\documentclass[lettersize,journal]{IEEEtran}
\usepackage{graphicx}
\usepackage[caption=false,font=footnotesize]{subfig}
\usepackage{amsmath,amsfonts}
\usepackage{algorithmic}
\usepackage{algorithm}
\usepackage{array}
\usepackage{textcomp}
\usepackage{stfloats}
\usepackage{url}
\usepackage{verbatim}
\usepackage{graphicx}
\usepackage{cite}
\usepackage{multirow} 
\usepackage{tabularx}
\usepackage{booktabs}

\begin{document}

\title{Saliency-Motion Guided Trunk-Collateral Network for Unsupervised Video Object Segmentation}

\author{Xiangyu Zheng, Wanyun Li, Songcheng He, Jianping Fan, Xiaoqiang Li, We Zhang*
\thanks{This work has been submitted to the IEEE for possible publication. Copyright may be transferred without notice, after which this version may no longer be accessible}
}

\maketitle

\begin{abstract}
Recent unsupervised video object segmentation (UVOS) methods predominantly adopt the motion-appearance paradigm, which leverages optical flow for motion cues and RGB image for appearance information to facilitate mask prediction. Mainstream motion-appearance approaches use either the bi-encoder structure to separately encode motion and appearance features, or the uni-encoder structure for joint encoding. However, these methods fail to properly balance the motion-appearance relationship. Consequently, even with complex fusion modules for motion-appearance integration, the extracted suboptimal features degrade the models' overall performance. Moreover, the quality of optical flow varies across scenarios, making it insufficient to rely solely on optical flow to achieve high-quality segmentation results. To address these challenges, we propose the \textbf{S}aliency-\textbf{M}motion guided \textbf{T}runk-\textbf{C}ollateral \textbf{Net}work (SMTC-Net), which better balances the motion-appearance relationship and incorporates model's intrinsic saliency information to enhance segmentation performance. Specifically, considering that optical flow maps are derived from RGB images, they share both commonalities and differences.Accordingly, we propose a novel Trunk-Collateral structure for motion-appearance UVOS. The shared trunk backbone captures the motion-appearance commonality, while the collateral branch learns the uniqueness of motion features. Furthermore, an Intrinsic Saliency guided Refinement Module (ISRM) is devised to efficiently leverage the model's intrinsic saliency information to refine high-level features, and provide pixel-level guidance for motion-appearance fusion, thereby enhancing performance without additional input. Experimental results show that SMTC-Net achieved state-of-the-art performance on three UVOS datasets (\textbf{89.2\%} \(\mathcal{J} \& \mathcal{F}\) on DAVIS-16, \textbf{76\%} \(\mathcal{J}\) on YouTube-Objects, \textbf{86.4\%} \(\mathcal{J}\) on FBMS) and four standard video salient object detection (VSOD) benchmarks with the notable increase, demonstrating its effectiveness and superiority over previous methods.
\end{abstract}


\begin{figure}[!t]
\centering
\subfloat[ \footnotesize Bi-Encoder Architecture]{\includegraphics[width=0.47\textwidth]{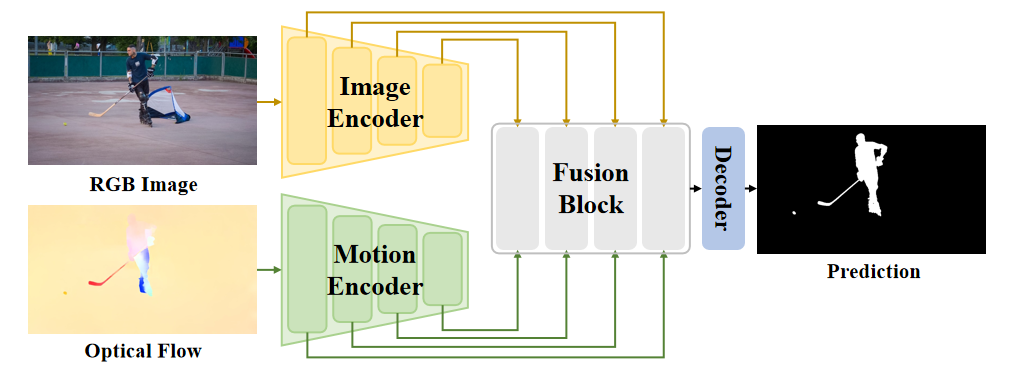}%
\label{bi-encoder fig}}
\hfil
\subfloat[ \footnotesize Uni-Encoder Architecture]{\includegraphics[width=0.47\textwidth]{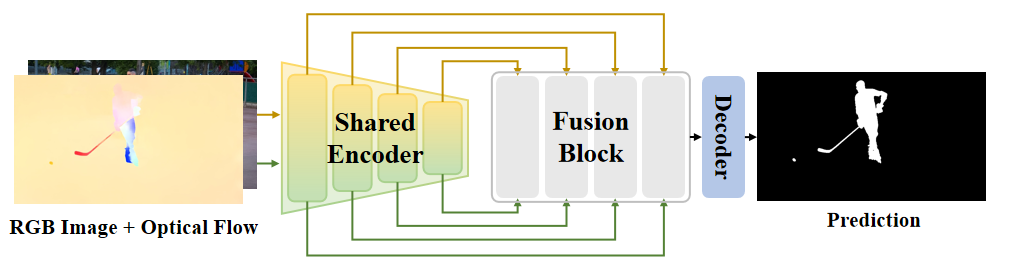}%
\label{Uni-Encoder fig}}
\hfil
\subfloat[ \footnotesize Trunk-Collateral Architecture]{\includegraphics[width=0.47\textwidth]{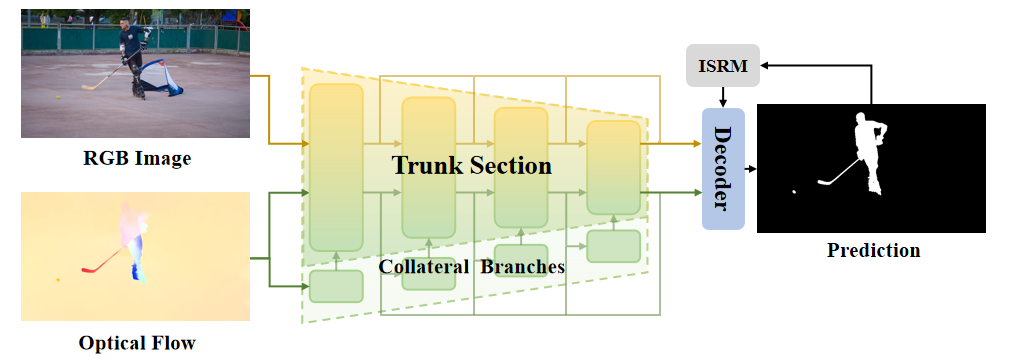}
\label{trunk-collateral fig}}
\caption{Illustration of typical motion-appearance UVOS paradigms and the proposed trunk-collateral architecture. (a) Bi-Encoder architecture with separate encoders for images and optical flows. (b) Uni-Encoder architecture with a shared encoder for images and optical flows. (c) Our Trunk-Collateral architecture which consists of both shared and specific parameters for images and optical flows.}
\label{UVOS manner}
\end{figure} 

\section{Introduction}
\IEEEPARstart{U}{nsupervised} video object segmentation (UVOS) requires generating a pixel-wise segmentation map for the most salient object within a video sequence. Unlike semi-supervised video object segmentation (SVOS), which relies on the ground truth annotation from the first frame to identify objects, UVOS focuses on autonomous segmentation without any prior information. Due to its robust generalization capabilities in unseen scenarios, UVOS has become a pivotal task in the field of computer vision, providing a solid foundation for video processing techniques and various real-world applications.

A fundamental challenge for UVOS is the necessity of generating precise pixel-wise segmentation maps without any prior knowledge. Optical flow offers per-pixel motion information between consecutive video frames, enabling the identification of moving objects, which are often considered as the salient targets in videos. Consequently, recent UVOS approaches\cite{MATNet, 3DCSeg, RTNet, FSNet, TransportNet, AMC-Net, HFAN, PMN, TMO, SimulFlow, DPA, GFA} predominantly utilize optical flow maps as motion features, combined with appearance features derived from RGB images to produce segmentation results. This methodology can be referred to as motion-appearance approaches or two-stream approaches.

Despite recent advances in motion-appearance techniques, several critical challenges still persist. First of all, the general architectures of motion-appearance approaches do not accommodate the relationship between motion and appearance features. Recent motion-appearance methods can be categorized into two types: bi-encoder architecture and uni-encoder architecture. As depicted in Fig. \ref{bi-encoder fig}, the bi-encoder architecture\cite{MATNet, RTNet, FSNet, AMC-Net, PMN, TMO, Isomer} employs distinct encoders to separately process RGB images and optical flow maps, yet overlooks the inherent motion-appearance correlations, leading to modality misalignment. In contrast, the uni-encoder architecture\cite{HFAN, SimulFlow, GFA}, as shown in Fig. \ref{Uni-Encoder fig}, encodes both features concurrently to reduce computational burden, but fails to capture the motion-appearance distinctions adequately. Neither of these two architectures can well enough balance motion-appearance relationship. Accordingly, previous methods often adopted complex interaction or fusion modules, either at the encoding stage or at the decoding stage, to compensate for the inferior feature extraction. Complicated fusion schemes bring about heavy model design, yet can not fundamentally resolve the problem, eventually leading to unsatisfactory performance.

\begin{figure}[!t]
    \centering
    \setlength{\tabcolsep}{1pt}
    \begin{tabular}{ccccc}
        &\textbf{Image} & \textbf{Mask} & \textbf{Optical Flow} \\ 
        \rotatebox{90}{\hspace{0.1cm} \textbf{Case1}} & 
        \includegraphics[width=0.15\textwidth]{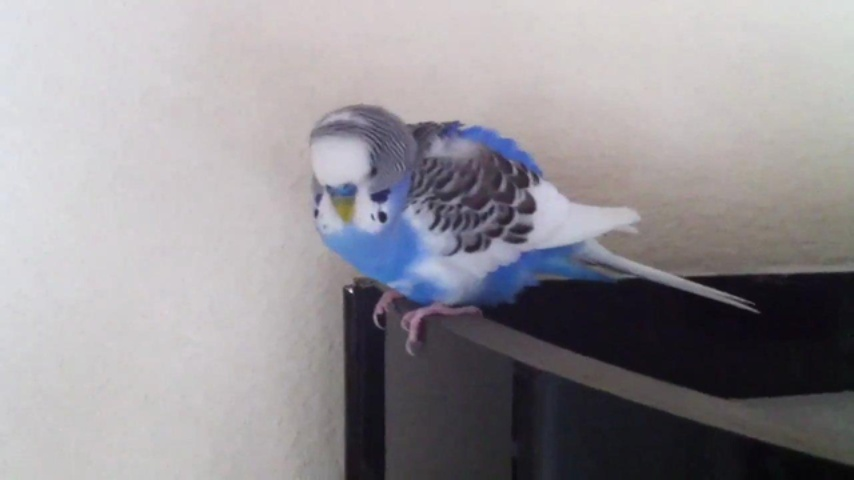} & 
        \includegraphics[width=0.15\textwidth]{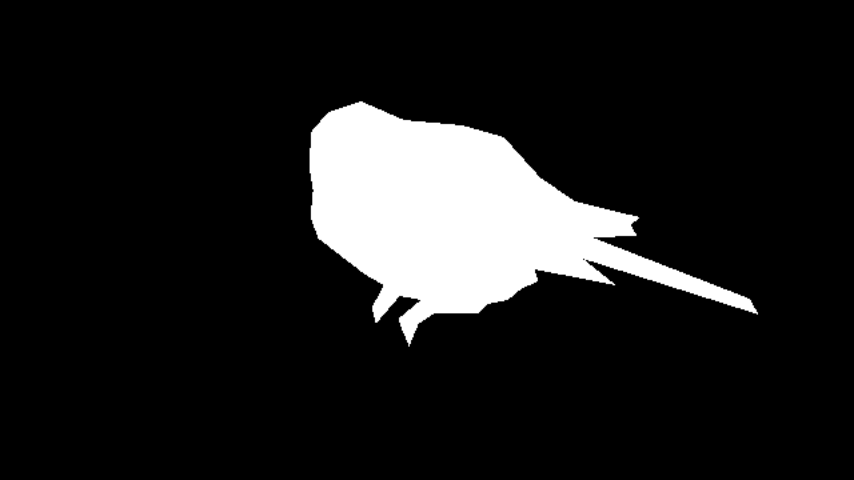} & 
        \includegraphics[width=0.15\textwidth]{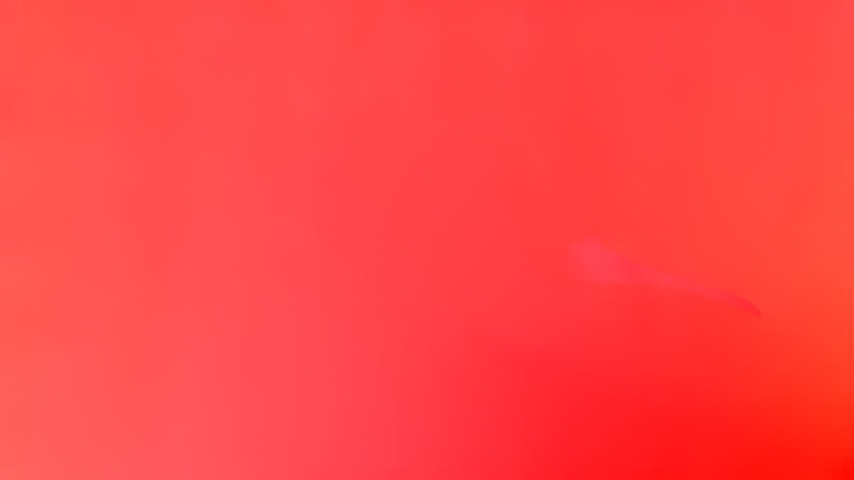} \\ 
        \rotatebox{90}{\hspace{0.1cm} \textbf{Case2}} & 
        \includegraphics[width=0.15\textwidth]{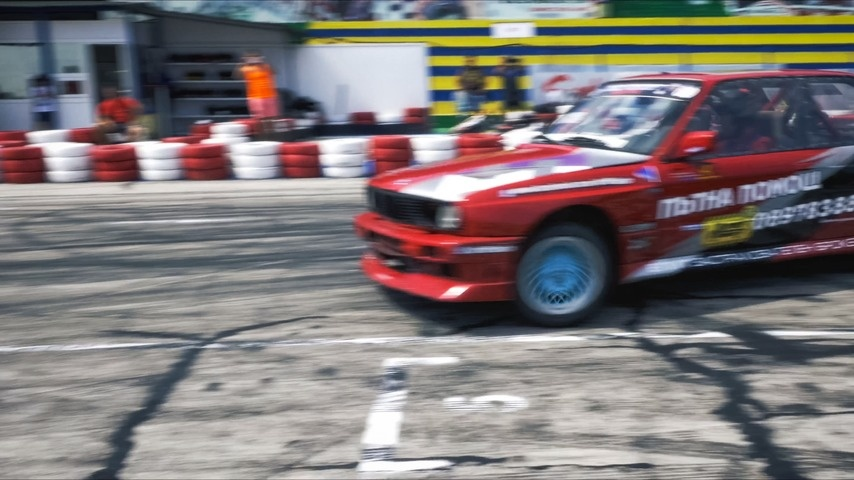} & 
        \includegraphics[width=0.15\textwidth]{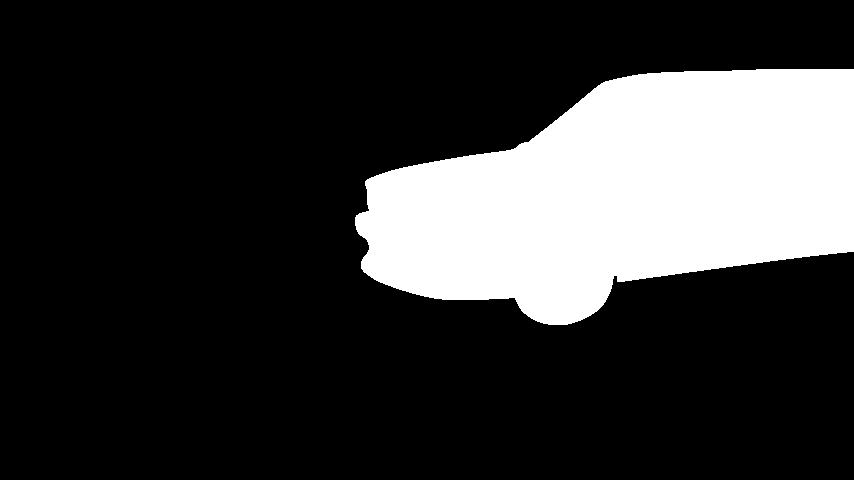} & 
        \includegraphics[width=0.15\textwidth]{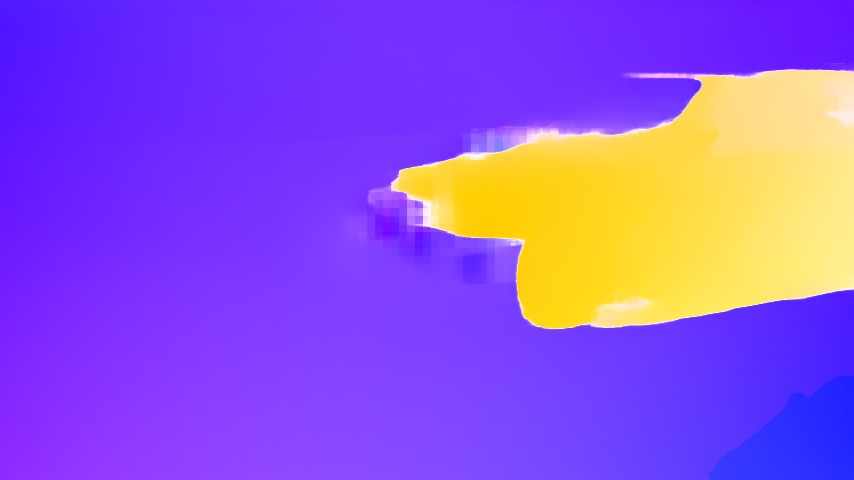} \\ 
        \rotatebox{90}{\hspace{0.1cm} \textbf{Case3}} & 
        \includegraphics[width=0.15\textwidth]{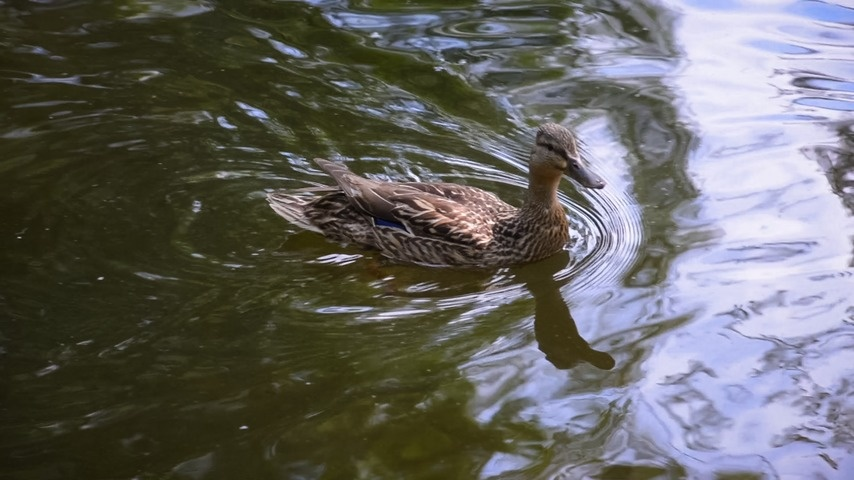} & 
        \includegraphics[width=0.15\textwidth]{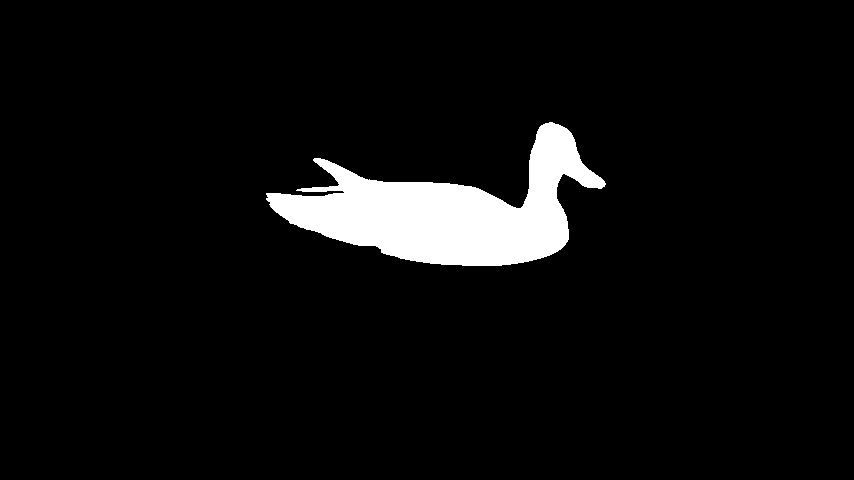} & 
        \includegraphics[width=0.15\textwidth]{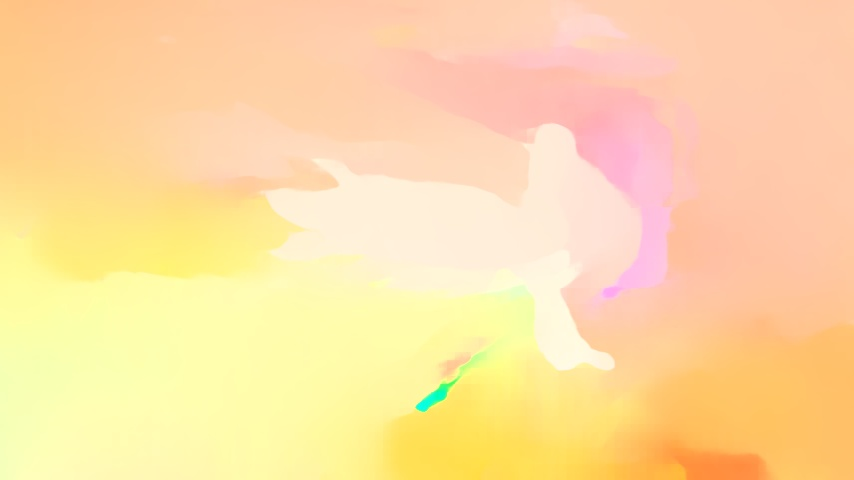} \\ 
        \rotatebox{90}{\hspace{0.1cm} \textbf{Case4}} & 
        \includegraphics[width=0.15\textwidth]{Image/Section1/failed_flow/case4_img.png} & 
        \includegraphics[width=0.15\textwidth]{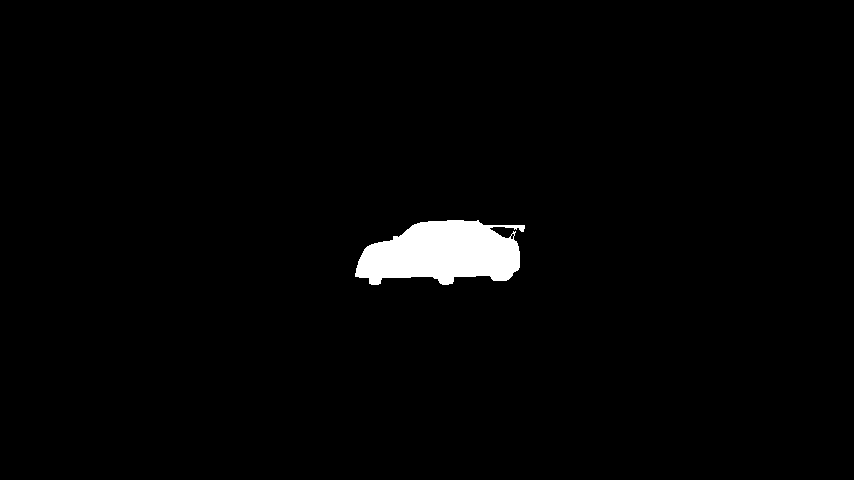} & 
        \includegraphics[width=0.15\textwidth]{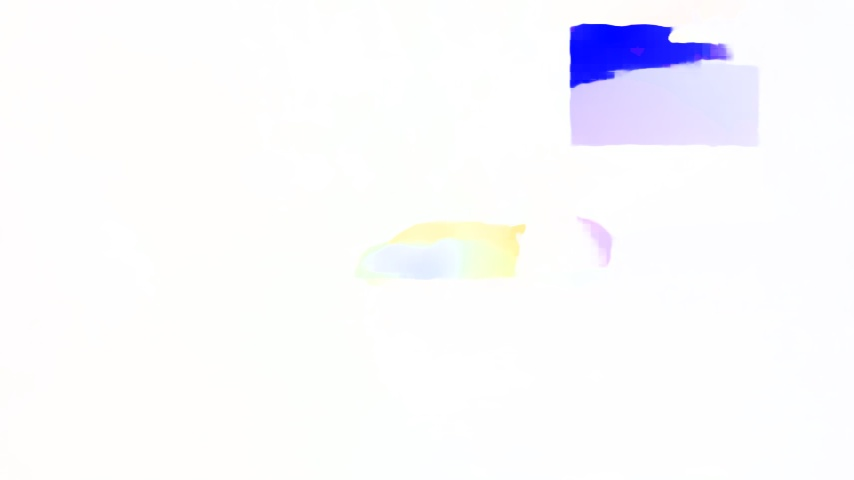} \\ 
    \end{tabular}
    \caption{Examples of optical flow maps in scenarios with partial estimation failure or suboptimal performance. Case 1–4 respectively represent stationary objects, motion blur, co-moving background, and misidentification of background objects.}
    \label{Flow fail fig}
\end{figure}

Secondly, relying solely on optical flow as auxiliary information is inadequate. Optical flow may fail to capture motion cues in scenarios involving stationary objects, motion blur, or co-moving background elements, as illustrated in Fig. \ref{Flow fail fig}. Furthermore, in the absence of instructive information, conventional fusion modules struggle to handle low-quality motion features. To address these limitations, additional information is essential. Saliency maps do not depend on motion cues to detect primary objects, which offer a natural complement in circumstances where optical flow fails. For example, saliency maps can accurately identify stationary foreground objects and filter out co-moving indistinctive elements. However, directly employing external saliency features, such as pre-generated saliency maps, is rather cumbersome. And simply integrating a new feature, like via a three-stream structure for instance, may introduce significant computational and memory overhead. Therefore, we argue that exploring a more efficient pattern to incorporate saliency features is worth further investigation.

In this study, we propose a novel saliency-motion guided trunk-collateral network (SMTC-Net) which effectively addresses the aforementioned challenges. To address the first challenge, \textbf{we introduce a brand-new trunk-collateral architecture tailored for motion-appearance UVOS methods}. This architecture, as shown in Fig. \ref{trunk-collateral fig}, leverages a shared backbone to jointly encode motion and appearance features, which captures their shared characteristics; complemented by a set of collateral branches to extract the motion-specific features that are distinct from those shared with appearance features. This innovative pattern simultaneously accounts for both motion-appearance similarity and divergence, eliminating the need for complex fusion modules. Theoretically, the parameter size of the collateral branch should remain relatively small to avoid negative impacts on the learning capacity of the trunk section. At the same time, it must sufficiently capture the motion uniqueness. Low-rank adaptation (LoRA)\cite{LoRA} is a low-rank and lightweight learning structure that can effectively captures additional information. The low-rank design also enables significant parameter reduction which avoids interfering with learning capacity of the main structure. Therefore, we implement LoRA in the design of the collateral branches, which realizes an optimal balance between computational efficiency and accuracy.

To address the second challenge, \textbf{we propose the simple yet effective intrinsic saliency guided refinement module (ISRM), which efficiently incorporates the model's intrinsic saliency to aid segmentation}. We observe that the segmentation results after simple decoding have already achieved a certain level of accuracy. Therefore, instead of using external saliency, we directly adopt the self-generated saliency in the first decoding round as an inherent instruction. In the second decoding round, ISRM leverages this saliency map to self-refine high-level semantic features and directs the motion-appearance integration. ISRM subtly leverages the model's internal saliency with straightforward design, enabling the model to reduce the negative impacts of unreliable optical flow while improving encoded features with high efficiency, contributing to superior segmentation results.

In summary, the main contributions of our work can be outlined as follows:

1.We propose an saliency-motion guided trunk-collateral network (SMTC-Net) which simultaneously adopts motion, appearance and saliency features for UVOS, and possesses better potential to manage the motion-appearance relationship while effectively incorporating saliency features.

2.We present a novel trunk-collateral architecture for motion-appearance UVOS, which considers both similarity and divergence of motion-appearance relationship without complex fusion schemes. Additionally, we propose the intrinsic saliency guided refinement module (ISRM), which leverages the model's internal saliency as an instruction to refine high-level representations and promote motion-appearance integration. Through careful design, our model achieves a balance between accuracy and speed.

3.Extensive experiments demonstrate the superiority of our SMTC-Net. SMTC-Net achieved state-of-the-art (SOTA) performance across all three widely used UVOS datasets (89.2\% \(\mathcal{J} \& \mathcal{F} \) on DAVIS-16\cite{DAVIS-16}, 76\% \(\mathcal{J}\) on YouTube-Objects\cite{YouTube-Objects}, and 86.4\% \(\mathcal{J}\) on FBMS\cite{FBMS}) and four standard video salient object detection (VSOD) benchmarks.

\section{Related Work}
\subsection{Incremental Structure}
Incremental structure refers to portions with extra parameters that are added on model baselines, in order to adapt to additional tasks or modal features. Typical applications of the incremental structure include continual learning and parameter-efficient fine-tuning (PEFT).

Continual learning requires models to constantly learn new knowledge or tasks. The challenge lies in avoiding catastrophic forgetting of previously learned knowledge while acquiring new knowledge. Parameter extension approaches\cite{extension1, extension2, extension3, extension4, extension5} are a typical category of continual learning. By allocating extra parameters for new tasks, these methods are capable of learning new information while retaining old knowledge. For instance, mixture-of-experts (MOE)\cite{expert1, expert2, expert3, expert4, expert5} assigns specialized experts to handle different tasks, which enables continuous growth of the model's knowledge through the increasing number of experts.

Incremental structure is also widely applied in parameter-efficient fine-tuning. Large language models (LLM) \cite{GPT2, GPT3, LLaMA, LLaMA2, LLaMA3} have become foundation models for various downstream applications, due to their striking performance in language understanding and processing. However, LLMs often have massive parameters, requiring substantial computational resources for full fine-tuning, which underscores the significance of PEFT. Some PEFT algorithms reduce the cost of fine-tuning by adding trainable parameters to the frozen baselines. For example, adapter-based methods \cite{adapter1, adapter2, adapter3, adapter4, adapter5, adapter6} insert trainable adapters among layers in the Transformer block, allowing the model to adjust to new tasks. LoRA-based approaches \cite{LoRA, adapter-lora1, adapter-lora2, adapter-lora3, adapter-lora4} add trainable modules comprising two low-rank matrices aside the models' fully connected layers, which empowers the model to accommodate to new assignments while remarkably reduces the computational burden. 

Our work innovatively applies the incremental structure to UVOS. By devising the trunk-collateral structure based on the intrinsic relationship between optical flow and image features, our SMTC-Net accounts for both motion-appearance commonality and differences, thus considerably enhancing the model performance.

\subsection{Unsupervised Video Object Segmentation}
As one of the foundational tasks in computer vision, video object segmentation (VOS) has witnessed numerous advancements over the past few years. A great number of representative works have emerged in both semi-supervised video object segmentation \cite{STCN, STM, Xmem, OneVOS, HFVOS} and unsupervised video object segmentation \cite{PDB, MOTAdapt, AGS, COSNet, AD-Net, AGNN, MATNet, WCS-Net, DFNet, 3DCSeg, F2Net, RTNet, FSNet, TransportNet, AMC-Net, 
 IMP, HFAN, PMN, TMO, SimulFlow, DPA, GFA}. Unlike SVOS, UVOS ought to automatically segment target objects from videos without any pre-defined information, which presents even more challenges.

Early non-learning UVOS methods rely on handcrafted features to distinguish between foreground and background objects. For example, background subtraction algorithms\cite{background1, background2, background3, background4} identify foreground objects by assuming the characteristics of background pixels. Point trajectory algorithms\cite{trajectory1, trajectory2, trajectory3, trajectory4} utilize long-term motion information in videos to obtain pixel trajectories, and cluster pixels accordingly to distinguish the main objects. Algorithms such as\cite{superpixel1, superpixel2, superpixel3} extract super-pixels from video frames based on pixel attributes like color, brightness, and image texture. Moreover, methods like\cite{object1, object2, object3} pre-generate foreground hypotheses for each frame as prior knowledge to assist segmentation.

With the advent of deep learning, learning-based methods have dominated the research of UVOS. Recent algorithms\cite{MATNet, WCS-Net, DFNet, 3DCSeg, F2Net, RTNet, FSNet, TransportNet, AMC-Net, IMP, HFAN, PMN, TMO, SimulFlow, DPA, GFA} primarily utilize two-stream networks to incorporate optical flow as the auxiliary motion information. These motion-appearance algorithms explicitly use both optical flow maps and RGB images as input, placing complex cross-modal fusion schemes to facilitate feature integration. For instance, MATNet\cite{MATNet} put forward an asymmetric attention module to differentially enhance motion and appearance features. RTNet\cite{RTNet} proposed a spatial temporal attentive fusion module to selectively blend two features. HFAN\cite{HFAN} introduced a feature alignment module and a feature adaptation module to align motion and appearance features at different levels within the encoder. SimulFlow\cite{SimulFlow} associated the motion-appearance fusion with the self-attention component of the Transformer block to facilitate merging. However, these merging-based algorithms have not adequately addressed the relationship between motion and appearance features. Consequently, their reliance on complicated fusion modules only yielded marginal improvements, yet critically hindering model running speed.

Our SMTC-Net differs from traditional motion-appearance UVOS approaches by designing a novel trunk-collateral structure which better corresponds to the intrinsic relationship between motion and appearance features, thus eliminating the need for complex fusion mechanisms. Furthermore, we exploit the model's inherent saliency to optimize high-level representations and motion-appearance combination, which further elevates the model effects.

\section{Method}

\begin{figure*}[!t]
\centering
\includegraphics[width=7in]{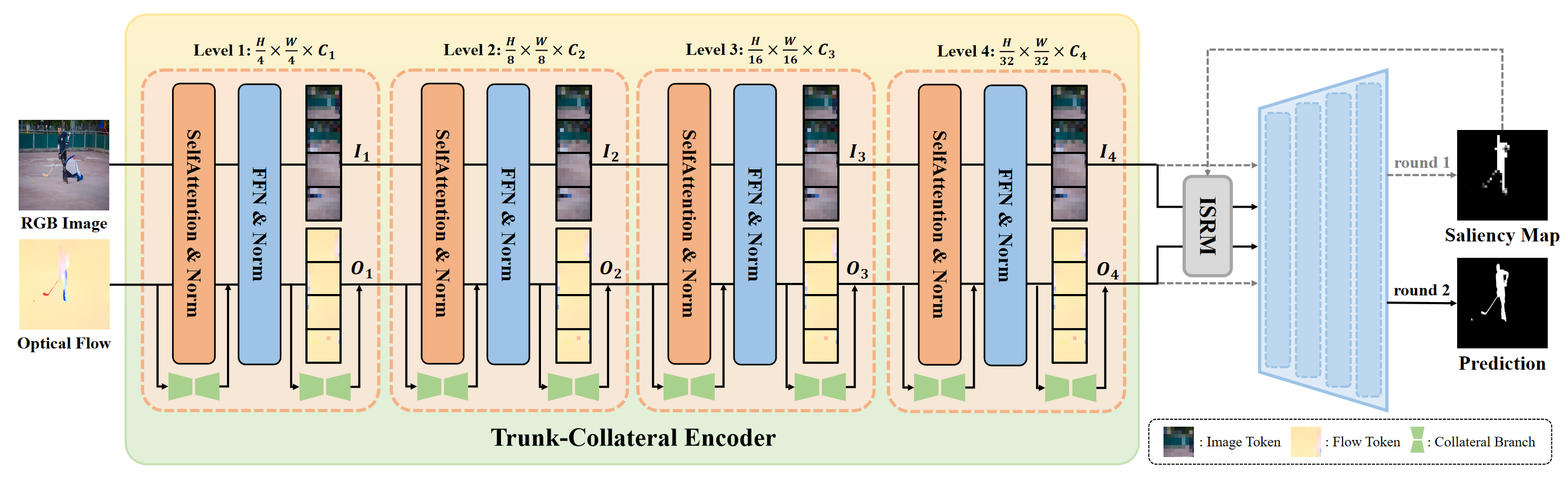}
\caption{Overall pipeline of SMTC-Net, which consists of a trunck-collateral encoder, an intrinsic saliency guided refinement module (ISRM) and a multi-level decoder. Given an input image and its corresponding optical flow, the trunk-collateral encoder captures the shared attributes of motion and appearance features through the common trunk section, while encoding distinctive motion characteristics via the collateral branches alongside the trunk section. ISRM optimizes feature representations and enhances motion-appearance integration using the intrinsic saliency feature generated in the first decoding round. The ultimate segmentation result is output in the second decoding round.}
\label{Overal_Pipeline}
\end{figure*}
\subsection{Overall Pipeline}

In this section, we first provide a comprehensive overview of our model pipeline, which is illustrated in Fig. \ref{Overal_Pipeline}. Specifically, the proposed saliency-motion guided trunk-collateral network (SMTC-Net) is an end-to-end UVOS framework based on the encoder-decoder architecture. Given an input video frame \( I \in \mathbb{R}^{H \times W \times 3}\) and its corresponding optical flow map \( O \in \mathbb{R}^{H \times W \times 3}\), SMTC-Net processes \( I \) and \( O \) to generate a segmentation mask \( M_{\text{pred}} \in \mathbb{R}^{H \times W \times 1}\).

To enhance feature extraction and representation, we adopt a multi-level encoder-decoder pattern following prior works \cite{MATNet, RTNet, HFAN, TMO, SimulFlow}, which enables the model to capture multiscale features and integrate both texture and semantic information. During the encoding phase, the appearance and motion features at the \( i \)-th level (\( i \in \{1, 2, 3, 4\} \)) are represented as \( I_{i} \in \mathbb{R}^{H_{i}W_{i} \times C_{i}}\) and \( O_{i} \in \mathbb{R}^{H_{i}W_{i} \times C_{i}}\) respectively, where \( H_iW_i = \frac{H}{2^{i+1}} \times \frac{W}{2^{i+1}} \). The appearance feature \( I_{i} \) and motion feature \( O_{i} \) extracted at the \( i \)-th encoder level are fused to produce a combined feature representation \( F_{i} \). This fused feature \( F_{i} \) is subsequently passed to the 
\( i \)-th decoder level for decoding.

Our multi-level decoder is constructed following the design in\cite{TMO}. After the encoding stage, the encoded features \( I_i \) and \( O_i \) are processed through the decoder in two iterations. During the decoding phase, features \( F'_{i} \) at level \(i\) are derived by directly adding \( I_i \) and \( O_i \), followed by concatenation with the feature \( F_{i-1} \) from the previous level. To further refine the merged features, \( F'_{i} \) is passed through a convolutional layer and a CBAM module\cite{CBAM}, which enhances feature fusion through attention mechanisms. The resulting feature is then upsampled by a factor of two to produce \( F_{i} \), which is subsequently passed to the next decoding level. 

After the final decoding level, the channel number of \( F_1 \) is reduced to from \text{$C_{1}$} to 1 via a convolution layer. The feature \( F_1 \) is then upsampled to match the spatial resolution of the input, yielding the segmentation prediction \( S \). The output \( S \) is a non-binarized probability map that represents the pixel-wise likelihood of belonging to the foreground object. As such, it can be regarded as an intrinsic saliency feature, which can effectively facilitate the differentiation between foreground and background objects. The general decoding process is expressed in \eqref{decoder_level}

\begin{equation}
\label{decoder_level}
\begin{aligned}
    F'_i & = I_i + O_i \\
    F''_i & =
    \begin{cases}
        \text{Concat}(F'_i, F_{i-1}), & i \in \{1, 2, 3\} \\
       \text{Concat}(F'_i), & i = 4
    \end{cases} \\[1ex]
    F_i & = \text{CBAM}(\text{ConvReLU}(F''_i))\\
    S & = \text{UpScale}(\text{Conv}_{C_{1} \rightarrow \text{1}}(F_{1}))\\
\end{aligned}
\end{equation}

During the first decoding round, \( I_i \) and \( O_i \) are directly processed by a naive decoder without leveraging the intrinsic saliency guided refinement module (ISRM). In the second round, the saliency map \( S \) generated during the first decoding round is utilized as input to the ISRM to refine the high-level features \( I_4 \) and \( O_4 \). This refinement produces the enhanced merged feature \(F^{r}_4\). Subsequently, the refined feature \(F^{r}_4\) and the feature \( F'_i \) for \( i \in \{1, 2, 3\} \) are propagated through the decoder again to generate the final segmentation mask \( M_{\text{pred}} \).
\begin{equation}
\begin{gathered}
\label{trunk-collareral general}
 S = \text{Decoder}( F'_1, F'_2, F'_3, F'_4)\\
 F^{r}_4= \text{ISRM}( I_4, O_4, S) \\
 M_{\text{pred}} = \text{Decoder}( F'_1, F'_2, F'_3, F^{r}_4)\\
\end{gathered}
\end{equation}

\subsection{Trunk-Collateral Paradigm}

\begin{figure}[t]
\centering
\includegraphics[width=3.5in]{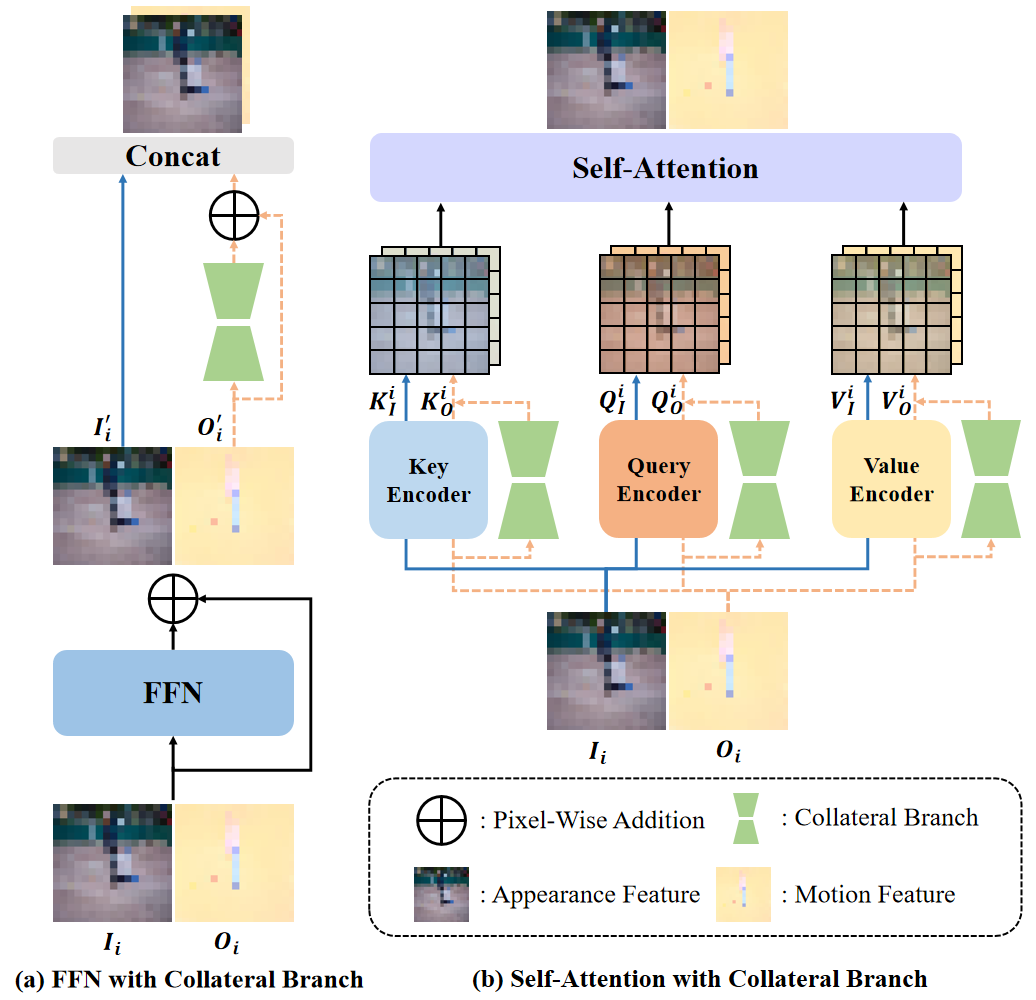}
\caption{Illustration of the trunk-collateral structure in the Transformer block. (a) Trunk-collateral structure in the feed-forward network. (b) Trunk-collateral structure in the multi-head self-attention.}
\label{Trunk_Collateral_Module}
\end{figure}

As illustrated in Fig. \ref{trunk-collateral fig} and Fig. \ref{Overal_Pipeline}, we propose a novel trunk-collateral paradigm for motion-appearance UVOS, formulated to effectively capture both the similarities and divergences between motion and appearance features. The trunk component, denoted as \( P_{\text{trunk}} \), serves as a shared backbone responsible for encoding both motion and appearance features, thereby facilitating the extraction of their common characteristics. Complementing this, the collateral branch \(P_{\text{collateral}}\) is specifically assigned for processing motion features, isolating the unique characteristics of motion. These distinct motion features are then integrated with the trunk-encoded motion characteristics to generate the final motion representation. Following prior works \cite{HFAN, TMO, DTTT}, we adopt Segformer\cite{Segformer} as the backbone, which constitutes the trunk component \( P_{\text{trunk}} \). To align with the Transformer-based \cite{Transformer} architecture, the collateral branch \(P_{\text{collateral}}\) is incorporated into every dense layer within the Transformer blocks, including the feed-forward network (FFN) and multi-head self-attention (MHSA). Fig. \ref{Trunk_Collateral_Module} provides a detailed visualization of the trunk-collateral paradigm applied within the Transformer-like structure. Specifically, Fig. \ref{Trunk_Collateral_Module} (a) and Fig. \ref{Trunk_Collateral_Module} (b) illustrate the implementation of the trunk-collateral paradigm within the FFN and MHSA, respectively.

In general, taking an image \( I \) and its corresponding optical flow \( O \) as inputs, multiscale features \( I_i \)( \( i \in \{1, 2, 3, 4\} \) ) and \( O_i \)(\( i \in \{1, 2, 3, 4\} \)) are encoded as:

\begin{equation}
\begin{gathered}
\label{trunk-collareral general}
O_i = P_{\text{trunk}}(O) + P_{\text{collateral}}(O) \\
I_i = P_{\text{trunk}}(I)
\end{gathered}
\end{equation}

Specifically, a linear layer is typically applied to \( I_{i}\in \mathbb{R}^{H_iW_i  \times C_{i}} \) and \( O_{i} \in \mathbb{R}^{H_iW_i \times C_{i}}\) to generate their key, query, and value features during MHSA in each Transformer block. In addition, the collateral branches \(P_{\text{collateral}}\) are incorporated alongside these linear layers exclusively for processing \( O_{i} \). Taking key generation as an example, the encoding of key features \( K^{i}_{I}\) and \( K^{i}_{O}\) for \( I_{i}\) and \( O_{i}\) respectively, is performed as expressed in \eqref{key-gen equation}. 

\begin{equation}
\begin{gathered}
\label{key-gen equation}
K^{i}_{O} = \text{Linear}_{K}(O_{i}) + P_{\text{collateral}}(O_{i}) \\
K^{i}_{I} = \text{Linear}_{K}(I_{i}) \\
\end{gathered}
\end{equation}

Rather than directly appending a collateral branch alongside the FFN, we take the features \( I'_{i}\) and \( O'_{i}\), which are outputs of the FFN combined with the residual connection as in \eqref{FFN_res}, for further refinement. 

\begin{equation}
\begin{gathered}
\label{FFN_res}
O'_{i}  = \text{FFN}(O_{i}) +  O_{i}\\
I'_{i}  = \text{FFN}(I_{i}) +  I_{i}\\
\end{gathered}
\end{equation}

As demonstrated in \eqref{FFN}, this process encompasses the scenario of directly integrating \(P_{\text{collateral}}\) to FFN, while simultaneously providing more flexibility for adaptive learning.

\begin{equation}
\begin{aligned}
\label{FFN}
O_{i} & = O'_{i} + P_{\text{collateral}}(O'_{i})\\
 & = \text{FFN}(O_{i}) +  O_{i} + P_{\text{collateral}}(\text{FFN}(O_{i}) +  O_{i}) \\ 
 & = \mathbf{O_{i} + P_{\text{collateral}} (O_{i})} \\
 & + \text{FFN}(O_{i})  + P_{\text{collateral}} (\text{FFN}(O_{i}))\\
\end{aligned}
\end{equation}

The collateral branch \(P_{\text{collateral}}\) functions as an auxiliary component to extract distinct motion characteristics, independent of the shared motion-appearance representation. Accordingly, it is plausible to hypothesize that \(P_{\text{collateral}}\) operates within a low-dimensional intrinsic space. As shown in Fig. \ref{Trunk_Collateral_Module}, we implement \(P_{\text{collateral}}\) following the principles of low-rank adaptation \cite{LoRA}, thereby significantly reducing computational complexity. The low-rank adaptation procedure is formalized in \eqref{Lora}, where \(B \in \mathbb{R}^{d \times r}\) , \(A \in \mathbb{R}^{r \times k}\), with the condition \(r \ll \min(d, k)\).

\begin{equation}
\label{Lora}
    f(x) = W_0 x + \Delta W x = W_0 x + B A x
\end{equation}

\subsection{Intrinsic Saliency Guided Refinement Module}
\begin{figure*}[!t]
\centering
\includegraphics[width=6.5in]{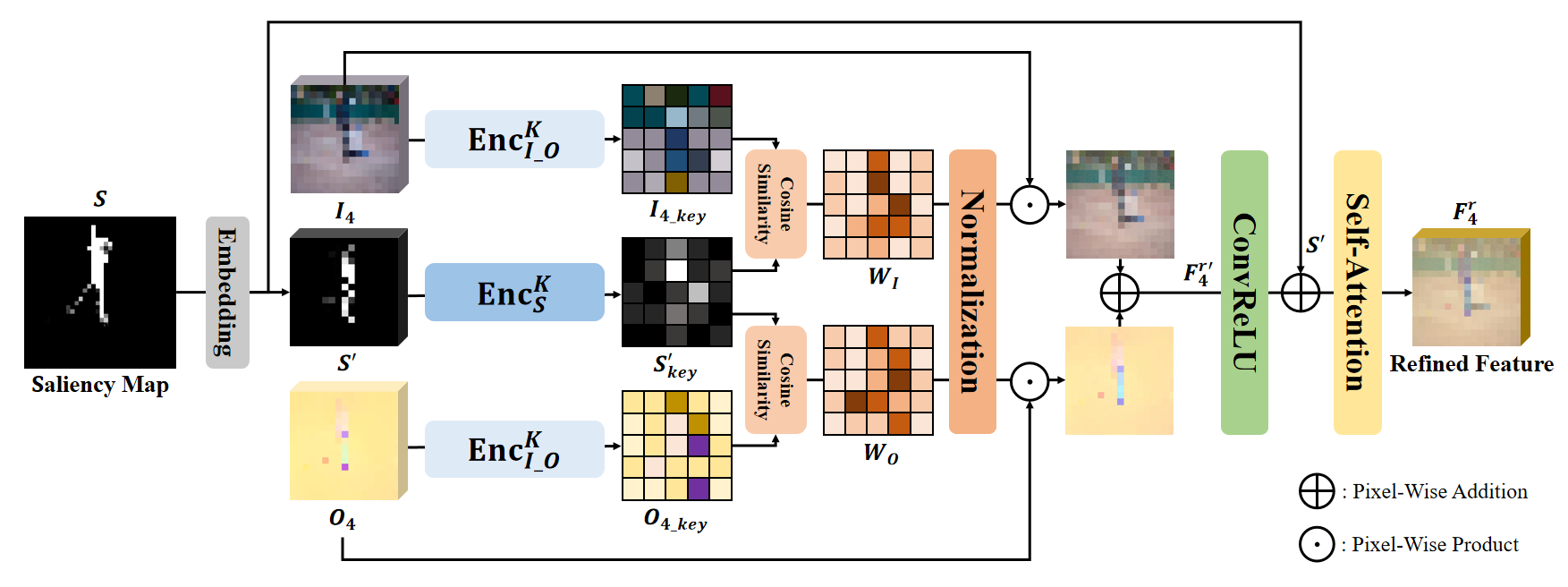}
\caption{Illustration of the intrinsic saliency guided refinement module (ISRM). ISRM utilizes the intrinsic saliency feature produced in the first decoding round to optimize high-level representations, and guide motion-appearance integration to advance the ultimate performance.}
\label{ISRM}
\end{figure*}

We further introduce an intrinsic saliency guided refinement module (ISRM) to effectively take advantage of the model's inherent saliency feature, as illustrated in Fig. \ref{ISRM}. After the encoding phase, multiscale image features \( I_i \) for \( i \in \{1, 2, 3, 4\} \) and motion features \( O_i \) for \( i \in \{1, 2, 3, 4\} \) are passed to the decoder. In the initial decoding round, ISRM is not applied, and a single-channel map \( S \) of the original image size is generated. This rough segmentation prediction map \( S \) is utilized as the intrinsic saliency feature, revealing the per-pixel probability of belonging to the foreground object.

In the second decoding round, ISRM leverages \( S \) to guide the refinement and fusion of high-level features \( O_4 \) and \( I_4 \). Initially, \( S \) is embedded to \( S' \in \mathbb{R}^{ \frac{H_{i}}{32} \times \frac{W_{i}}{32} \times C_{i}}\) through a convolution layer. Two distinct key encoders are employed to extract key features \( S'_{\text{key}} \), \( O_{4\_\text{key}} \), and \( I_{4\_\text{key}} \) from \( S' \), \( O_4 \), and \( I_4 \), respectively, as expressed in \eqref{SGR key-gene}.

\begin{equation}
\begin{gathered}
\label{SGR key-gene}
S' = \text{ConvReLu}(S) \\ 
S'_{\text{key}} = \text{Enc}_{S}^{K}(S') \\
O_{4\_\text{key}} = \text{Enc}_{{I\_O}}^{K}(O_4) \\
I_{4\_\text{key}} = \text{Enc}_{{I\_O}}^{K}(I_4) \\
\end{gathered}
\end{equation}

The similarity maps \( W_o \in \mathbb{R}^{\frac{H}{32} \times \frac{W}{32}}\) and \( W_i \in \mathbb{R}^{\frac{H}{32} \times \frac{W}{32}}\) are derived by computing the pixel-wise cosine similarity between \( O_{4\_\text{key}} \), \( I_{4\_\text{key}} \), and \( S'_{\text{key}} \). Since \( S' \) provides relatively reliable guidance, the map \( W_o \) and \( W_i \) represent the pixel-wise confidence of the feature maps \( O_4 \) and \( I_4 \). Subsequently, \( W_o \) and \( W_i \) are normalized and applied as the summation weights to \( O_4 \) and \( I_4 \), leading to the final combined feature \( F_{4o} \).

\begin{equation}
\begin{gathered}
\label{weighted sum}
W_o = \text{Cosine\_Sim}( S'_{\text{key}}, O_{4\_\text{key}} )\\
W_i = \text{Cosine\_Sim}( S'_{\text{key}}, I_{4\_\text{key}} )\\
W_o = \frac{W_o}{W_o + W_i +1e^{-6}}\\
W_i = \frac{W_i}{W_o + W_i +1e^{-6}}\\[1ex]
F^{r'}_{4} = W_o \cdot O_4 + W_i \cdot I_4
\end{gathered}
\end{equation}

We apply self-attention to the sum of \( F^{r'}_{4} \) and \( S' \) to further facilitate information interaction and refinement. This process produces the final saliency-guided merged feature \( F^{r}_{4} \) for the second decoding phase. Features \( F'_i \) for \( i \in \{1, 2, 3\} \) are computed in the same way as in the first decoding round.

\begin{equation}
\begin{gathered}
\label{feature refinement}
F^{r}_{4}=  \text{Self\_Attention}(F^{r'}_{4} +S')
\end{gathered}
\end{equation}

\subsection{Loss Function}

Different from former works, our training loss is formulated as a weighted linear combination of the focal loss \( L_{Focal} \), binary cross-entropy loss \( L_{BCE} \), and Dice loss \( L_{Dice} \). Focal loss effectively mitigates the class imbalance between positive and negative samples, enabling the model to prioritize hard-to-classify pixels. Dice loss provides a more holistic measure of segmentation performance beyond pixel-level accuracy, and is particularly effective in addressing extreme data imbalances. When combined with the traditional BCE Loss, our loss function facilitates smoother model training, improves management of the imbalance between positive and negative samples, and enhances the model's performance in segmenting small objects.

During training, loss is calculated between the predicted mask \(M_{\text{pred}}\) and the goundtruth \(M_{\text{gt}}\). Additionally, we apply supervision to the output saliency map \(S\) generated in the first decoding round with a weight of \(\omega=0.3 \). Our general loss function is given in \eqref{loss}, with hyperparameters \(\alpha=20, \beta=10\) and \(\gamma=1\).

\begin{equation}
\begin{gathered}
\label{loss}
L_{\text{loss}} = \alpha \cdot L_{\text{Focal}} +\beta \cdot L_{\text{BCE}} + \gamma \cdot L_{\text{Dice}} \\
L_{\text{total}}= L_{\text{loss}}(M_{\text{pred}}, M_{\text{gt}}) + \omega \cdot L_{\text{loss}}(S, M_{\text{gt}})
\end{gathered}
\end{equation}

\section{Experiments}

\subsection{Implementation Details}

\subsubsection{Traning and Inference} 
The training process of SMTC-Net comprises two phases. First, we pre-train the model on the YouTube-VOS dataset \cite{YouTube-VOS} for 50 epochs to enhance its generalization capability. Subsequently, the model is fine-tuned on specific downstream datasets, such as DAVIS-16 \cite{DAVIS-16} or FBMS \cite{FBMS}, to evaluate its performance on the UVOS task. Furthermore, to assess the model's effectiveness on the VSOD task, we fine-tune the pre-trained model on a mixed dataset comprising DAVIS-16 and DAVSOD \cite{DAVSOD} like the preceding works, or directly adopt the fine-tuned UVOS model for testing. We employ the AdamW optimizer \cite{AdamW}. The learning rate is fixed to \( 6 \times 10^{-5} \) during the pre-training phase,  and adjusted to \( 1 \times 10^{-4} \) with a learning rate decay strategy in the fine-tuning phase. To enhance training robustness, data augmentation techniques, such as horizontal flipping, are applied. The model is trained on two NVIDIA 3090 GPUs, with a mini-batch size of 16 during pre-training and 8 during fine-tuning.

During both training and inference phases, input images and optical flow maps are resized to \( 512 \times 512 \) before being fed into the model. No data augmentation techniques are applied during the inference phase. In line with prior approaches\cite{HFAN, TMO, SimulFlow, DPA, GSA}, we utilize RAFT\cite{RAFT} to pre-generate the optical flow maps for each video, which are subsequently stored for training and inference. In scenarios where multiple segmentation targets are present within a single video, we treat all targets as a single entity for segmentation during both training and inference.

\subsubsection{Evaluation Metrics}
We evaluate our model's performance on the UVOS task by adhering to the validation methodologies employed in prior VOS algorithms. Specifically, region similarity \(\mathcal{J}\) quantifies the overall segmentation accuracy during inference by calculating the intersection over union (IoU) between the predicted mask \(M_{\text{pred}}\) and the ground truth \(M_{\text{gt}}\) as in \eqref{J}.

\begin{equation}
\label{J}
\mathcal{J} = \frac{|M_{\text{gt}} \cap M_{\text{pred}}|}{|M_{\text{gt}} \cup M_{\text{pred}}|}
\end{equation}

Meanwhile, boundary accuracy \( \mathcal{F}\) evaluates the precision of the predicted mask's boundaries by computing the F1 score between the edge maps extracted from the predicted masks \(M_{\text{pred}}\) and ground truth \(M_{\text{gt}}\), as defined in equation \eqref{F}. Furthermore, the combined metric \(\mathcal{J} \& \mathcal{F} \), which represents the average of \(\mathcal{J}\) and \(\mathcal{F} \), provides a comprehensive assessment of the model's overall performance.

\begin{equation}
\begin{gathered}
\label{F}
\text{Precision} = \frac{|M_{\text{gt}} \cap M_{\text{pred}}|}{|M_{\text{pred}}|} \quad
\text{Recall} = \frac{|M_{\text{gt}} \cap M_{\text{pred}}|}{|M_{\text{gt}}|}\\
\mathcal{F} = \frac{2 \times \text{Precision} \times \text{Recall}}{\text{Precision} + \text{Recall}}
\end{gathered}
\end{equation}

For the VSOD task, we also use standard evaluation metrics to assess the performance of SMTC-Net, including the mean absolute error ($MAE$), maximum F-measure ($F_m$), maximum enhanced-alignment measure ($E_m$) and structure-measure ($S_m$). To be more specific, $MAE$ computes the average absolute error on a per-pixel basis; $F_m$ integrates precision and recall under the optimal threshold; $E_m$ accounts for both pixel-level and image-level discrepancies, and $S_m$ quantifies structural similarity from both region-aware and object-aware perspectives.

\subsubsection{Datasets} 
We provide an introduction of the datasets utilized in our work. SMTC-Net is pre-trained on the YouTube-VOS dataset and fine-tuned for specific tasks. For the UVOS task, the model is evaluated on the test sets of DAVIS-16, FBMS, and YouTube-Objects. For the VSOD task, we evaluate the performance of SMTC-Net on the test sets of DAVIS-16, FBMS, ViSal, and DAVSOD.

\textbf{Youtube-VOS}
The YouTube-VOS dataset \cite{YouTube-VOS} is one of the most widely used benchmarks for video object segmentation (VOS). Its training set contains 3,471 videos spanning 65 different categories, encompassing a total of 6,459 unique object instances. The validation and test sets each include over 500 videos, featuring both seen and unseen categories relative to the training set. However, as the validation and test sets of YouTube-VOS are specifically designed for the SVOS task and only provide ground truth for the first frame, we utilize YouTube-VOS exclusively for training in this work.

\textbf{DAVIS-16}
The DAVIS-16 dataset \cite{DAVIS-16} consists of a training set with 30 video clips and a validation set with 20 video clips, primarily focusing on single-object segmentation. As one of the most widely employed benchmarks in VOS research, it facilitates direct and consistent comparisons among different VOS models.

\textbf{FBMS} The FBMS dataset \cite{FBMS} contains 59 video clips, divided into 29 training clips and 30 test clips, with annotations provided for 720 frames. 

\textbf{YouTube-Objects} The YouTube-Objects dataset \cite{YouTube-Objects}, compiled from videos sourced from YouTube, consists of 10 semantic categories, with each category containing 9 to 24 video clips. Since the dataset does not provide explicit splits for training and validation, prior UVOS studies typically use the entire dataset for testing purposes. 

\textbf{ViSal} The ViSal dataset \cite{ViSal} is a VSOD dataset which includes 17 video sequences, totaling 193 annotated frames, which spans various categories such as humans, animals, and vehicles. 

\textbf{DAVSOD} The DAVSOD dataset \cite{DAVSOD} includes 61 training videos and 41 validation videos, covering a wide range of real-world scenes, objects, and actions for the VSOD task.

\begin{table*}[!t]
    \centering
    \caption{Quantitative results and comparative analysis on the UVOS benchmarks: DAVIS-16, FBMS, and YouTube-Objects. In the table, 'OF' denotes the use of optical flow estimation models, while 'PP' represents the application of post-processing techniques. The top two performances are highlighted by \textbf{bold} and \underline{underline} respectively}
    \renewcommand{\arraystretch}{1.25}
    \begin{tabularx}{\textwidth}{p{2.5cm}p{2cm}p{2.5cm}XXXXXXX} 
        \hline
        \multirow{2}{*}{\textbf{Method}} & \multirow{2}{*}{\textbf{Publication}} & \multirow{2}{*}{\textbf{Backbone}} & \multirow{2}{*}{\textbf{OF}} & \multirow{2}{*}{\textbf{PP}} & \multicolumn{3}{c}{\textbf{DAVIS-16}} & \textbf{FBMS} & \textbf{YTO} \\ 
        \cline{6-10}
        & & & & & \(\mathcal{J} \& \mathcal{F}\) & \(\mathcal{J}\) & \(\mathcal{F}\) & \(\mathcal{J}\) & \(\mathcal{J}\)\\
        \hline
        PDB\cite{PDB} & ECCV'18 & ResNet-50 & &  \checkmark &75.9 & 77.2 & 74.5 & 74.0 & -  \\ 
        MOTAdapt\cite{MOTAdapt} & ICRA'19 & ResNet-18 &  & \checkmark &77.3 & 77.2 & 77.4 & - & - \\ 
        AGS\cite{AGS} & CVPR'19 & ResNet-101 &  &\checkmark & 78.6 & 79.7 & 77.4 & - & 69.7 \\ 
        COSNet\cite{COSNet} & CVPR'19 & DeepLabv3 & & \checkmark & 80.0 & 80.5 & 79.4 & 75.6 & 70.5 \\ 
        AD-Net\cite{AD-Net} & ICCV'19 & DeepLabv3 & &\checkmark & 81.1 & 81.7 & 80.5 & - & - \\ 
        AGNN\cite{AGNN} & ICCV'19 & DeepLabV3 & & \checkmark  & 79.9 & 80.7 & 79.1 & - & 70.8 \\ 
        MATNet\cite{MATNet} & AAAI'20 & ResNet-101 & \checkmark & \checkmark & 81.6 & 82.4 & 80.7 & 76.1 & 69.0 \\ 
        WCS-Net\cite{WCS-Net} & ECCV'20 & ResNet-101 & & & 81.5 & 82.2 & 80.7 & - & 70.5 \\ 
        DFNet\cite{DFNet} & ECCV'20 & DeepLabv3 & & \checkmark &  82.6 & 83.4 & 81.8 & - & - \\ 
        RTNet\cite{RTNet} & CVPR'21 & ResNet-101 & \checkmark &\checkmark & 85.2 & 85.6 & 84.7 & - & 71.0 \\ 
        FSNet\cite{FSNet} & ICCV'21 & ResNet-50 & \checkmark & \checkmark & 83.3 & 83.4 & 83.1 & - & - \\ 
        TransportNet\cite{TransportNet} & ICCV'21 & ResNet-101 & \checkmark & & 84.8 & 84.5 & 85.0 & 78.7 & - \\ 
        AMC-Net\cite{AMC-Net} & ICCV'21 & ResNet-101 & \checkmark & \checkmark & 84.6 & 84.5 & 84.6 &  76.5 & 71.1 \\ 
        IMP\cite{IMP} & AAAI'22 & ResNet-50 &  &  & 85.6 & 84.5 & 86.7 & 77.5 & -\\ 
        HFAN\cite{HFAN} & ECCV'22 & Mit-b1 & \checkmark &  & 86.7 & 86.2 & 87.1 & - & 73.4 \\ 
        HCPN\cite{HCPN} & TIP'23 & ResNet-101 & \checkmark & \checkmark & 85.6 & 85.8 & 85.4 & 78.3 & 73.3 \\
        PMN\cite{PMN} & WACV'23 & VGG-16 & \checkmark &  & 85.9 & 85.4 & 86.4 & 77.7 & 71.8 \\ 
        TMO\cite{TMO} & WACV'23 & ResNet-101 & \checkmark &  & 86.1 & 85.6 & 86.6 & 79.9 & 71.5\\ 
        OAST\cite{OAST} & ICCV'23 & MobileViT & \checkmark & & 87.0 & 86.6 & 87.4 & 83.0 & - \\
        SimulFlow\cite{SimulFlow} & ACMMM'23 & Mit-b1 & \checkmark &  & 87.4 & 86.9 & 88.0 & 80.4 & 72.9 \\ 
        HGPU\cite{HGPU} & TIP'24 &  ResNet-101 & \checkmark &  & 86.1 & 86.0 & 86.2 & - & 73.9 \\
        DPA\cite{DPA} & CVPR'24 & VGG-16 & \checkmark &  & 87.6 & 86.8 & 88.4 & \underline{83.4} & 73.7\\  
        GSA\cite{GSA} & CVPR'24 & ResNet-101 & \checkmark &  & 87.7 & 87.0 & 88.4 & 83.1 & -\\ 
        DTTT\cite{DTTT} & CVPR'24 & Mit-b1 & \checkmark &  & 87.2 & 85.8 & 88.5 & 78.8 & -\\  
        GFA\cite{GFA} & AAAI'24 & - & \checkmark &  & \underline{88.2} & \underline{87.4} & \underline{88.9} & 82.4 & \underline{74.7} \\ 
        GFA\cite{GFA} & AAAI'24 & ResNet-101 & \checkmark &  & 86.3 & 85.9 & 86.7 & 80.1 & 73.6 \\
        \hline
        \textbf{Ours(SMTC-Net)} & - & Mit-b1 & \checkmark &  & \textbf{89.2} & \textbf{88.2} & \textbf{90.2} & \textbf{86.4} & \textbf{76.0} \\ 
        \hline
    \end{tabularx}
    \label{tab:UVOS_method_comparison}
\end{table*}

\subsection{Quantitative Comparison with the State-of-The-Arts}

\subsubsection{Unsupervised Video Object Segmentation Results} To assess the effectiveness of SMTC-Net on the UVOS task, we fine-tuned and tested SMTC-Net on the DAVIS-16 and FBMS dataset. For the YouTube-Objects dataset, we directly leveraged the pre-trained model for testing. Results of SMTC-Net on UVOS metrics are presented in Table \ref{tab:UVOS_method_comparison}, with comparison to past UVOS works. In particular, for studies reporting multiple results, we only present those results derived using standard evaluation protocols and widely adopted backbones.

Previous motion-appearance UVOS approaches have predominantly focused on devising complex fusion mechanisms to enhance model performance. For instance, MATNet applied separate encoders for motion and appearance features and enhanced appearance representations by utilizing motion information. Similarly, RTNet proposed a reciprocal transformation module to refine motion and appearance features bidirectionally. However, neither MATNet nor RTNet could outperform TMO, which employed the same backbone yet discarded any merging modules, highlighting the limited improvements that fusion schemes alone can achieve. HFAN introduced Transformer-based architectures to UVOS, performing asymmetric mutual refinement and weighted fusion between motion and appearance features. However, according to experimental verification, the sophisticated alignment schemes only brought marginal improvement. SimulFlow incorporated motion-appearance interaction directly into the attention blocks of Segformer \cite{Segformer}, but still attained indistinctive effect advancement. 

These prior approaches ignored the fundamental motion-appearance relationship, which resulted in the marginal improvements. In contrast, the trunk-collateral architecture of SMTC-Net accounts for motion-appearance distinctions and commonalities simultaneously, which facilitates feature encoding without relying on complex fusion strategies. Additionally, some methods, such as \cite{DTTT}, incorporated additional inputs (e.g., depth maps or external saliency cues) to aid in segmentation. While effective to some extent, this fashion severely impedes the model's efficiency and incurs greater storage costs. SMTC-Net instead leverages the model’s intrinsic saliency, which promotes model performance while streamlining the overall process.

\begin{table*}[ht]
    \centering
    \caption{Quantitative results and comparative analysis on the YouTube-Objects dataset, presenting the mean region similarity \(\mathcal{J}\) for each category and the entire dataset. The top two performances are highlighted by \textbf{bold} and \underline{underlined} respectively}
    \renewcommand{\arraystretch}{1.25}
    \begin{tabularx}{\textwidth}{p{2cm}|>{\centering}X>{\centering}X>{\centering}X>{\centering}X>{\centering}X>{\centering}X>{\centering}X>{\centering}X>{\centering}X>{\centering}X| c}
        \toprule
        \textbf{Method} &\textbf{Aeroplane} & \textbf{Bird} & \textbf{Boat} & \textbf{Car} & \textbf{Cat} & \textbf{Cow} & \textbf{Dog} & \textbf{Horse} & \textbf{Motorbike} & \textbf{Train} & \textbf{Average} \\ 
        \midrule
        AGS\cite{AGS} & \underline{87.7} & 76.7 & 72.2 & 78.6 & 69.2 & 64.6 & 73.3 & 64.4 & 62.1 & 48.2 & 69.7 \\ 
        COSNet\cite{COSNet} & 81.1 & 75.7 & 71.3 & 77.6 & 66.5 & 69.8 & 76.8 & 67.4 & 67.7 & 46.8 & 70.5 \\ 
        AGNN\cite{AGNN} & 81.1 & 75.9 & 70.7 & 78.1 & 67.9 & 69.7 & 77.4 & 67.3 & \underline{68.3} & 47.8 & 70.8 \\ 
        WCS-Net\cite{WCS-Net} & 81.8 & 81.1 & 67.7 & 79.2 & 64.7 & 65.8 & 73.4 & 68.6 & \textbf{69.7} & 49.2 & 70.5 \\
        MATNet\cite{MATNet} & 72.9 & 77.5 & 66.9 & 79.0 & 73.7 & 67.4 & 75.9 & 63.2 & 62.6 & 51.0 & 69.0 \\ 
        RTNet\cite{RTNet} & 84.1 & 80.2 & 70.1 & 79.5 & 71.8 & 70.1 & 71.3 & 65.1 & 64.6 & 53.3 & 71.0 \\ 
        AMC-Net\cite{AMC-Net} & 78.9 & 80.9 & 67.4 & 82.0 & 69.0 & 69.6 & 75.8 & 63.0 & 63.4 & 57.8 & 71.1 \\ 
        HFAN\cite{HFAN} & 84.7 & 80.0 & 72.0 & 76.1 & 76.0 & 71.2 & 76.9 & \textbf{71.0} & 64.3 & 61.4 & 73.4 \\ 
        TMO\cite{TMO} & 85.7 & 80.0 & 70.1 & 78.0 & 73.6 & 70.3 & 76.8 & 66.2 & 58.6 & 47.0 & 71.5 \\
        HGPU\cite{HGPU} & \textbf{89.7} & 84.6 & 69.8 & 75.1 & 67.6 & \underline{71.7}&  75.8 & 69.5 & 65.1 & \underline{70.5} & 73.9 \\
        HCPN\cite{HCPN} & 84.5 & 79.6 & 67.3 & \textbf{87.8} & 74.1 & 71.2 & 76.5 & 66.2 & 65.8 & 59.7 & 73.3 \\ 
        DPA\cite{DPA} & 87.5 & \underline{85.6} & 70.1 & 77.7 & \underline{81.2} & 69.0 & \underline{81.8} & 61.9 & 62.1 & 51.3 & 73.7 \\
        GFA\cite{GFA} & 87.2 & 85.5 & \textbf{74.7} & 82.9 & 80.4 & \textbf{72.0} & 79.6 & \underline{67.8} & 61.3 & 55.8 & \underline{74.7}\\
       \textbf{Ours(SMTC-Net)} & 85.9 & \textbf{86.4} & \underline{74.3} & \underline{86.5} & \textbf{82.3} & 70.8 & \textbf{83.1} & 67.1 & 47.9 & \textbf{74.1} & \textbf{76.0} \\
       \bottomrule
    \end{tabularx}
    \label{tab:ytbobj_category}
\end{table*}

A series of outstanding experimental results demonstrate the superiority and effectiveness of our model. SMTC-Net achieves state-of-the-art performance across all three standard UVOS benchmarks. On the DAVIS-16 validation set, SMTC-Net attained 89.2\% on the \(\mathcal{J} \& \mathcal{F}\) metric, outperforming all previous models by a margin of at least 1\%. On the FBMS test set, our model recorded a \(\mathcal{J}\) score of 86.4\% , marking a 3.0\% improvement over the best recent results. Moreover, SMTC-Net outperformed all existing UVOS methods on the YouTube-Objects dataset, achieving a 1.3\% \(\mathcal{J}\) value improvement over the recent top-performing approach. As detailed in Table \ref{tab:ytbobj_category}, our model secured either the best or second-best scores in six out of all ten categories in the YouTube-Objects dataset, underscoring the exceptional efficacy of SMTC-Net.

\begin{table*}[h]
    \centering
    \caption{Quantitative results and comparative analysis on the VSOD benchmark: DAVIS-16, DAVSOD, ViSal and FBMS. In the table, $\uparrow$ indicates that higher values correspond to better performance, whereas $\downarrow$ denotes the opposite. * signifies the results were obtained by executing the publicly available code provided in the respective paper. The top two performances are highlighted by \textbf{bold} and \underline{underline} respectively.}
    \renewcommand{\arraystretch}{1.4}
    \begin{tabularx}{\textwidth}{p{2.1cm}|p{0.7cm}XXX|p{0.7cm}XXX|p{0.7cm}XXX|p{0.7cm}XXX} 
        \hline
        \multirow{2}{*}{\textbf{Method}} & \multicolumn{4}{c|}{\textbf{DAVIS-16}} & \multicolumn{4}{c|}{\textbf{DAVSOD}} & \multicolumn{4}{c|}{\textbf{ViSal}} & \multicolumn{4}{c}{\textbf{FBMS}} \\ 
        \cline{2-17}
    
        & $MAE\downarrow$ & $F_m\uparrow$ & $E_m\uparrow$ & $S_m\uparrow$ & $MAE\downarrow$ & $F_m\uparrow$ & $E_m\uparrow$ & $S_m\uparrow$ & $MAE\downarrow$ & $F_m\uparrow$ & $E_m\uparrow$ & $S_m\uparrow$ & $MAE\downarrow$ & $F_m\uparrow$ & $E_m\uparrow$ & $S_m\uparrow$ \\
        \hline

        AD-Net\cite{AD-Net} & 0.044 & 0.808 & - & - & - & - & - & - & 0.030 & 0.904 &  - & - & 0.064 & 0.812 & - & -\\
        MATNet*\cite{MATNet} & 0.048 & 0.752 & 0.890 & 0.776 & 0.098 & 0.628 & 0.789 & 0.707 & 0.041 & 0.891 & 0.967 & 0.863 & 0.091 & 0.751 & 0.852 & 0.760 \\
        DFNet\cite{DFNet} & 0.018 & 0.899 & - & - & - & - & - & - &0.017 & 0.927 & - & - & 0.054 & 0.833 & - & -\\
        RTNet*\cite{RTNet} & 0.012 & 0.928 & 0.978 & 0.933 & 0.068 & 0.647 & 0.782 & 0.743 & 0.019 & 0.938 & 0.975 & 0.936 & 0.057 & 0.845 & 0.892 & 0.855 \\ 
        FSNet\cite{FSNet} & 0.020 & 0.907 & 0.970 & 0.920 & 0.072 & 0.685 & 0.825 & 0.773 & - & - & - & - & 0.041 & 0.888 & 0.935 &0.890 \\
        TransportNet\cite{TransportNet} & 0.013 & 0.928 & - & - &- & - & - & - & \textbf{0.012} & \underline{0.953} & - & - & 0.045 & 0.885 & - & - \\
        HFAN*\cite{HFAN} & 0.014 & 0.930 & \underline{0.984} & \underline{0.939} & 0.078 & 0.656 & 0.795 & 0.763 & 0.029 & 0.860 & 0.928 & 0.891 & 0.065 & 0.794 & 0.877 & 0.818 \\
        HCPN*\cite{HCPN} & 0.017 & 0.923 & 0.980 & 0.932 & 0.072 & 0.684 & 0.818 & 0.774 & 0.016 & 0.942 & 0.986 & 0.945 & 0.060 & 0.850 & 0.903 & 0.851 \\
        TMO*\cite{TMO} & 0.013 & 0.925 & 0.982 & 0.936 & \underline{0.062} & \underline{0.731} & \underline{0.849} & \underline{0.805} & \underline{0.013} & 0.951 & \underline{0.989} & \underline{0.951} & 0.036 & 0.887 & \underline{0.933} & 0.893 \\
        OAST\cite{OAST} & \underline{0.011} & 0.926 & - & 0.935 & 0.070 & 0.712 & - & 0.786 & - & - & - & - & \textbf{0.025} & \underline{0.919} & - & \underline{0.917}\\
        SimulFlow\cite{SimulFlow} & \textbf{0.009} & \underline{0.936} & - & 0.937 & 0.069 & 0.722 & - & 0.771 & \textbf{0.012} & 0.943 & - & 0.946 & - & - & - & -\\
        \textbf{Ours(SMTC-Net)} & \textbf{0.009} & \textbf{0.944} & \textbf{0.989} & \textbf{0.949} & \textbf{0.059} & \textbf{0.757} & \textbf{0.862} & \textbf{0.812} & \underline{0.013} & \textbf{0.962} & \textbf{0.992} & \textbf{0.961} & \underline{0.029} & \textbf{0.939} & \textbf{0.970} & \textbf{0.923} \\
        
        \hline
    \end{tabularx}
    \label{tab:VSOD_method_comparison}
\end{table*}

\subsubsection{Video Salient Object Detection Results}
Given the shared objective of detecting visually salient objects in a video sequence between UVOS and VSOD tasks, we also evaluate the performance of SMTC-Net on four VSOD benchmarks: DAVIS-16, FBMS, ViSal, and DAVSOD. To demonstrate the effectiveness of our model on these benchmarks, we employ four evaluation metrics and compare SMTC-Net's results against those of previous UVOS methods, which either reported relevant results or provided publicly available code. The comparative results are presented in Table \ref{tab:VSOD_method_comparison}.

For previous UVOS algorithms, the reported results were either directly extracted from their respective papers or obtained by running the original code and weights provided by the authors. When executing the code, we strictly adhered to the guidelines outlined in their publications and the instructions within the code repositories. As shown in Table \ref{tab:VSOD_method_comparison}, prior algorithms demonstrate varying strengths across different datasets. For instance, TransportNet achieved the best performance on ViSal, yet its results on FBMS were less competitive. Similarly, HFAN delivered excellent performance on DAVIS-16, but its outcomes on the other three datasets were less satisfactory. TMO demonstrated favorable results on DAVSOD and ViSal, but struggled to compete with more recent approaches on DAVIS-16 and FBMS.

Our SMTC-Net consistently outperforms existing UVOS algorithms, achieving state-of-the-art performance across all four VSOD benchmarks. Of the total 16 evaluation metrics used across these datasets, SMTC-Net secured the best on 14 metrics and ranked second on the remaining one. Notably, SMTC-Net achieved the highest scores for the $F_m$, $E_m$, and $S_m$ metrics across all benchmarks with substantial improvements over prior methods, while attaining either the best or second-best performance on the $MAE$ metric. These outstanding quantitative results underscore the robustness and effectiveness of SMTC-Net in identifying salient objects in video sequences, further demonstrating its superior capability in addressing VSOD and UVOS tasks.

\begin{figure*}[htbp]
    \centering
    \setlength{\tabcolsep}{1pt}
    \begin{tabular}{ccccccc}
        \rotatebox{90}{\hspace{0.1cm} \textbf{Case1}} & 
        \includegraphics[width=0.15\textwidth]{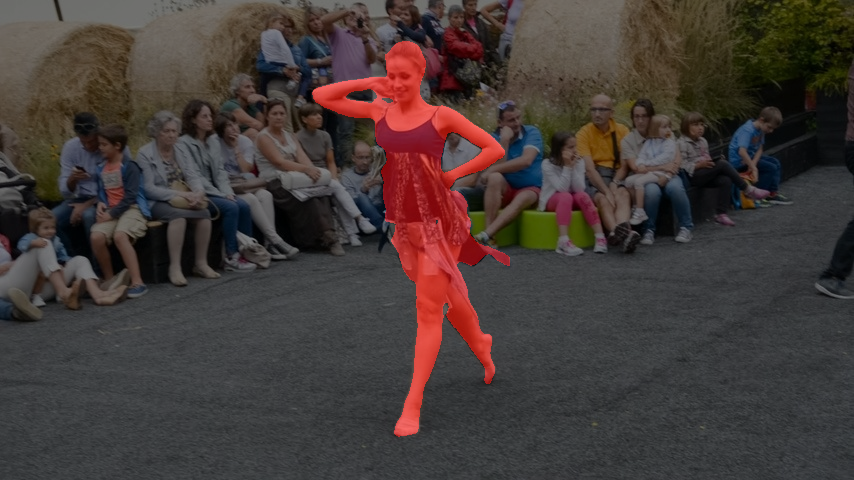} & 
        \includegraphics[width=0.15\textwidth]{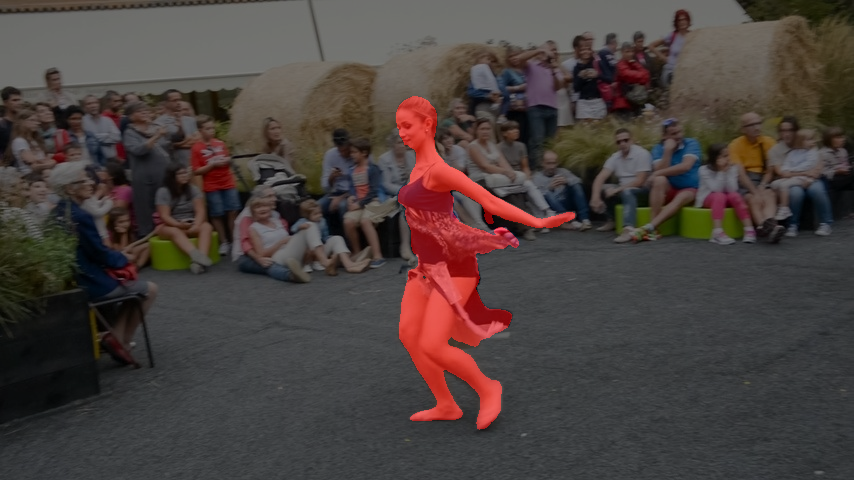} & 
        \includegraphics[width=0.15\textwidth]{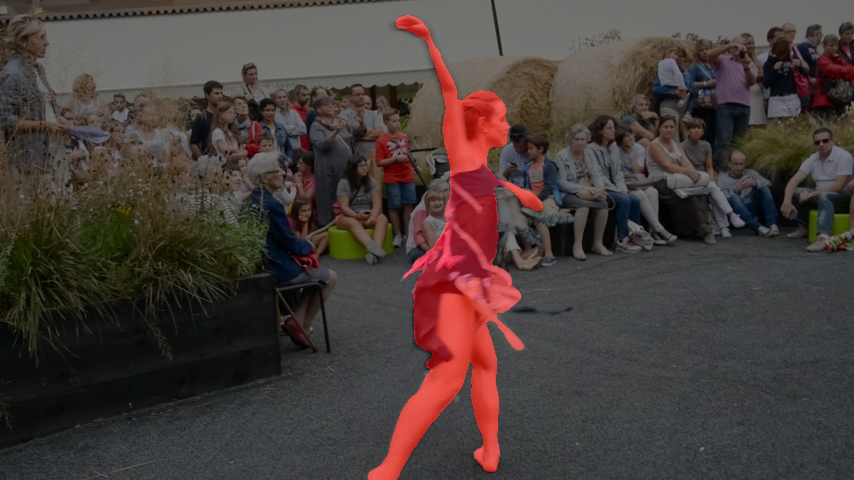} &
        \includegraphics[width=0.15\textwidth]{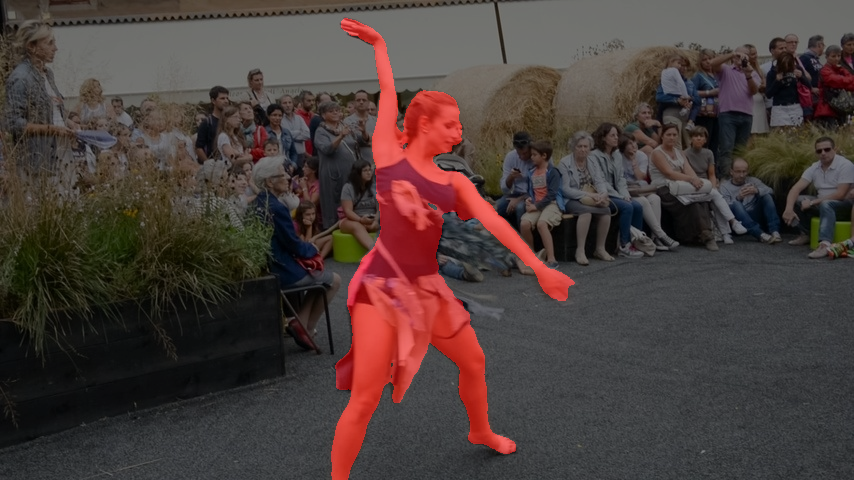} &
        \includegraphics[width=0.15\textwidth]{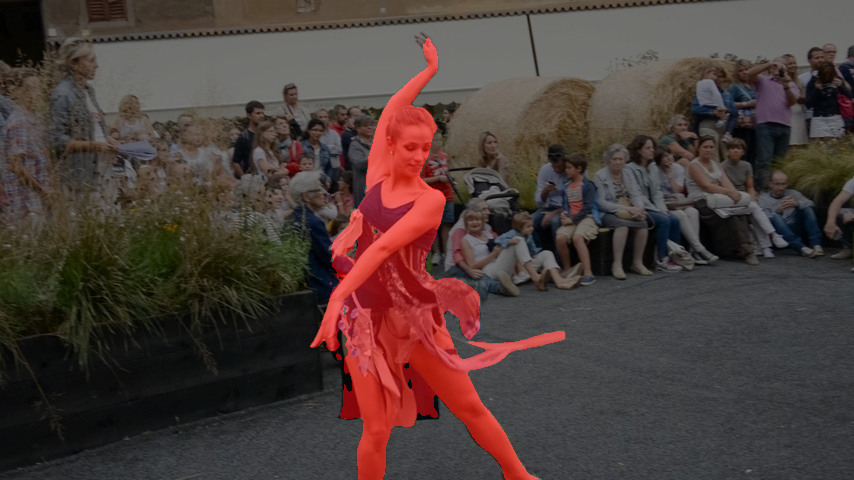} &
        \includegraphics[width=0.15\textwidth]{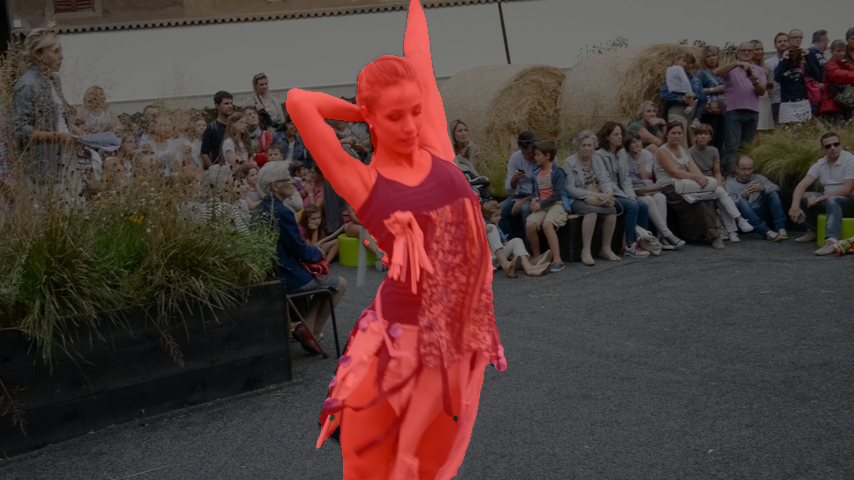} 
        \\ 
        \rotatebox{90}{\hspace{0.1cm} \textbf{Case2}} & 
        \includegraphics[width=0.15\textwidth]{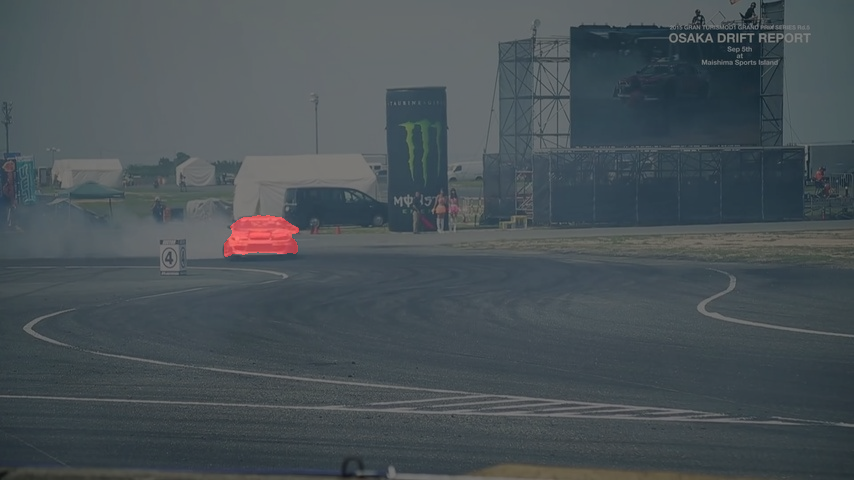} & 
        \includegraphics[width=0.15\textwidth]{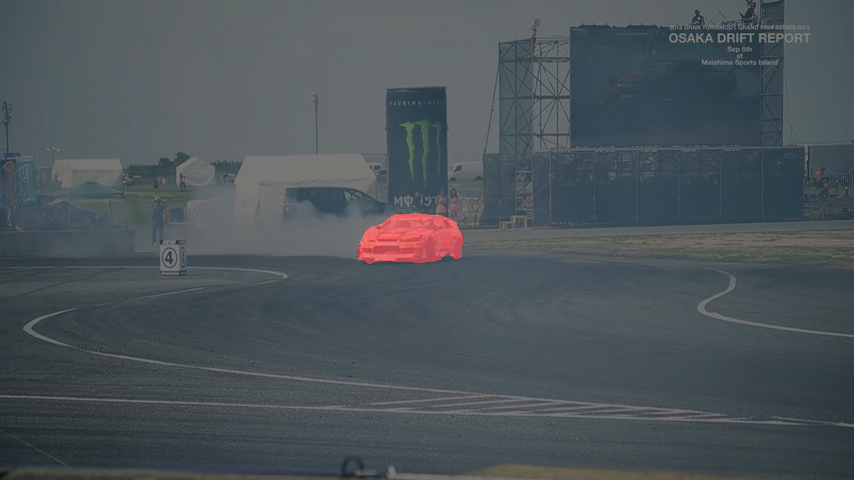} & 
        \includegraphics[width=0.15\textwidth]{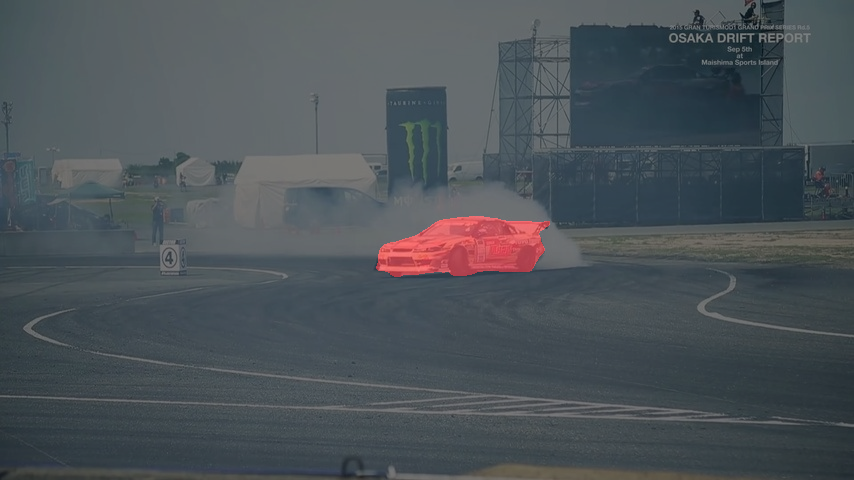} &
        \includegraphics[width=0.15\textwidth]{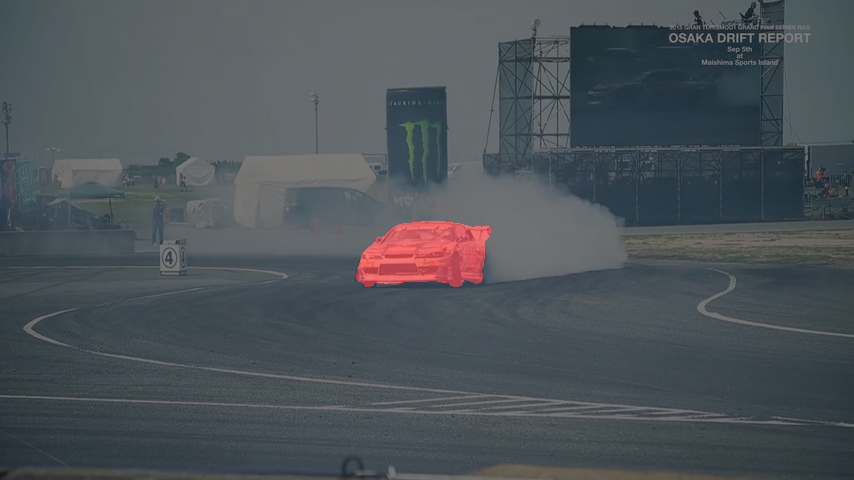} &
        \includegraphics[width=0.15\textwidth]{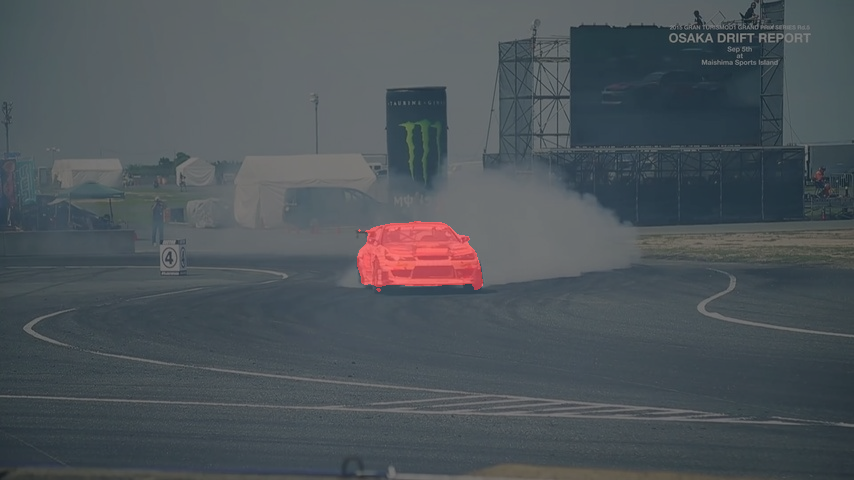} &
        \includegraphics[width=0.15\textwidth]{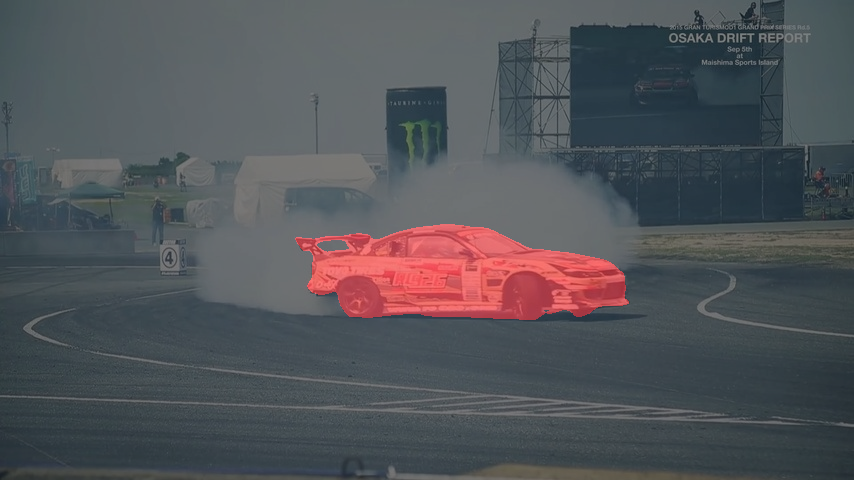} 
        \\  
        \rotatebox{90}{\hspace{0.1cm} \textbf{Case3}} & 
        \includegraphics[width=0.15\textwidth]{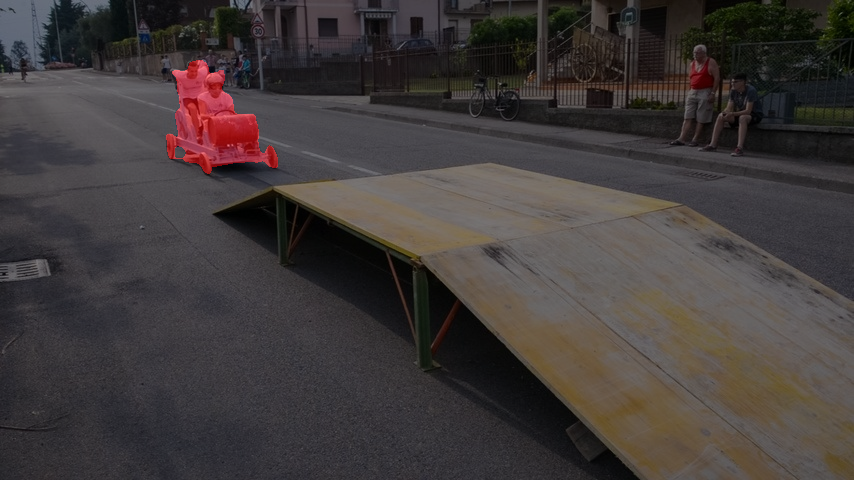} & 
        \includegraphics[width=0.15\textwidth]{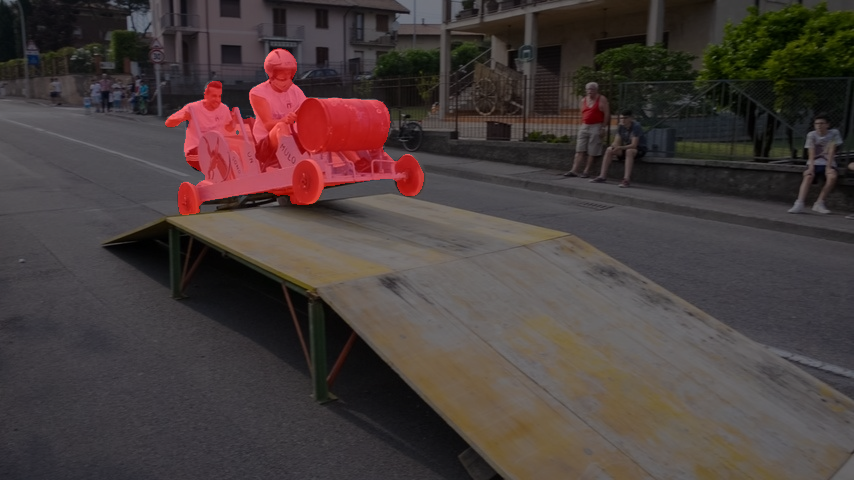} & 
        \includegraphics[width=0.15\textwidth]{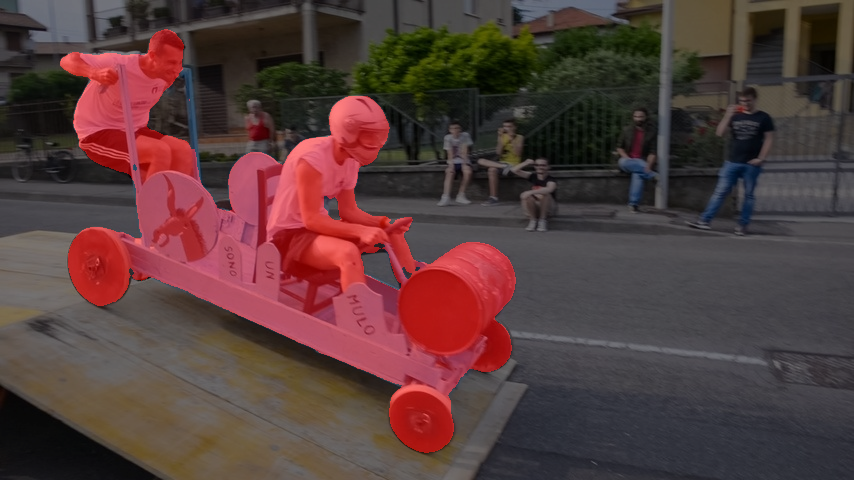} &
        \includegraphics[width=0.15\textwidth]{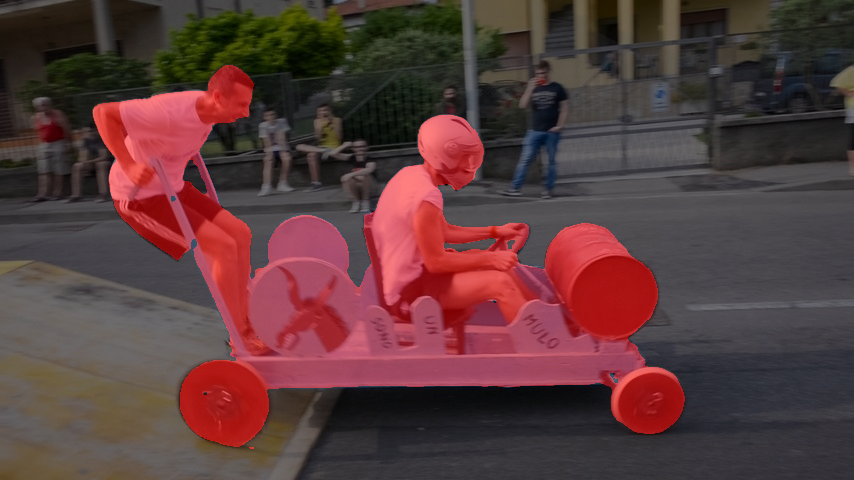} &
        \includegraphics[width=0.15\textwidth]{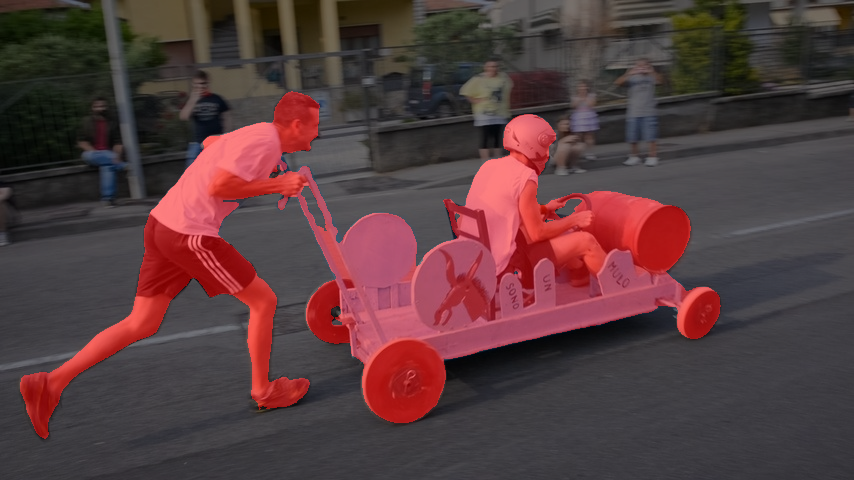} &
        \includegraphics[width=0.15\textwidth]{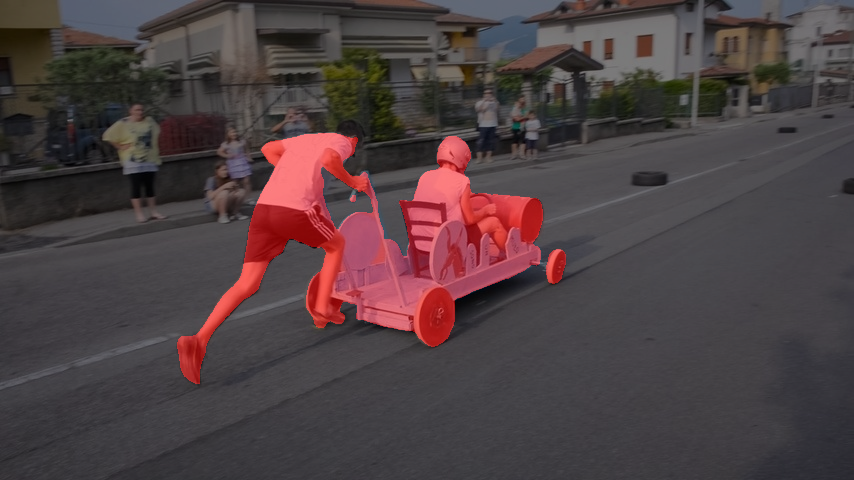} 
        \\ 
        \rotatebox{90}{\hspace{0.3cm} \textbf{Case4}} & 
        \includegraphics[width=0.15\textwidth]{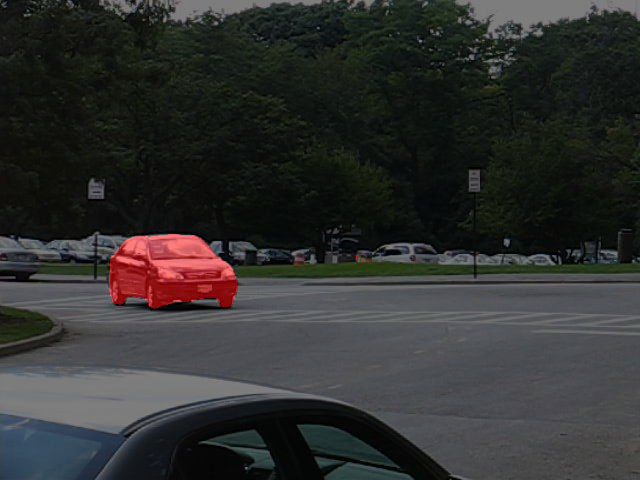} & 
        \includegraphics[width=0.15\textwidth]{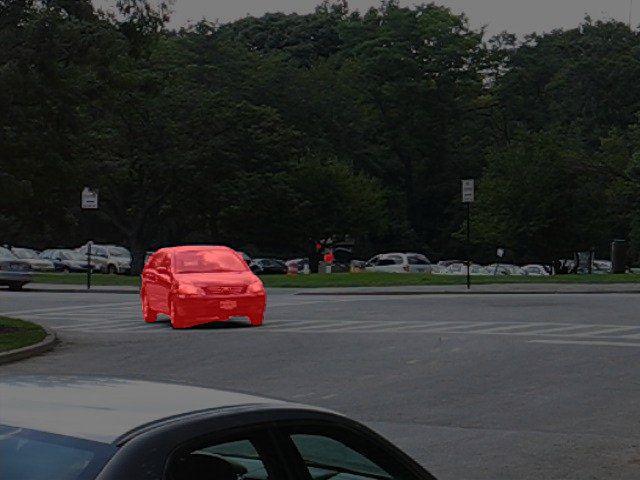} & 
        \includegraphics[width=0.15\textwidth]{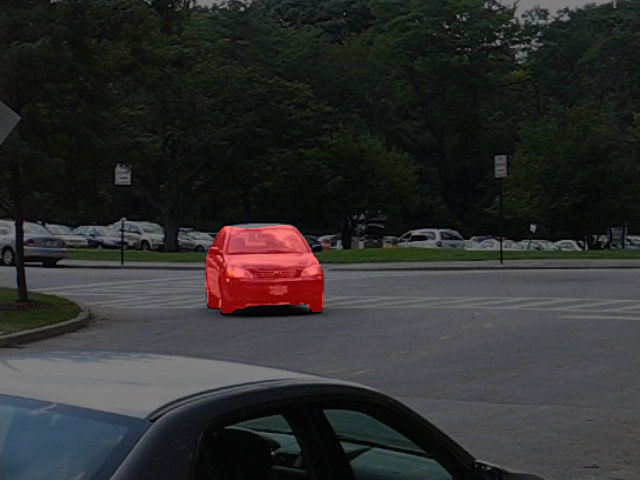} &
        \includegraphics[width=0.15\textwidth]{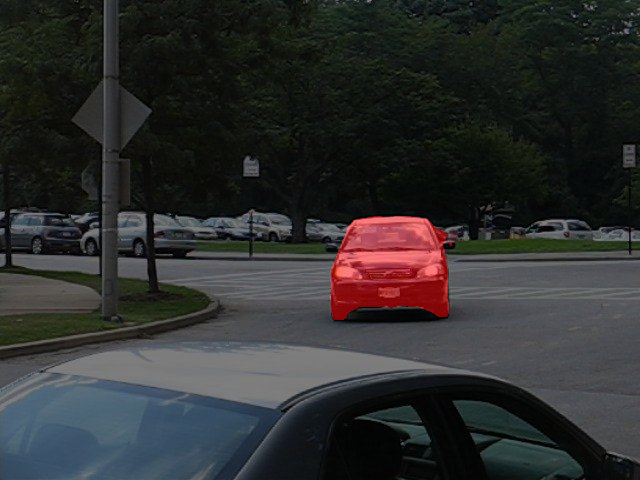} &
        \includegraphics[width=0.15\textwidth]{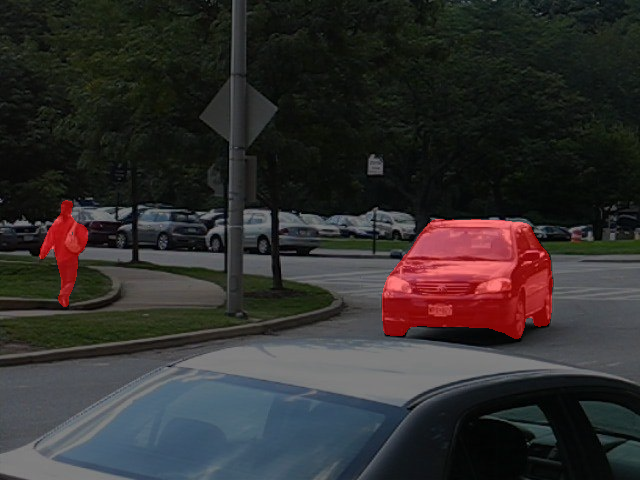} &
        \includegraphics[width=0.15\textwidth]{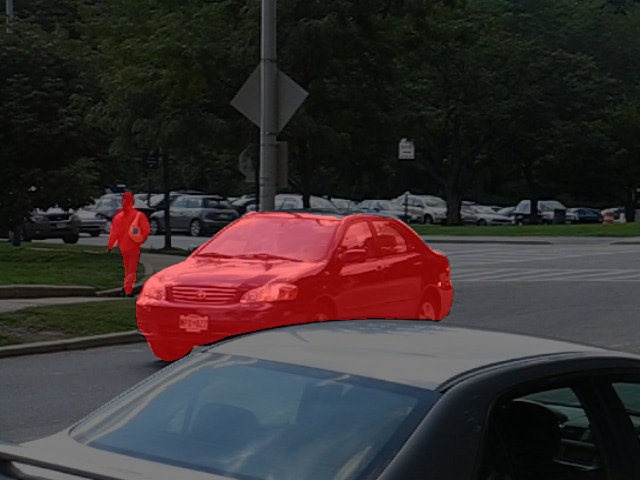} 
        \\ 
        \rotatebox{90}{\hspace{0.3cm} \textbf{Case5}} & 
        \includegraphics[width=0.15\textwidth]{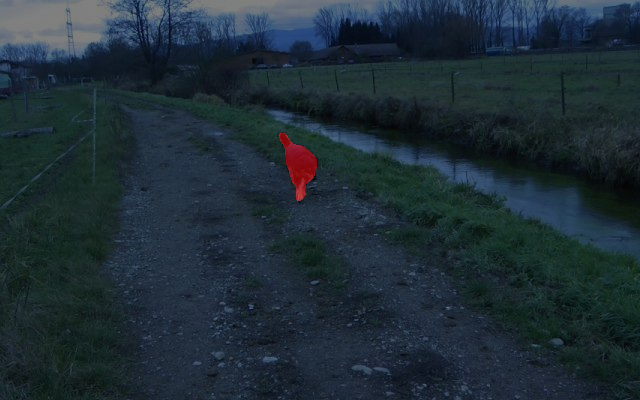} & 
        \includegraphics[width=0.15\textwidth]{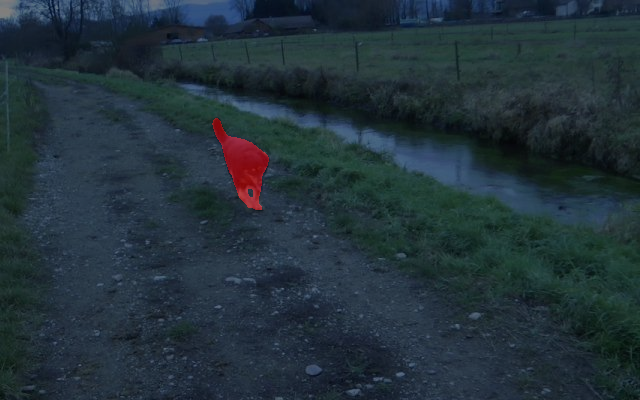} & 
        \includegraphics[width=0.15\textwidth]{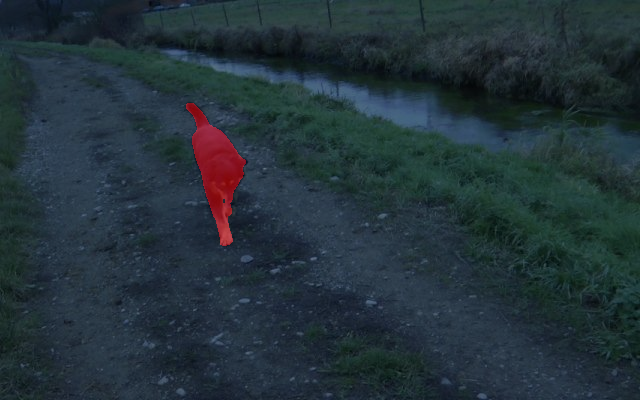} &
        \includegraphics[width=0.15\textwidth]{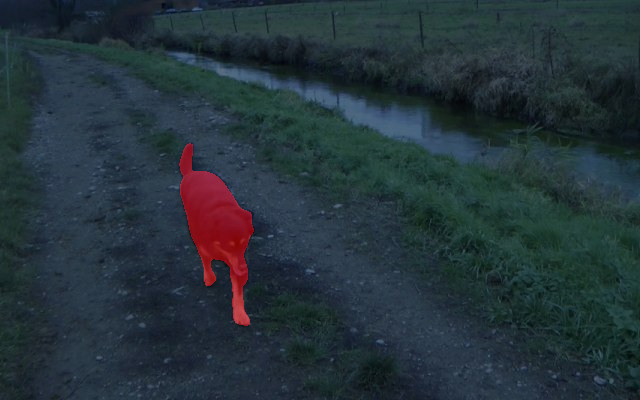} &
        \includegraphics[width=0.15\textwidth]{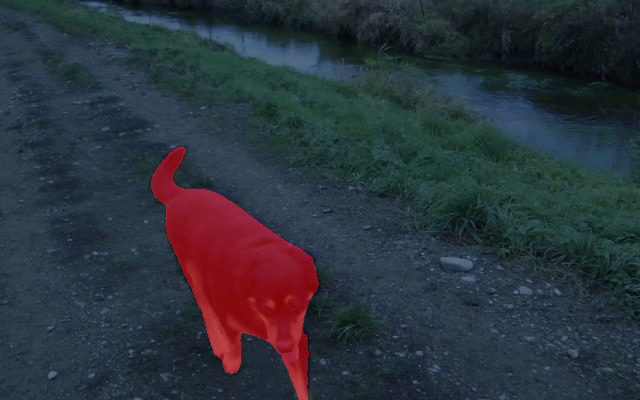} &
        \includegraphics[width=0.15\textwidth]{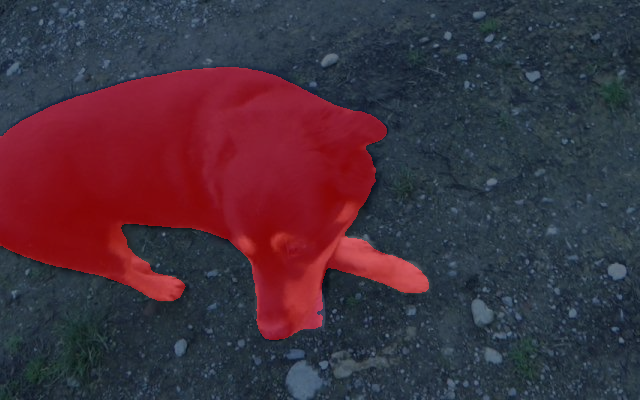} 
        \\ 
        \rotatebox{90}{\hspace{0.1cm} \textbf{Case6}} & 
        \includegraphics[width=0.15\textwidth]{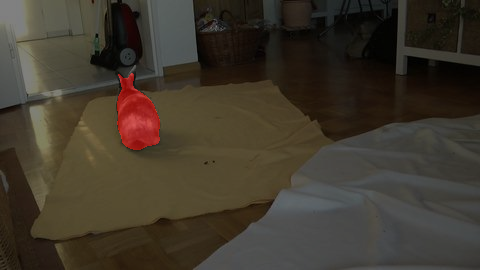} & 
        \includegraphics[width=0.15\textwidth]{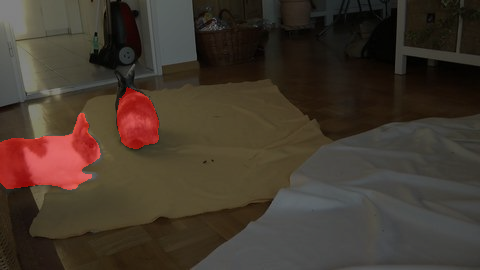} & 
        \includegraphics[width=0.15\textwidth]{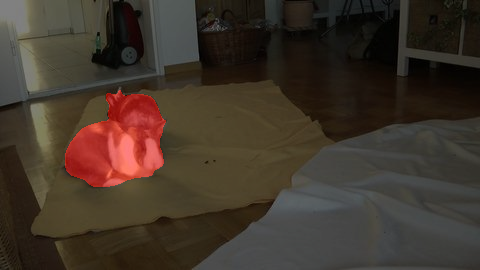} &
        \includegraphics[width=0.15\textwidth]{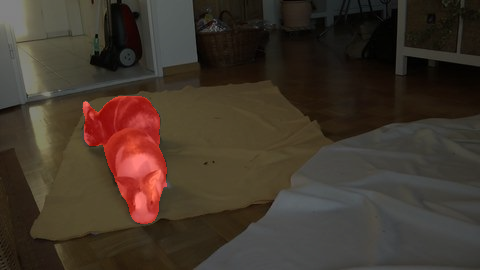} &
        \includegraphics[width=0.15\textwidth]{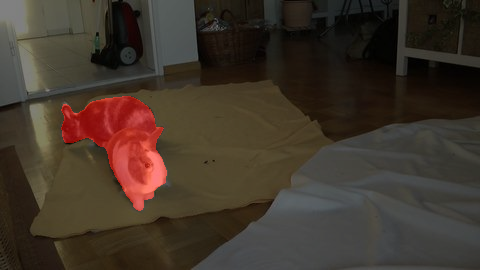} &
        \includegraphics[width=0.15\textwidth]{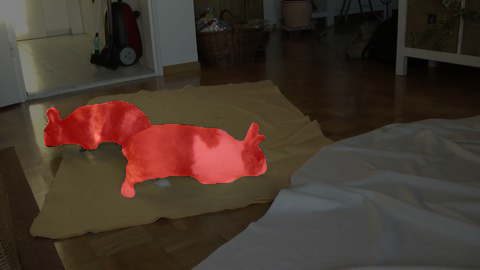} 
        \\ 
        \rotatebox{90}{\hspace{0.1cm} \textbf{Case7}} & 
        \includegraphics[width=0.15\textwidth]{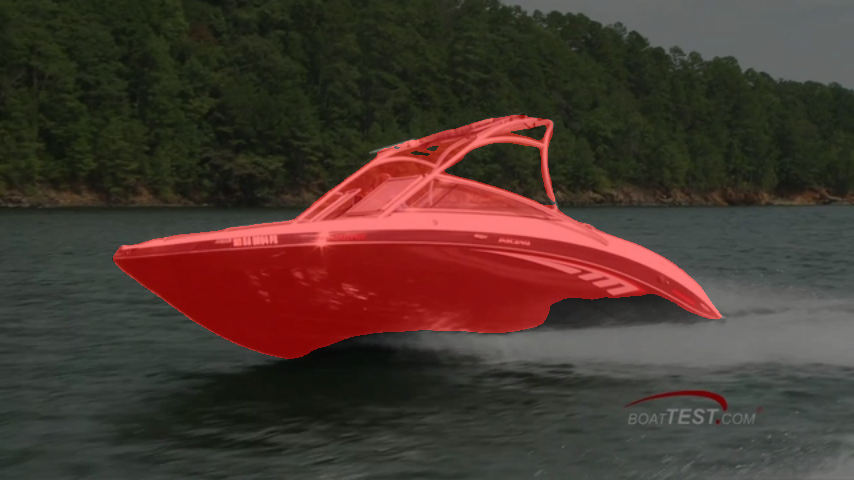} & 
        \includegraphics[width=0.15\textwidth]{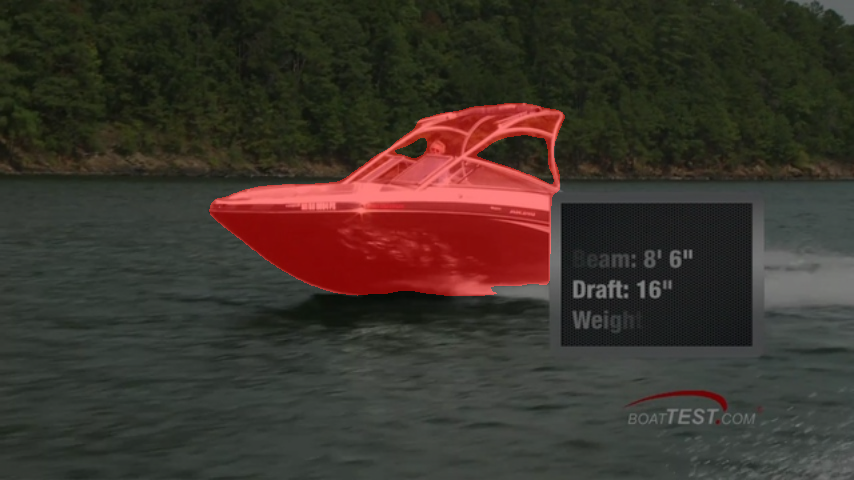} & 
        \includegraphics[width=0.15\textwidth]{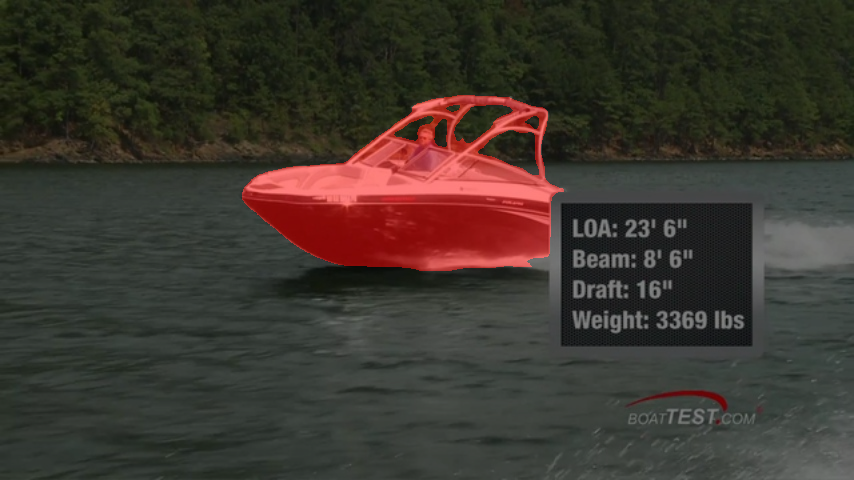} &
        \includegraphics[width=0.15\textwidth]{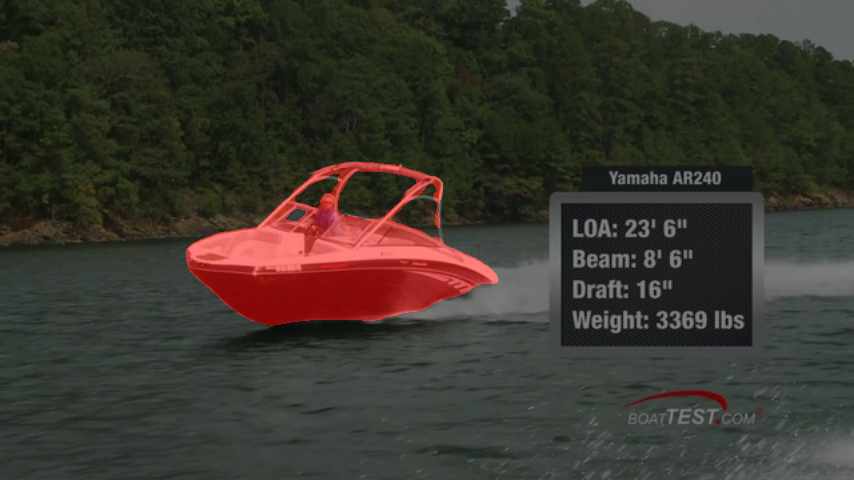} &
        \includegraphics[width=0.15\textwidth]{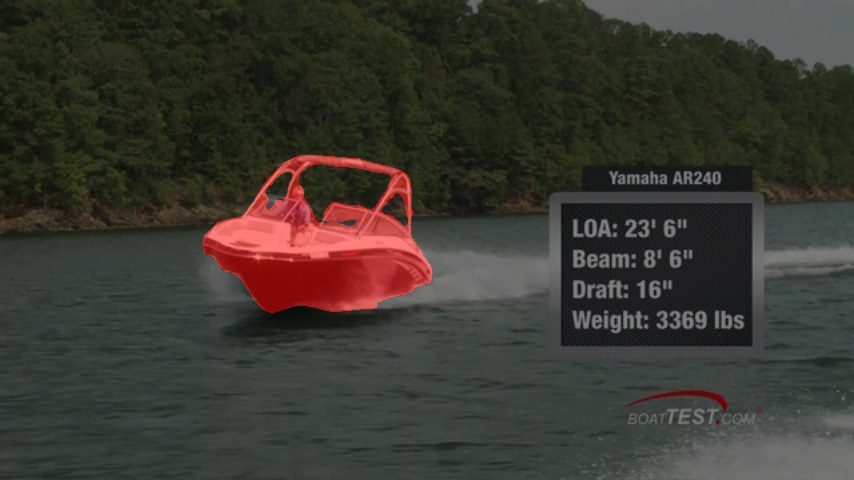} &
        \includegraphics[width=0.15\textwidth]{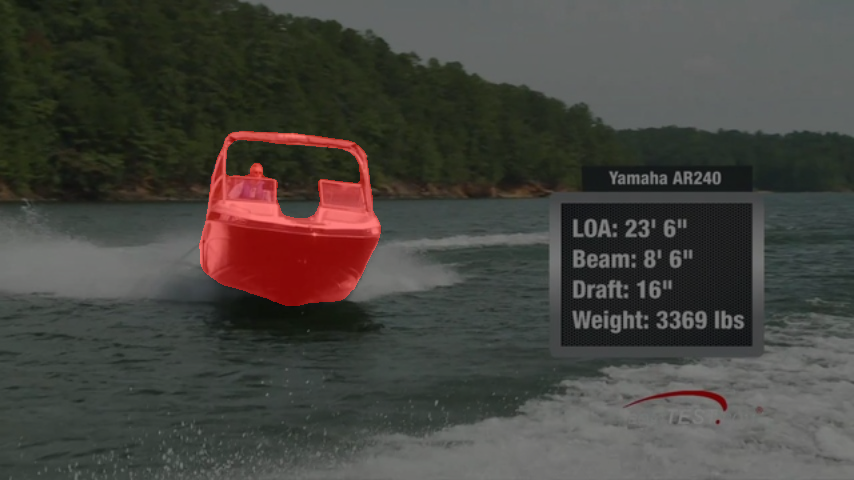} 
        \\ 
        \rotatebox{90}{\hspace{0.1cm} \textbf{Case8}} & 
        \includegraphics[width=0.15\textwidth]{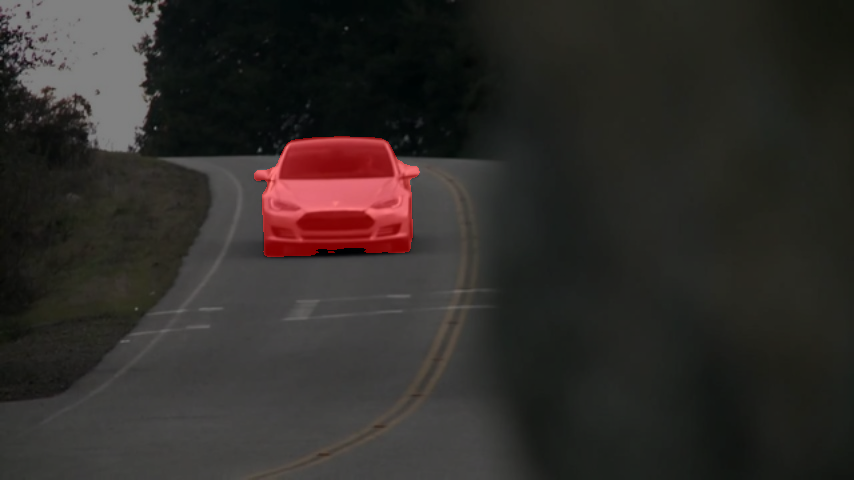} & 
        \includegraphics[width=0.15\textwidth]{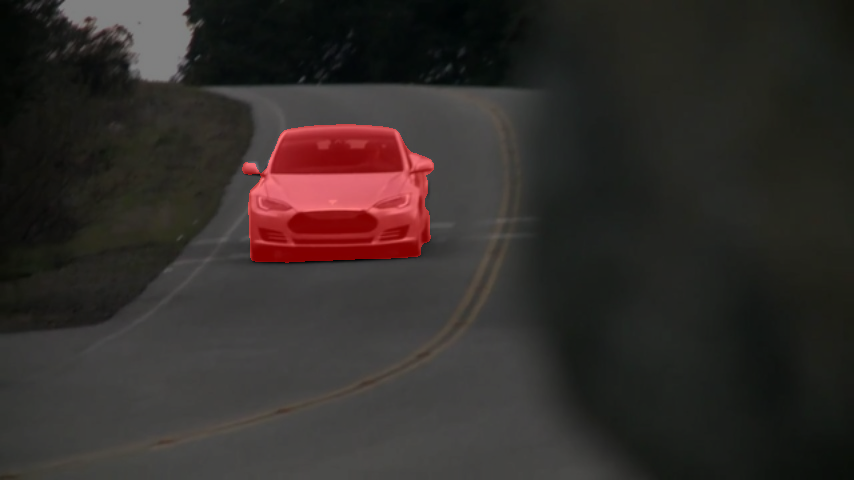} & 
        \includegraphics[width=0.15\textwidth]{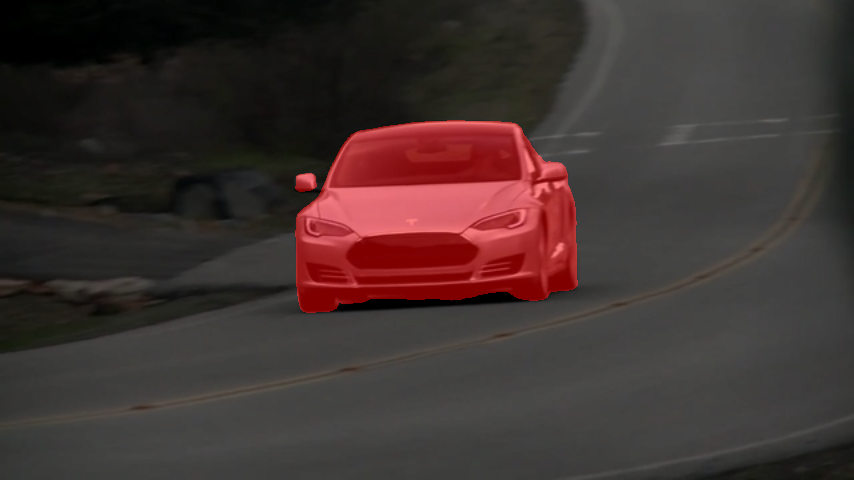} &
        \includegraphics[width=0.15\textwidth]{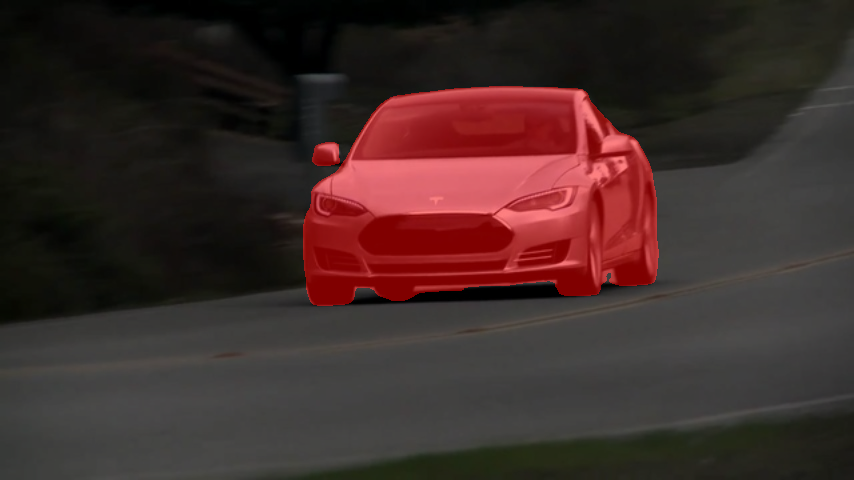} &
        \includegraphics[width=0.15\textwidth]{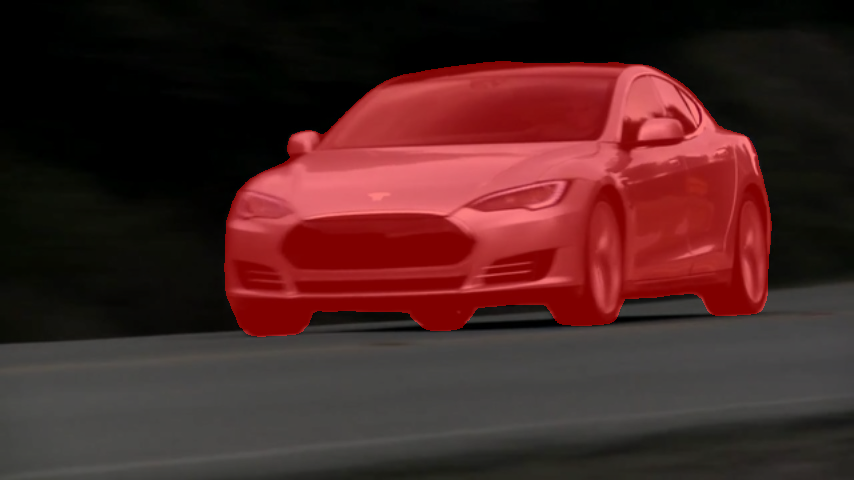} &
        \includegraphics[width=0.15\textwidth]{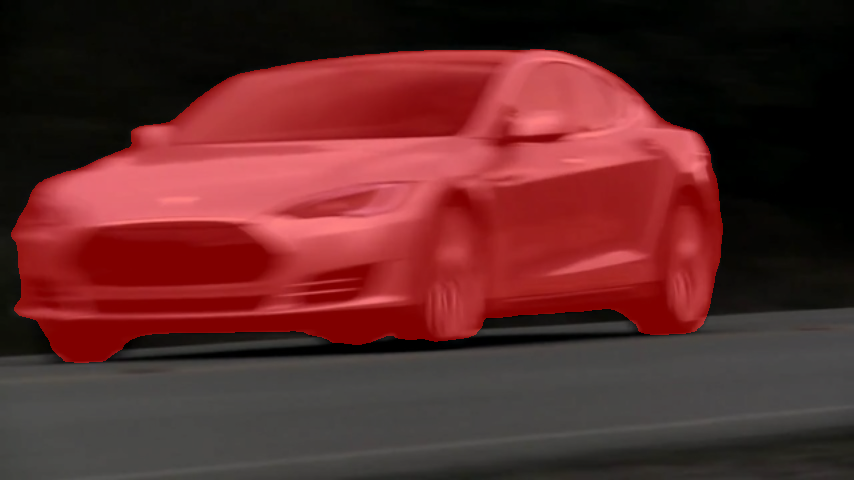} 
        \\  
        \rotatebox{90}{\hspace{0.1cm} \textbf{Case9}} & 
        \includegraphics[width=0.15\textwidth]{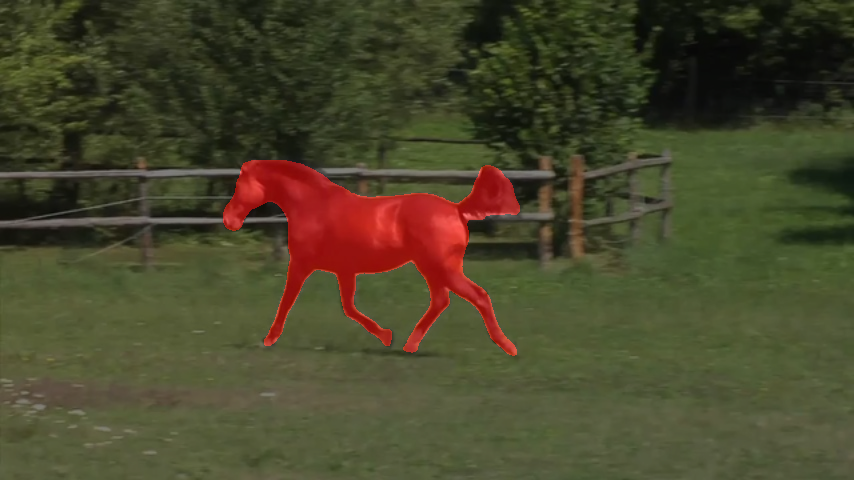} & 
        \includegraphics[width=0.15\textwidth]{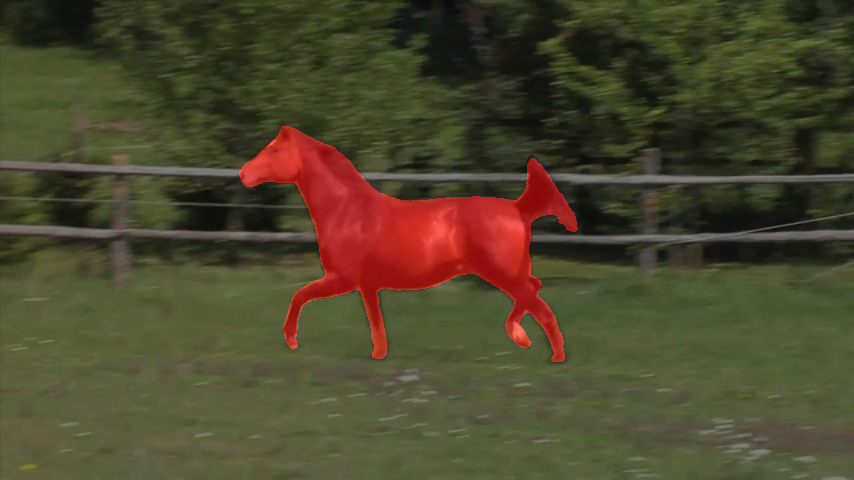} & 
        \includegraphics[width=0.15\textwidth]{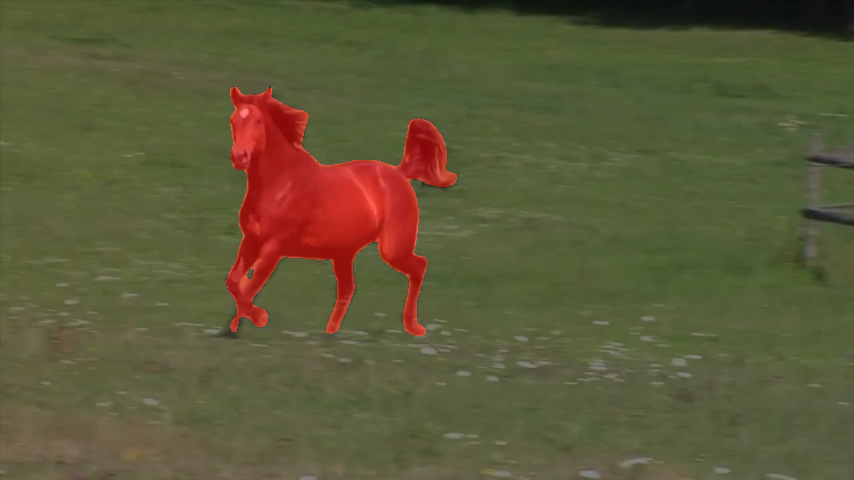} &
        \includegraphics[width=0.15\textwidth]{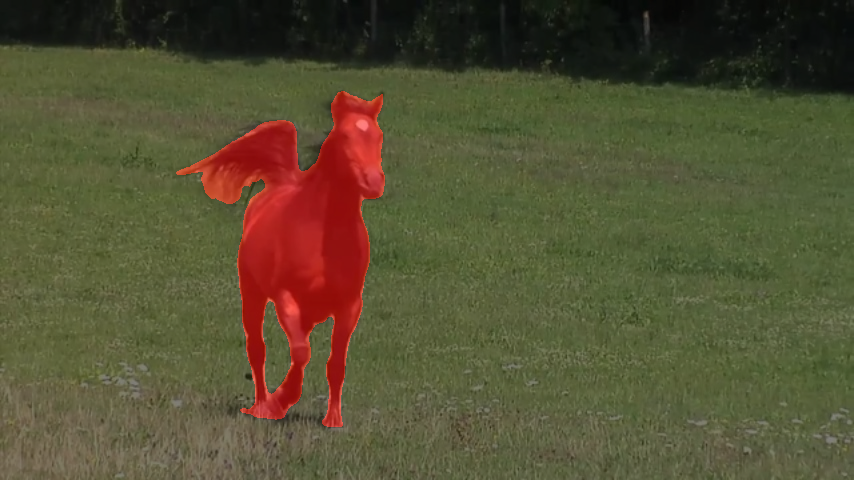} &
        \includegraphics[width=0.15\textwidth]{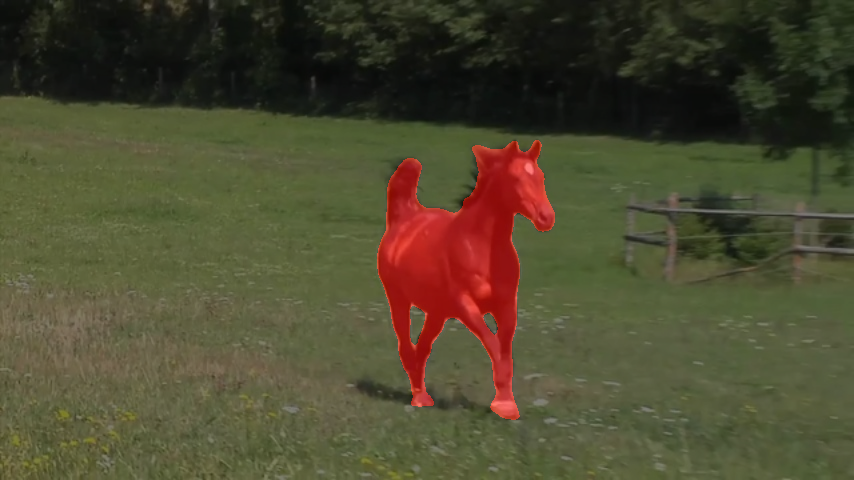} &
        \includegraphics[width=0.15\textwidth]{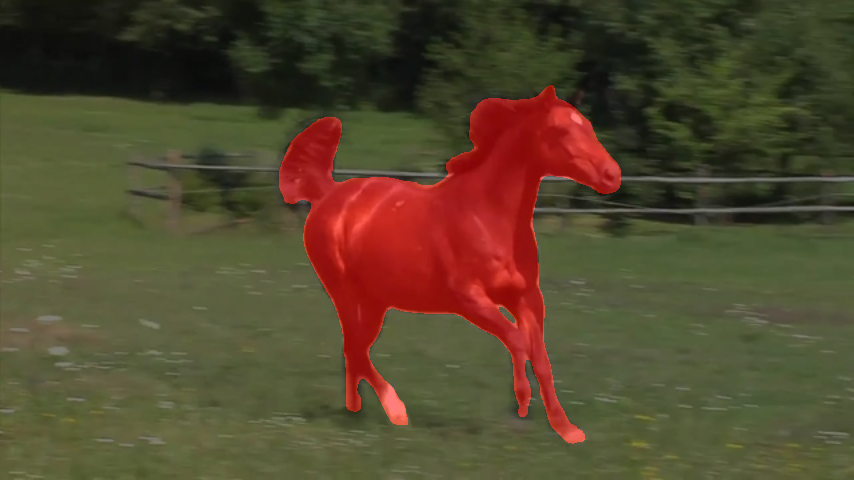} 
        \\ 
    \end{tabular}
    \caption{Qualitative demonstration of SMTC-Net's performance on diverse complex scenes from the DAVIS-16, FBMS, and YouTube-Objects datasets. The selected scenes include various challenging situations for the UVOS task, such as rapid object motion, co-moving background elements, multiple targets, intricate details, motion blur, and so on. Case 1-3: \textit{dance-twirl, drift-chicane, soapbox} from DAVIS-16; Case 4-6:  \textit{dogs02, cars4, rabbits02} from FBMS; Case 7-9:  \textit{boat0012, car0009, horse0011} from YouTube-Obects.}
    \label{Qualitative results UVOS}
\end{figure*}

\subsection{Qualitative Results}
\subsubsection{Unsupervised Video Object Segmentation} We qualitatively demonstrate the visual effectiveness of SMTC-Net on the UVOS task in Fig. \ref{Qualitative results UVOS}. Nine representative scenes were selected from three widely used UVOS benchmarks: DAVIS-16, FBMS, and YouTube-Objects. These videos cover a diverse range of challenging UVOS scenarios, including rapid object motion, co-moving background elements, multiple targets, intricate details, motion blur, etc.

In the first, third, and ninth cases, objects such as characters, vehicles, and horses exhibit complex, dynamic movements with numerous fine-grained details, including limbs or clothing. Our model consistently delivers accurate segmentation of the primary objects, capturing even subtle features with precision.

Some videos involve objects undergoing substantial visual changes in size due to shifting positions. For example, in the second, fifth, and eighth cases, the target objects transition from appearing small to large as their distance from the camera varies. Despite these volume transitions, SMTC-Net maintains highly accurate segmentation, demonstrating its adaptability to such challenges.

Furthermore, in the second and seventh cases, the presence of water ripples and smoke introduced by the movement of the primary objects is a common source of interference in UVOS tasks. These dynamic elements, which often move in sync with the background, can compromise the performance of optical flow in distinguishing foreground targets. Nevertheless, SMTC-Net effectively isolates the foreground objects from the background, preserving precise contours.

Additionally, SMTC-Net successfully handles a variety of other complex scenarios, such as multiple targets, motion blur, and the emergence of new objects, as shown in selected cases. These results highlight the robustness and exceptional segmentation capabilities of SMTC-Net, even in highly challenging conditions.

\begin{figure*}[htbp]
    \centering
    \setlength{\tabcolsep}{1pt}
    \begin{tabular}{cccccccccc}
         &  & \textbf{Case 1} &  &  & \textbf{Case 2} &  &  & \textbf{Case 3} &  \\
        \rotatebox{90}{\textbf{Frames}} &
        \includegraphics[width=0.1\textwidth, height=1.3cm]{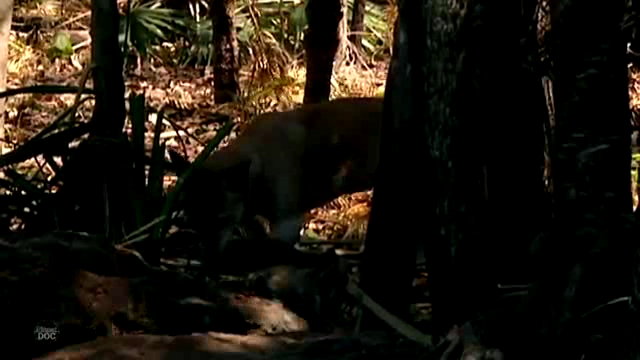} &
        \includegraphics[width=0.1\textwidth, height=1.3cm]{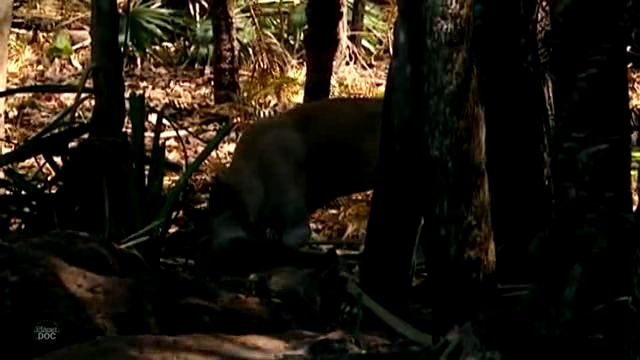} &
        \includegraphics[width=0.1\textwidth, height=1.3cm]{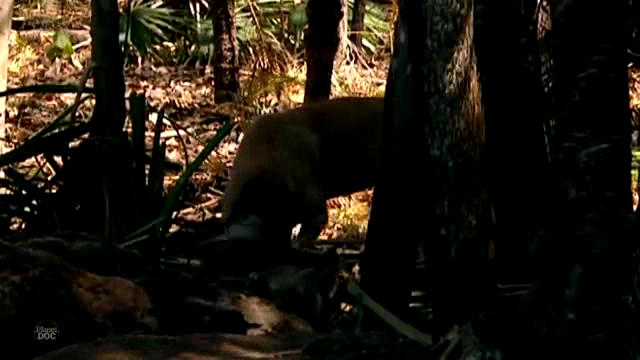} &
        \includegraphics[width=0.1\textwidth, height=1.3cm]{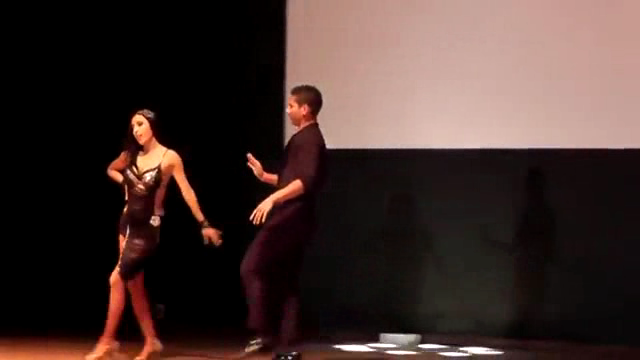} &
        \includegraphics[width=0.1\textwidth, height=1.3cm]{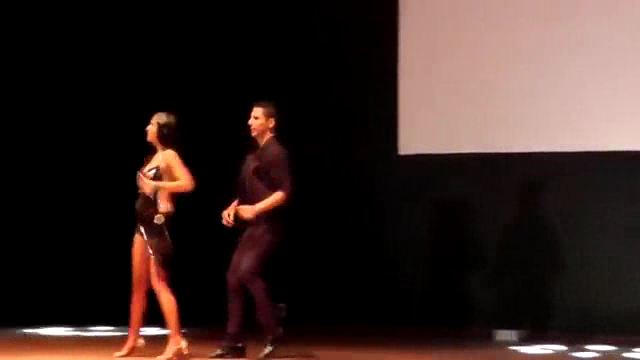} &
        \includegraphics[width=0.1\textwidth, height=1.3cm]{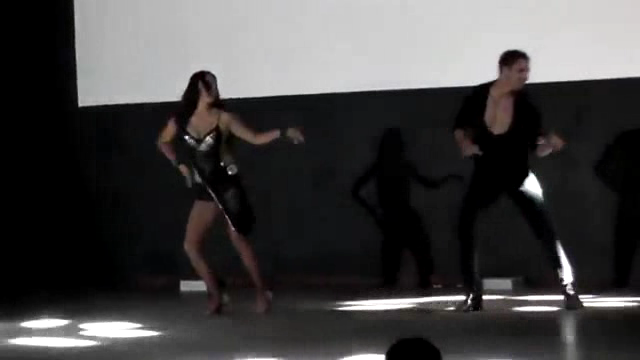}& 
        \includegraphics[width=0.1\textwidth, height=1.3cm]{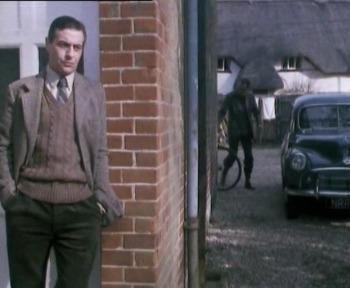} & 
        \includegraphics[width=0.1\textwidth, height=1.3cm]{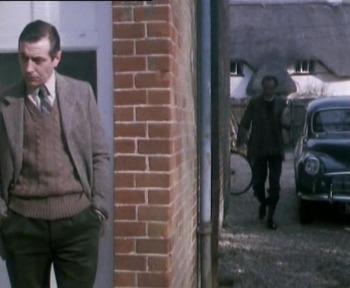} & 
        \includegraphics[width=0.1\textwidth, height=1.3cm]{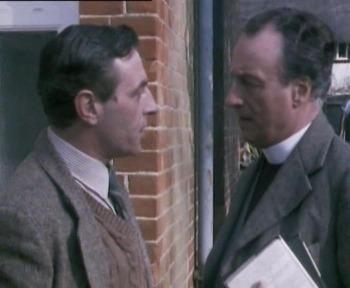}  \\

        \rotatebox{90}{\hspace{0.2cm} \textbf{GT}} &
        \includegraphics[width=0.1\textwidth, height=1.3cm]{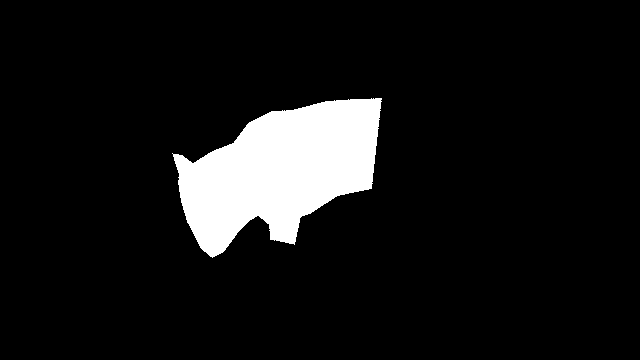} &
        \includegraphics[width=0.1\textwidth, height=1.3cm]{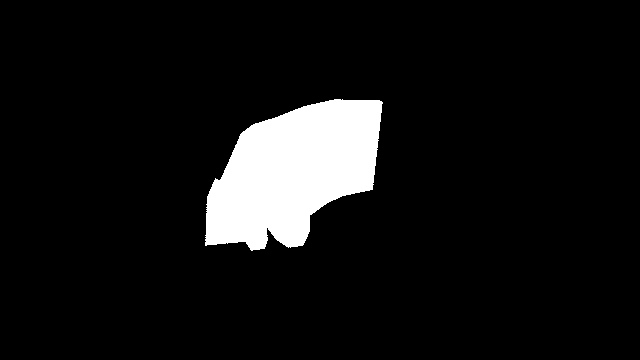} &
        \includegraphics[width=0.1\textwidth, height=1.3cm]{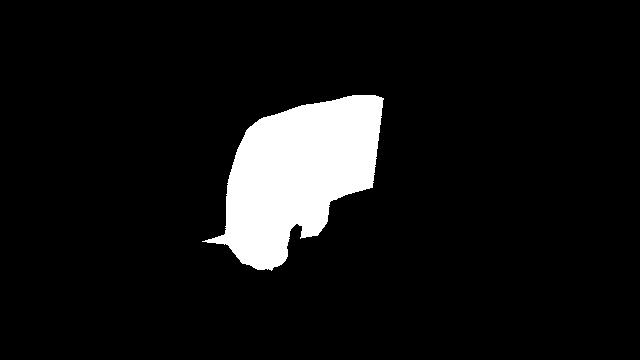} &
        \includegraphics[width=0.1\textwidth, height=1.3cm]{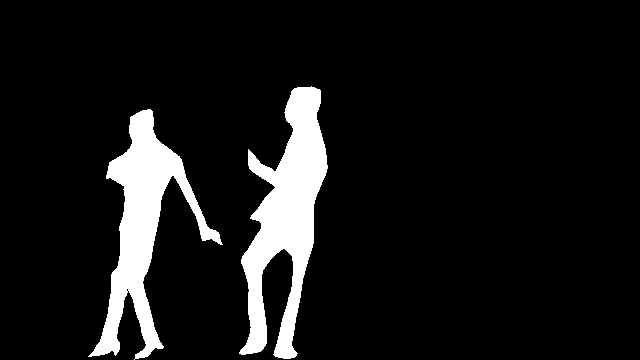} &
        \includegraphics[width=0.1\textwidth, height=1.3cm]{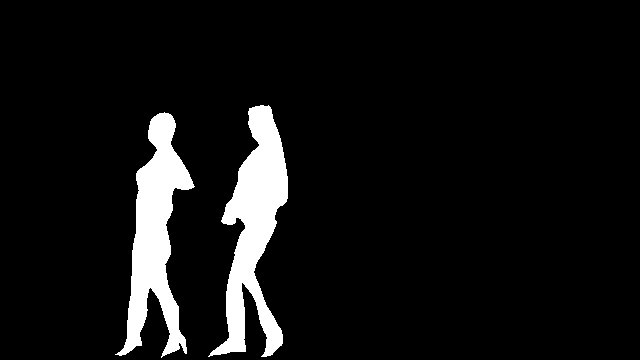} &
        \includegraphics[width=0.1\textwidth, height=1.3cm]{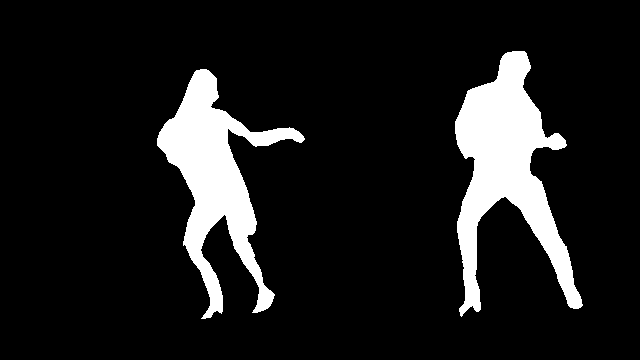}& 
        \includegraphics[width=0.1\textwidth, height=1.3cm]{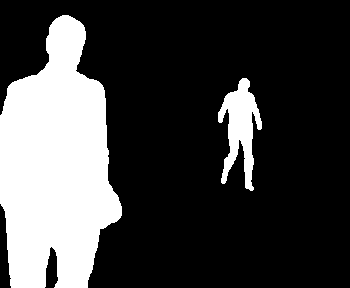} & 
        \includegraphics[width=0.1\textwidth, height=1.3cm]{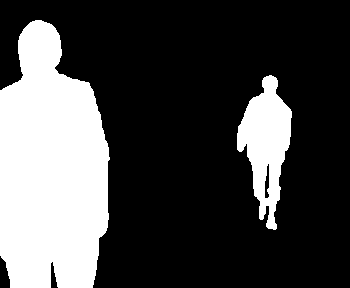} & 
        \includegraphics[width=0.1\textwidth, height=1.3cm]{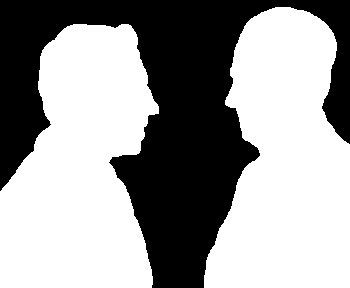}  \\

        \rotatebox{90}{\textbf{MATNet}} &
        \includegraphics[width=0.1\textwidth, height=1.3cm]{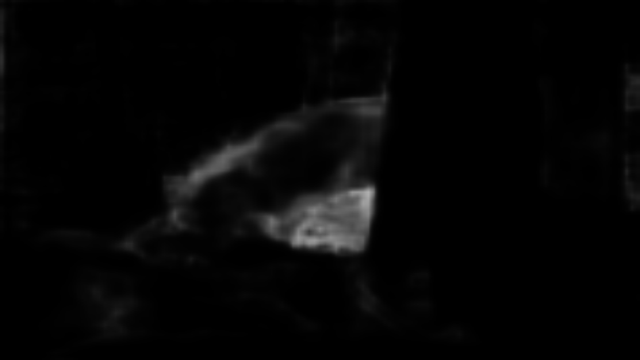} &
        \includegraphics[width=0.1\textwidth, height=1.3cm]{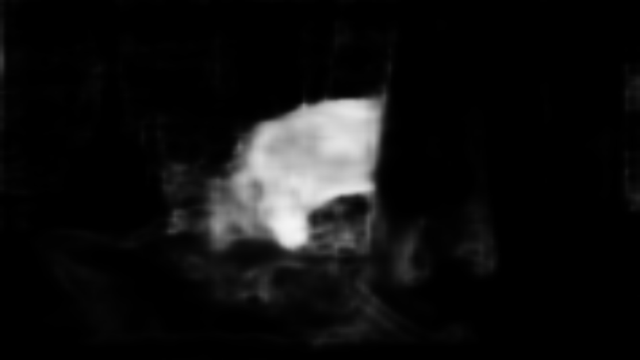} &
        \includegraphics[width=0.1\textwidth, height=1.3cm]{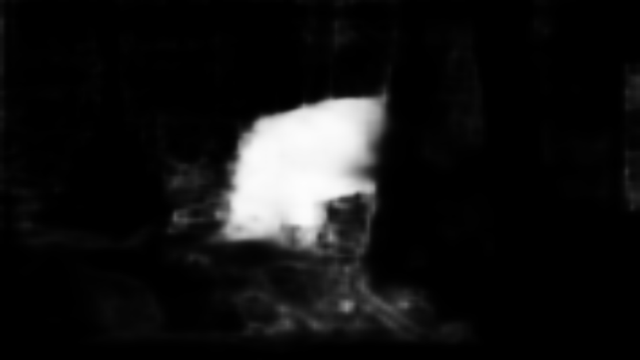} &
        \includegraphics[width=0.1\textwidth, height=1.3cm]{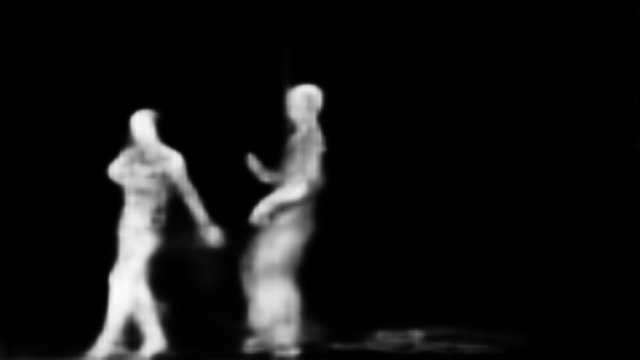} &
        \includegraphics[width=0.1\textwidth, height=1.3cm]{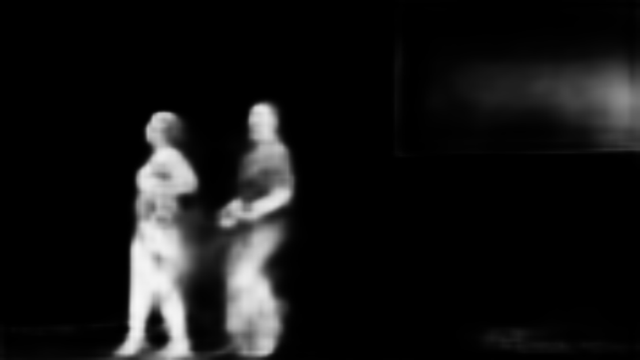} &
        \includegraphics[width=0.1\textwidth, height=1.3cm]{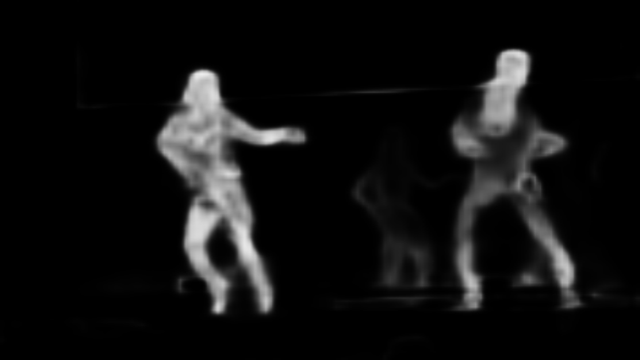}& 
        \includegraphics[width=0.1\textwidth, height=1.3cm]{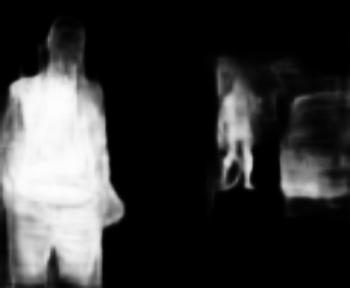} & 
        \includegraphics[width=0.1\textwidth, height=1.3cm]{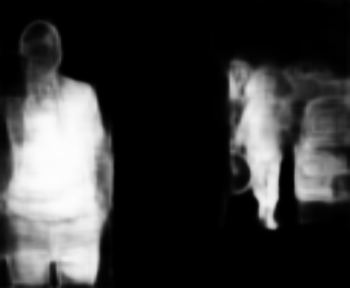} & 
        \includegraphics[width=0.1\textwidth, height=1.3cm]{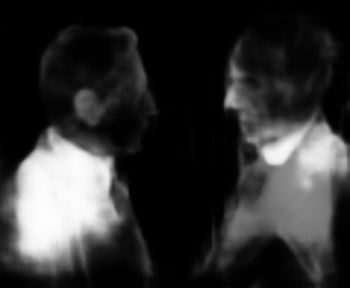}  \\

        \rotatebox{90}{\hspace{0.05cm} \textbf{RTNet}} &
        \includegraphics[width=0.1\textwidth, height=1.3cm]{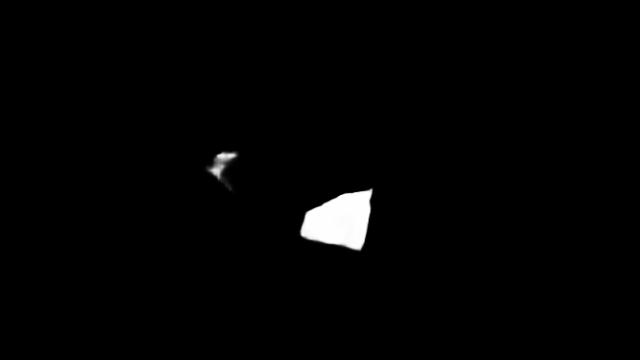} &
        \includegraphics[width=0.1\textwidth, height=1.3cm]{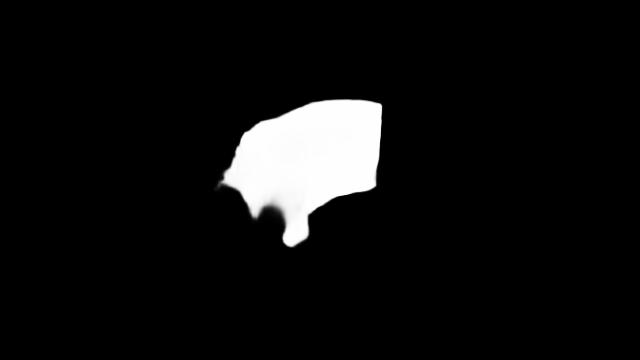} &
        \includegraphics[width=0.1\textwidth, height=1.3cm]{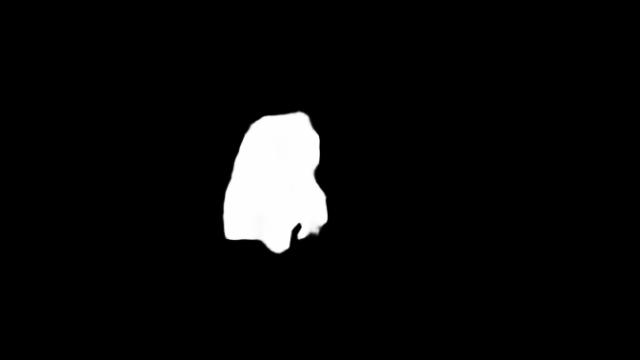} &
        \includegraphics[width=0.1\textwidth, height=1.3cm]{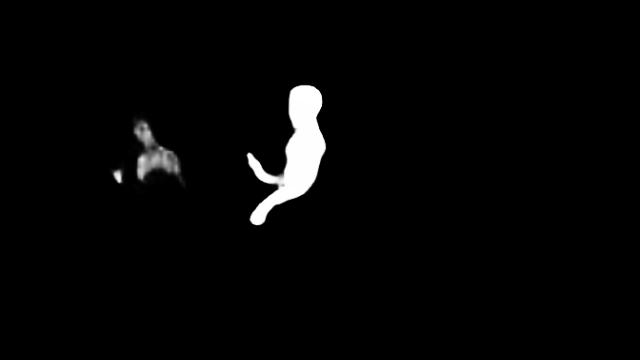} &
        \includegraphics[width=0.1\textwidth, height=1.3cm]{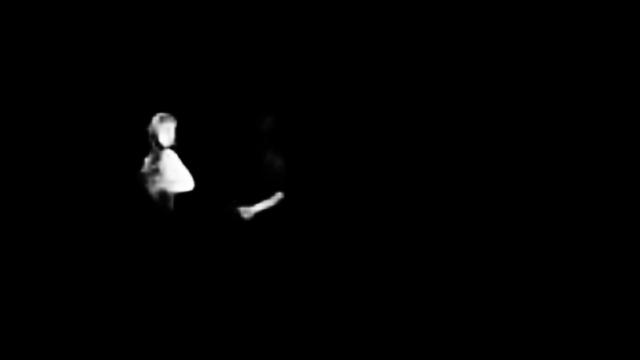} &
        \includegraphics[width=0.1\textwidth, height=1.3cm]{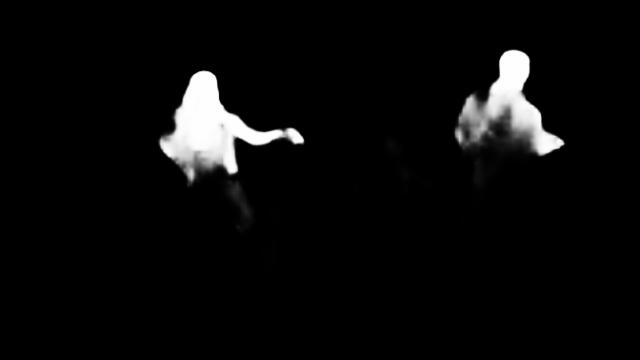}& 
        \includegraphics[width=0.1\textwidth, height=1.3cm]{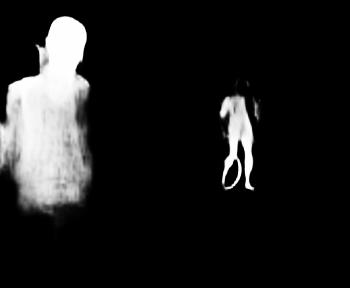} & 
        \includegraphics[width=0.1\textwidth, height=1.3cm]{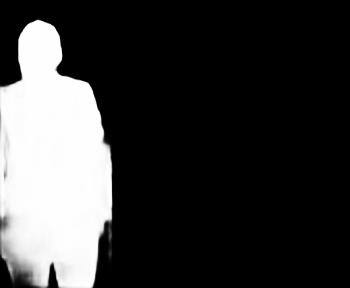} & 
        \includegraphics[width=0.1\textwidth, height=1.3cm]{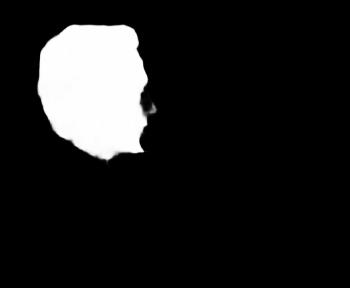}  \\

        \rotatebox{90}{\hspace{0.05cm} \textbf{HFAN}} &
        \includegraphics[width=0.1\textwidth, height=1.3cm]{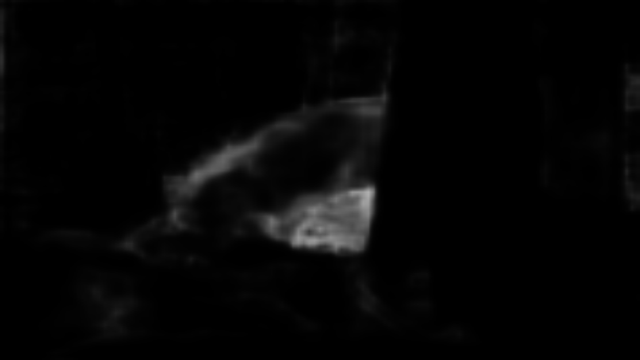} &
        \includegraphics[width=0.1\textwidth, height=1.3cm]{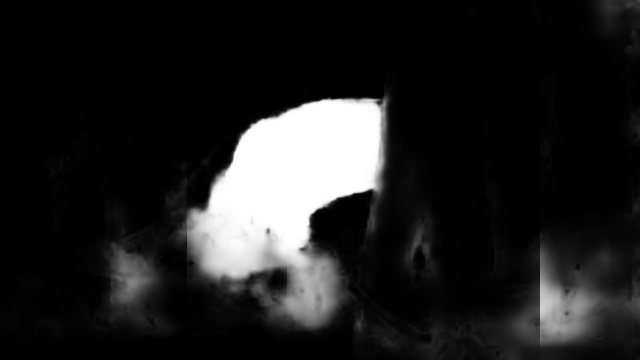} &
        \includegraphics[width=0.1\textwidth, height=1.3cm]{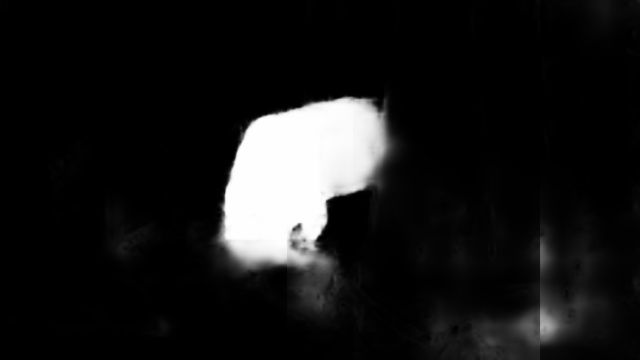} &
        \includegraphics[width=0.1\textwidth, height=1.3cm]{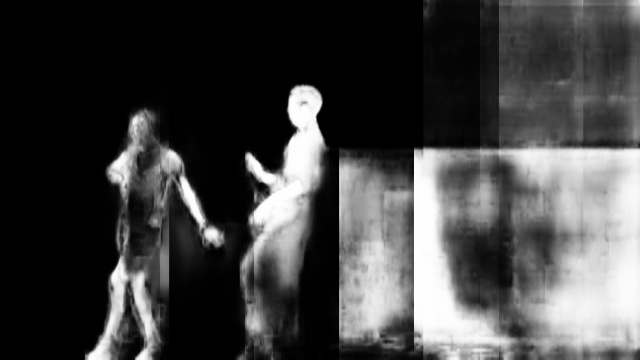} &
        \includegraphics[width=0.1\textwidth, height=1.3cm]{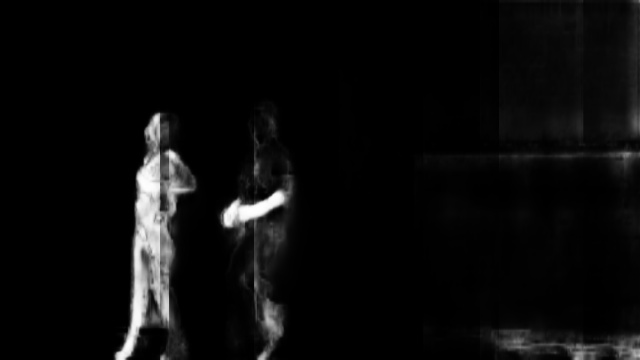} &
        \includegraphics[width=0.1\textwidth, height=1.3cm]{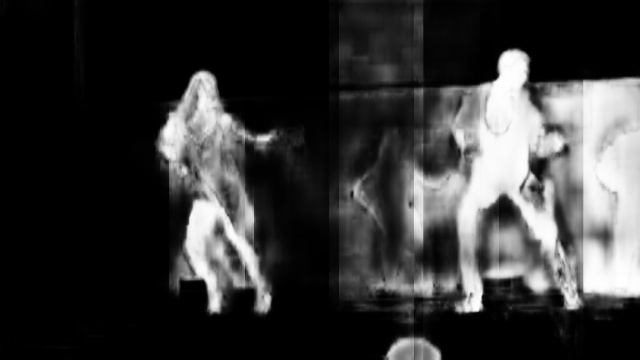}& 
        \includegraphics[width=0.1\textwidth, height=1.3cm]{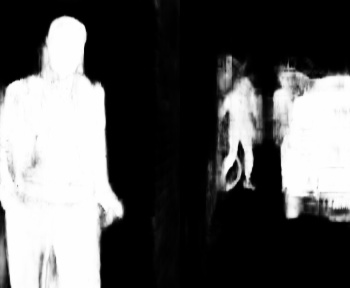} & 
        \includegraphics[width=0.1\textwidth, height=1.3cm]{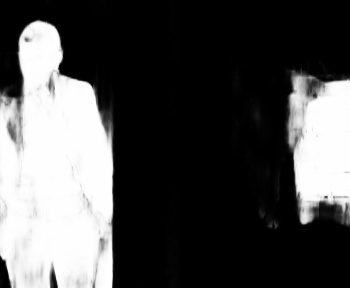} & 
        \includegraphics[width=0.1\textwidth, height=1.3cm]{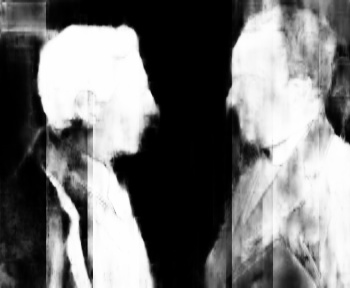}  \\

        \rotatebox{90}{\hspace{0.05cm} \textbf{HCPN}} &
        \includegraphics[width=0.1\textwidth, height=1.3cm]{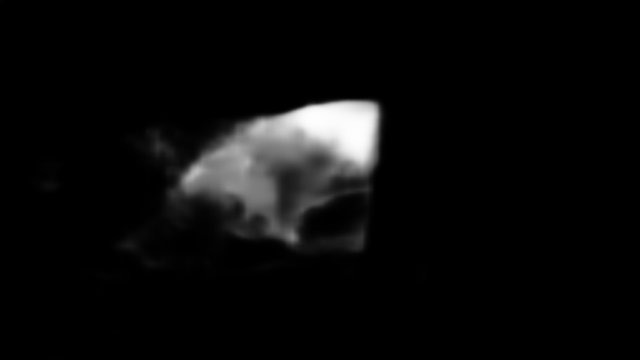} &
        \includegraphics[width=0.1\textwidth, height=1.3cm]{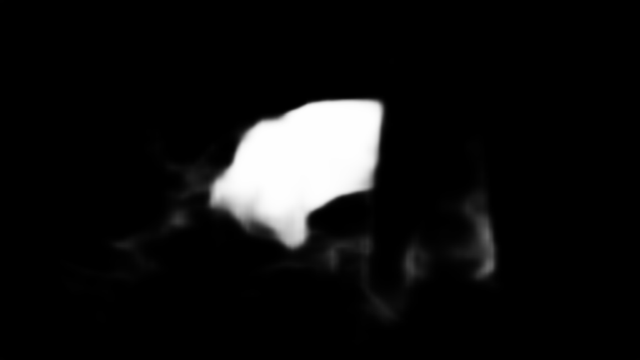} &
        \includegraphics[width=0.1\textwidth, height=1.3cm]{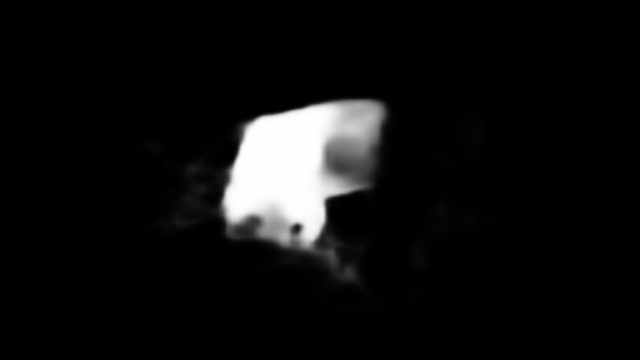} &
        \includegraphics[width=0.1\textwidth, height=1.3cm]{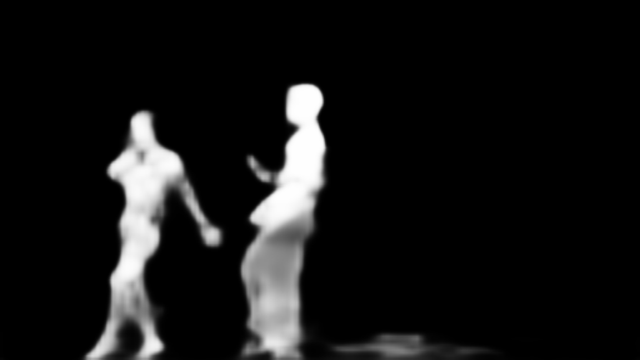} &
        \includegraphics[width=0.1\textwidth, height=1.3cm]{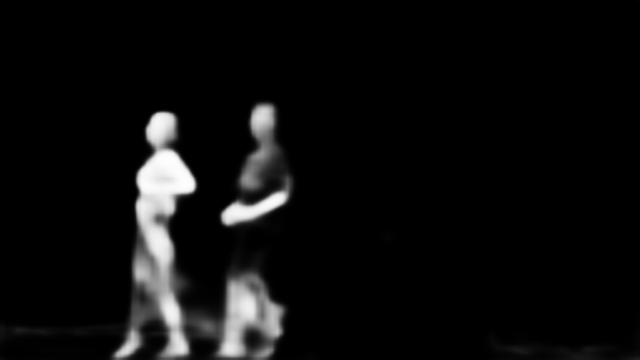} &
        \includegraphics[width=0.1\textwidth, height=1.3cm]{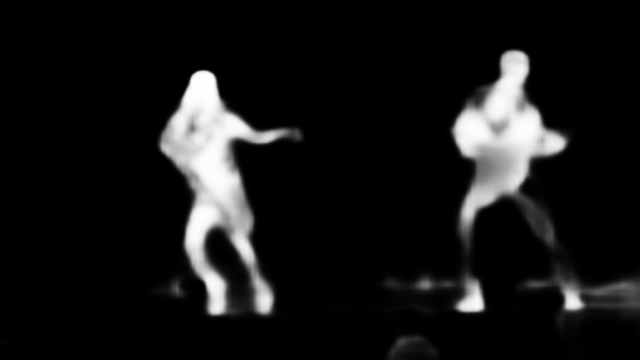}& 
        \includegraphics[width=0.1\textwidth, height=1.3cm]{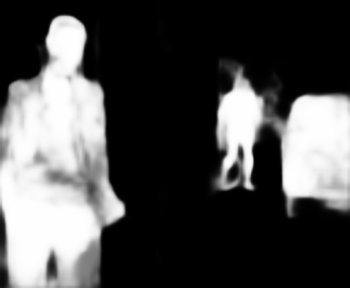} & 
        \includegraphics[width=0.1\textwidth, height=1.3cm]{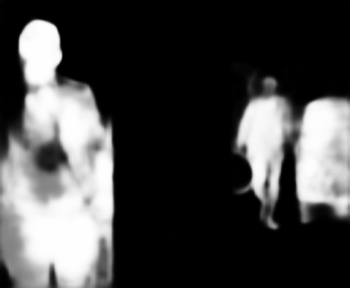} & 
        \includegraphics[width=0.1\textwidth, height=1.3cm]{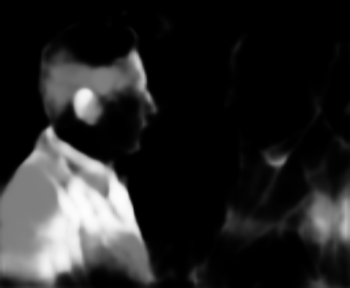}  \\

        \rotatebox{90}{\hspace{0.1cm} \textbf{TMO}} &
        \includegraphics[width=0.1\textwidth, height=1.3cm]{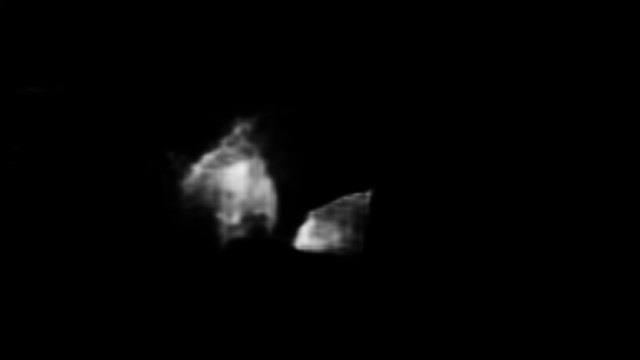} &
        \includegraphics[width=0.1\textwidth, height=1.3cm]{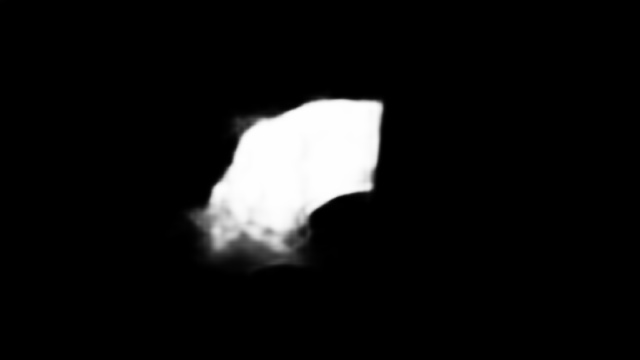} &
        \includegraphics[width=0.1\textwidth, height=1.3cm]{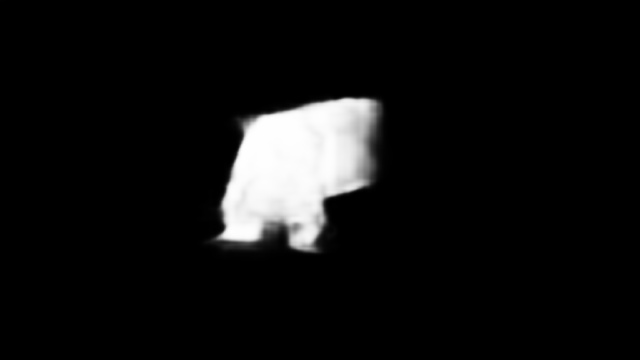} &
        \includegraphics[width=0.1\textwidth, height=1.3cm]{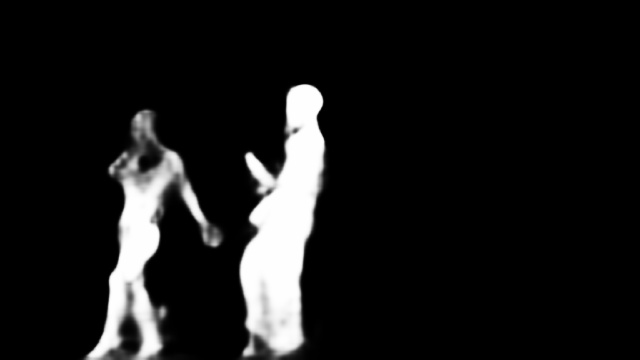} &
        \includegraphics[width=0.1\textwidth, height=1.3cm]{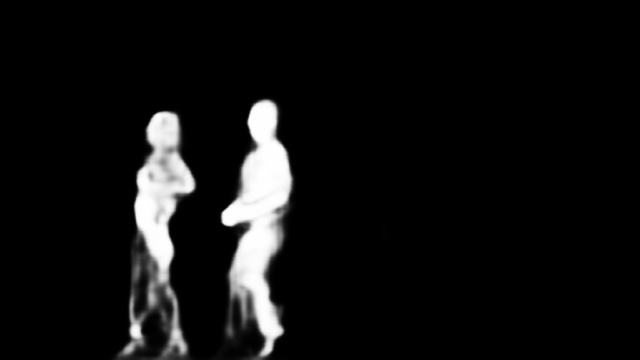} &
        \includegraphics[width=0.1\textwidth, height=1.3cm]{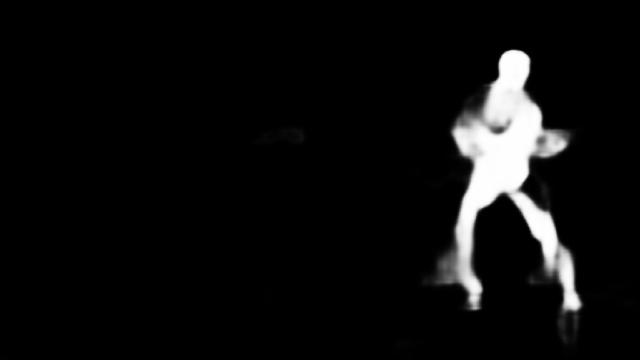}& 
        \includegraphics[width=0.1\textwidth, height=1.3cm]{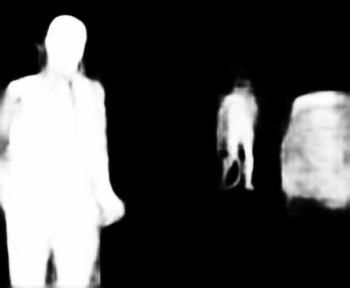} & 
        \includegraphics[width=0.1\textwidth, height=1.3cm]{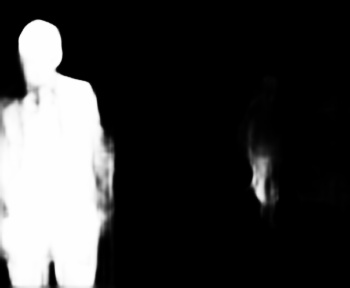} & 
        \includegraphics[width=0.1\textwidth, height=1.3cm]{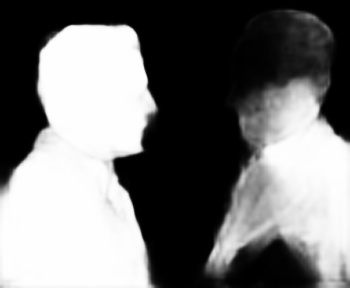}  \\

        \rotatebox{90}{\hspace{0.2cm} \textbf{Ours}} &
        \includegraphics[width=0.1\textwidth, height=1.3cm]{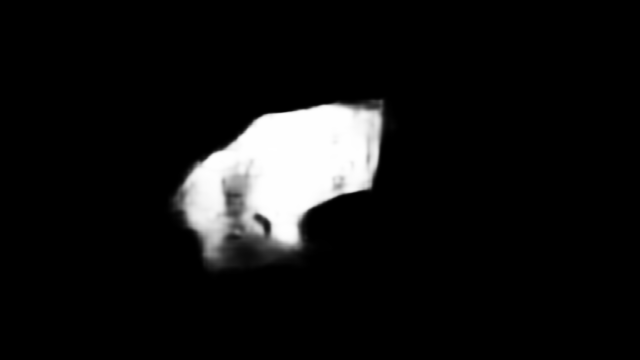} &
        \includegraphics[width=0.1\textwidth, height=1.3cm]{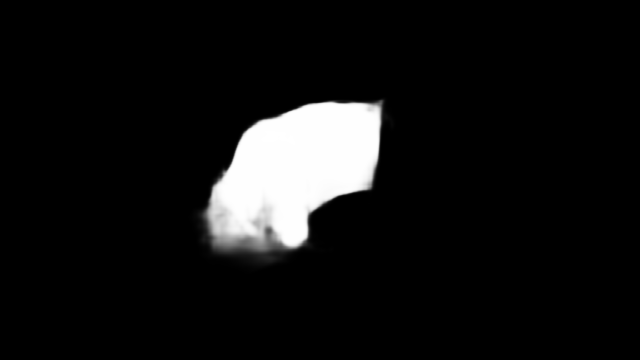} &
        \includegraphics[width=0.1\textwidth, height=1.3cm]{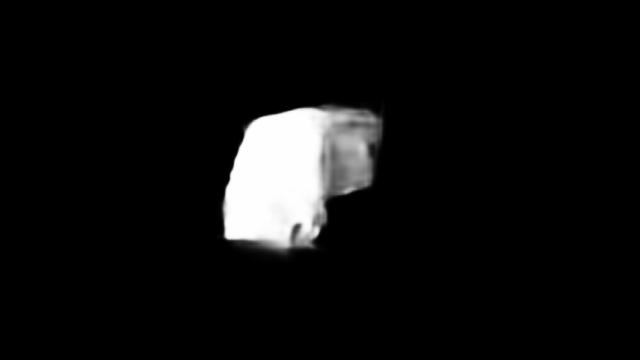} &
        \includegraphics[width=0.1\textwidth, height=1.3cm]{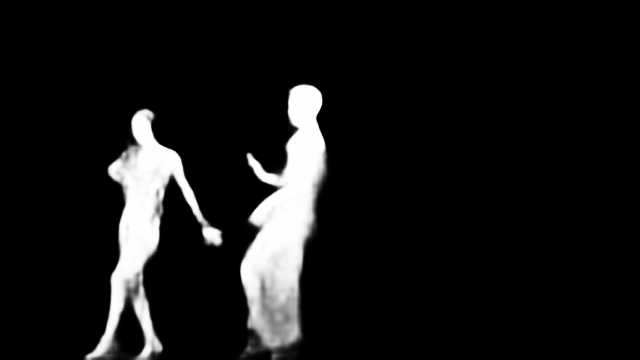} &
        \includegraphics[width=0.1\textwidth, height=1.3cm]{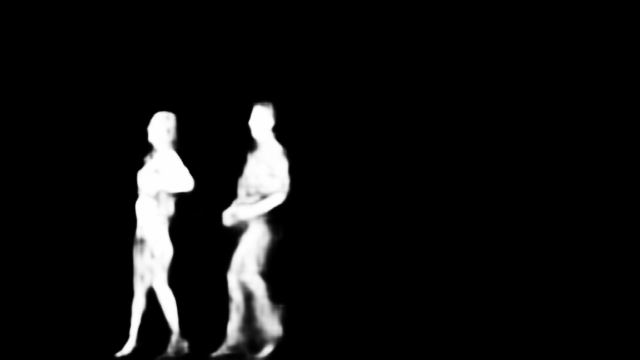} &
        \includegraphics[width=0.1\textwidth, height=1.3cm]{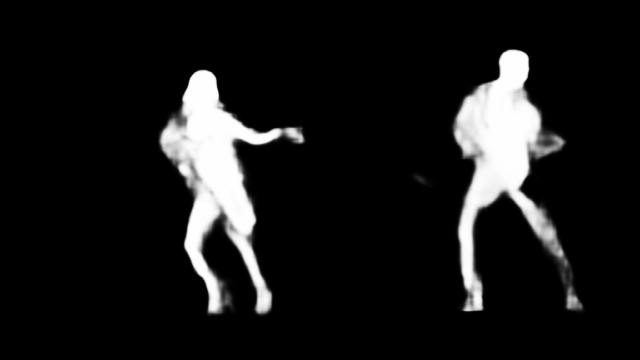}& 
        \includegraphics[width=0.1\textwidth, height=1.3cm]{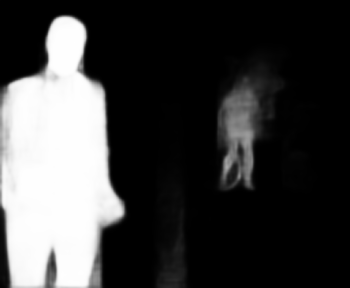} & 
        \includegraphics[width=0.1\textwidth, height=1.3cm]{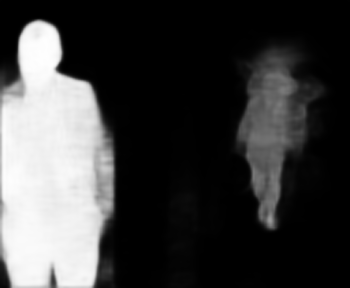} & 
        \includegraphics[width=0.1\textwidth, height=1.3cm]{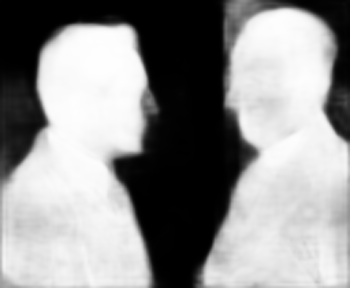}  \\
    
    \end{tabular}
    \caption{Quantitative visualization and comparative analysis of UVOS methods on the the VSOD benchmarks. Case1-2:  \textit{select\_0066}, \textit{select\_0218} from DAVSOD; Case 3: \textit{marple6} from FBMS. We selected five works with publicly available code to compare the visual results of the VSOD task: MATNet, RTNet, HFAN, HCPN, and TMO. Using the weights they provided, we executed the code on the aforementioned VSOD datasets and chose the predicted results of corresponding scene frames for comparison.}
    \label{Qualitative results VSOD}
\end{figure*}

\subsubsection{Video Salient Object Detection}
We further illustrate the visual results of SMTC-Net in recognizing video salient objects on three challenging videos selected from the VSOD and FBMS datasets. The saliency detection maps are presented in Fig. \ref{Qualitative results VSOD}, alongside predictions from prior UVOS algorithms for qualitative comparison.

The first video represents a low-light scenario, where previous methods often struggled to consistently distinguish the entire target across multiple frames. SMTC-Net accurately identifies salient objects even in dimly lit frames, and with significantly higher precision. The second video involves complex human motion and intricate details, posing significant challenges for the VSOD task. Prior approaches frequently produced incomplete detections or identified targets with low confidence. However, SMTC-Net precisely detects both individuals with high confidence, accurately capturing their limbs and other fine-grained details. In the third video, the target object undergoes drastic changes in shape and size. Previous methods either failed to consistently track the target or detected irrelevant objects. In comparison, SMTC-Net maintains high detection accuracy, reliably segmenting the target without interference.

The visual results presented in Fig. \ref{Qualitative results VSOD} highlight the remarkable performance of SMTC-Net across various complex scenarios. These findings emphasize the model's superior capability to handle diverse challenges in VSOD with robustness and accuracy.

\subsection{Ablation Study}
We performed a series of ablation studies to provide insights into the inner workings of SMTC-Net and validate its effectiveness. The conducted experiments focus on four aspects: the effectiveness of the trunk-collateral structure and ISRM; the contribution of appearance and motion features; the influence of the LoRA rank value; and the placement of the collateral branch within the Transformer block. All experiments are carried out on the DAVIS-16 dataset using the mit-b1 \cite{Segformer} backbone.

\subsubsection{Effectiveness of Different Modules}
To assess the contributions of individual modules, we conducted experiments under three conditions: (1) a simple baseline model, (2) the baseline augmented with the trunk-collateral structure, and (3) the baseline enhanced with the ISRM. These experiments aim to demonstrate the impact of each module on the overall model performance.

As shown in Table \ref{tab:module ablation}, both the trunk-collateral structure and ISRM contribute significantly to improving the model's performance over the baseline. Specifically, the inclusion of the trunk-collateral structure leads to a 0.6\% increase in the \(\mathcal{J} \& \mathcal{F}\) metric, while integrating the ISRM results in a 0.8\% improvement. When both the trunk-collateral structure and ISRM are incorporated, the model achieves an even more substantial enhancement, with a \(\mathcal{J} \& \mathcal{F}\) score increase of 1.1\% compared to the baseline.

These results confirm the effectiveness of both the trunk-collateral structure and ISRM, underscoring their critical role in the success of SMTC-Net and validating the overall pattern of our proposed framework.

\begin{table}[h]
\caption{Ablation study on the contribution of different modules: the Trunk-Collateral structure and ISRM}
\label{tab:module ablation}
\centering
\renewcommand{\arraystretch}{1.45}
\begin{tabularx}{0.95\linewidth}{l|c|c|c}
\hline
\textbf{Variant} & \(\boldsymbol{\mathcal{J}}\) & \(\boldsymbol{\mathcal{F}}\) & \(\boldsymbol{\mathcal{J}} \& \boldsymbol{\mathcal{F}}\)\\
\hline
Baseline (Mit-b1) & 87.2 & 88.4 & 87.8\\
Baseline + Trunk-Collateral & 87.8 & 89 & 88.4 (\textbf{+0.6}) \\
Baseline + ISRM & 87.7 & 89.5 & 88.6 (\textbf{+0.8}) \\
Baseline + Trunk-Collateral \& ISRM & 88.2 & 90.2 & 89.2 (\textbf{+1.4}) \\
\hline
\end{tabularx}
\end{table}

\subsubsection{Quantitative Demonstration of the Trunk-Collateral Structure}
\begin{figure}
    \centering
    \setlength{\tabcolsep}{1pt}
    \begin{tabular}{ccccc}
        \textbf{Image} & \textbf{Mask} & \textbf{Flow} & \textbf{Attn\_T} & \textbf{Attn\_T\&C} \\ 
        \includegraphics[width=0.092\textwidth,height=0.065\textwidth]{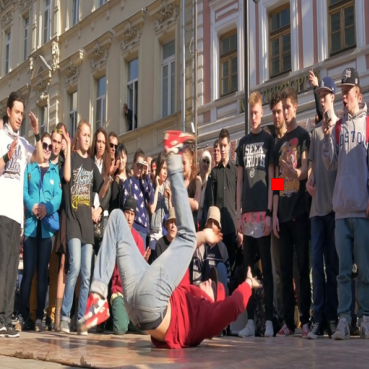} & 
        \includegraphics[width=0.092\textwidth,height=0.065\textwidth]{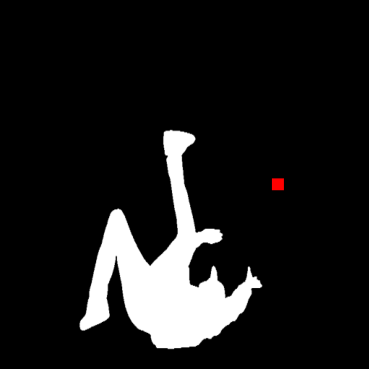} & 
        \includegraphics[width=0.092\textwidth,height=0.065\textwidth]{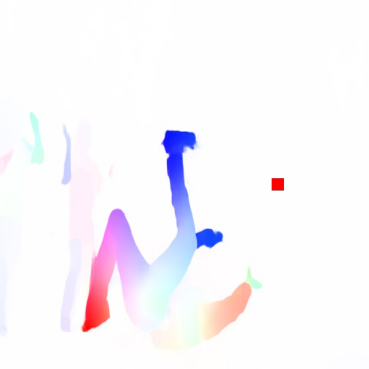}& 
        \includegraphics[width=0.092\textwidth,height=0.065\textwidth]{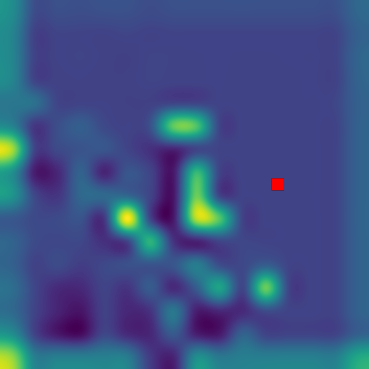}& 
        \includegraphics[width=0.092\textwidth,height=0.065\textwidth]{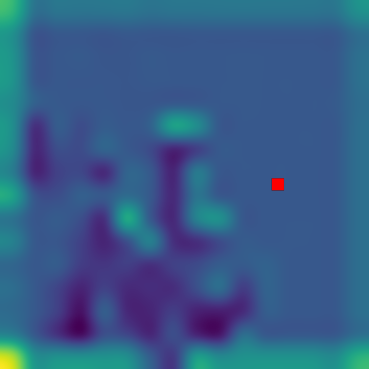} \\ 
        \includegraphics[width=0.092\textwidth,height=0.065\textwidth]{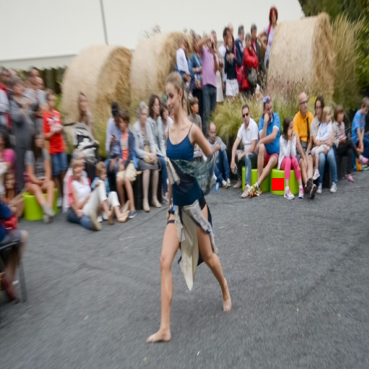} & 
        \includegraphics[width=0.092\textwidth,height=0.065\textwidth]{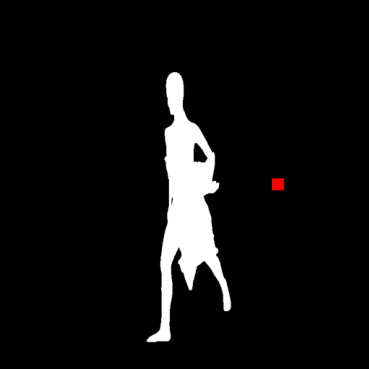} & 
        \includegraphics[width=0.092\textwidth,height=0.065\textwidth]{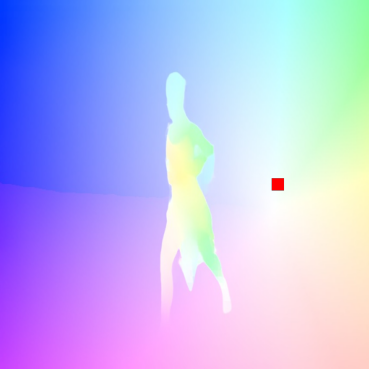}& 
        \includegraphics[width=0.092\textwidth,height=0.065\textwidth]{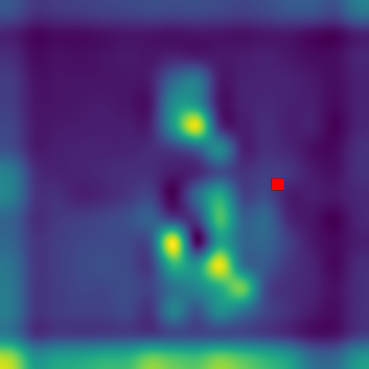}& 
        \includegraphics[width=0.092\textwidth,height=0.065\textwidth]{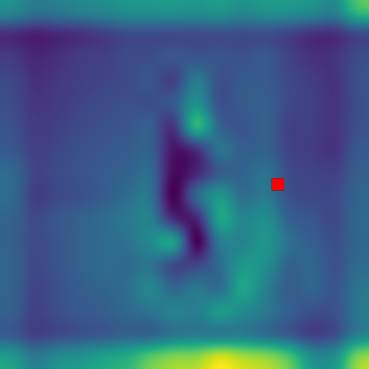} \\ 
         \includegraphics[width=0.092\textwidth,height=0.065\textwidth]{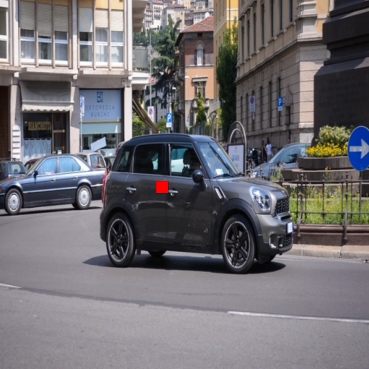} & 
        \includegraphics[width=0.092\textwidth,height=0.065\textwidth]{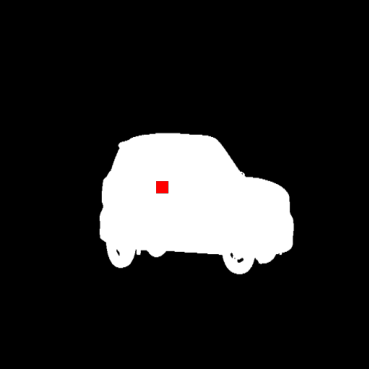} & 
        \includegraphics[width=0.092\textwidth,height=0.065\textwidth]{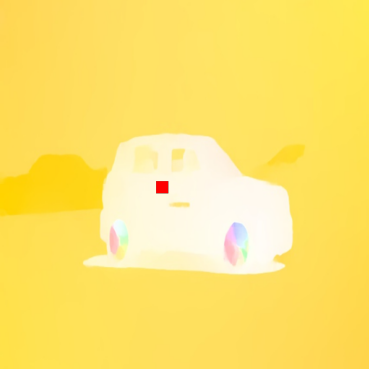}& 
        \includegraphics[width=0.092\textwidth,height=0.065\textwidth]{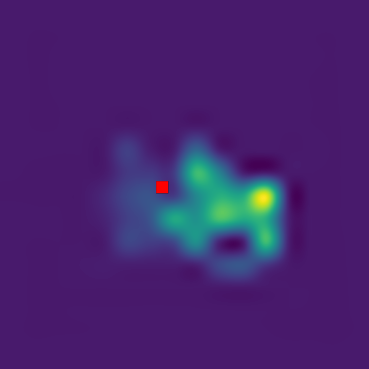}& 
        \includegraphics[width=0.092\textwidth,height=0.065\textwidth]{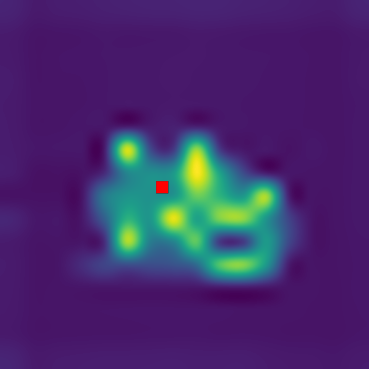} \\ 
         \includegraphics[width=0.092\textwidth,height=0.065\textwidth]{Image/Section4/TC_effect/motocross-jump/img.png} & 
        \includegraphics[width=0.092\textwidth,height=0.065\textwidth]{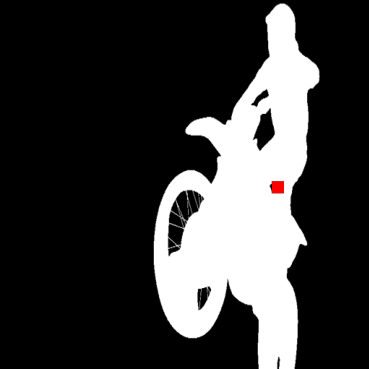} & 
        \includegraphics[width=0.092\textwidth,height=0.065\textwidth]{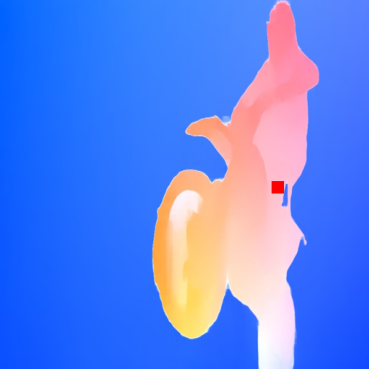}& 
        \includegraphics[width=0.092\textwidth,height=0.065\textwidth]{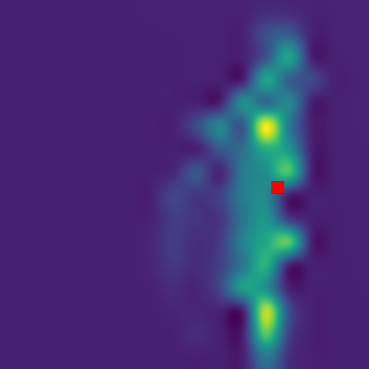}& 
        \includegraphics[width=0.092\textwidth,height=0.065\textwidth]{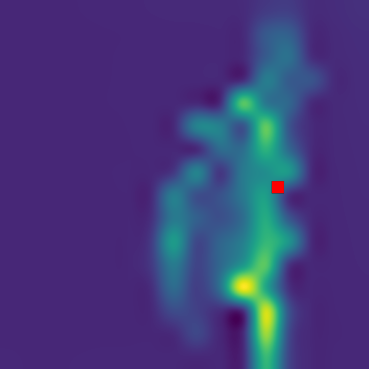} \\ 
         \includegraphics[width=0.092\textwidth,height=0.065\textwidth]{Image/Section4/TC_effect/soapbox/img.png} & 
        \includegraphics[width=0.092\textwidth,height=0.065\textwidth]{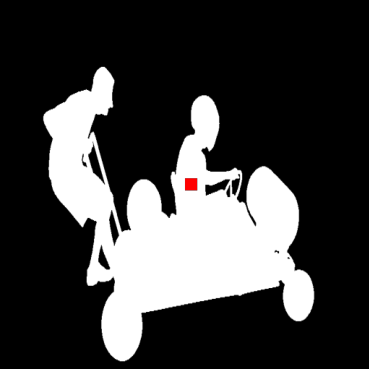} & 
        \includegraphics[width=0.092\textwidth,height=0.065\textwidth]{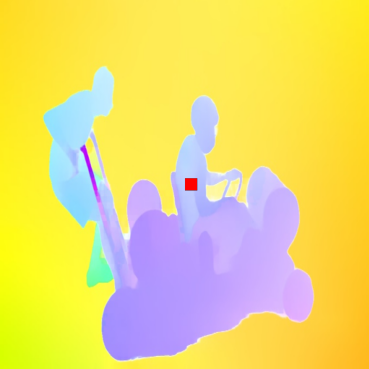}& 
        \includegraphics[width=0.092\textwidth,height=0.065\textwidth]{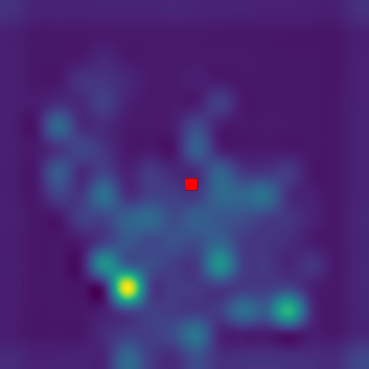}&
        \includegraphics[width=0.092\textwidth,height=0.065\textwidth]{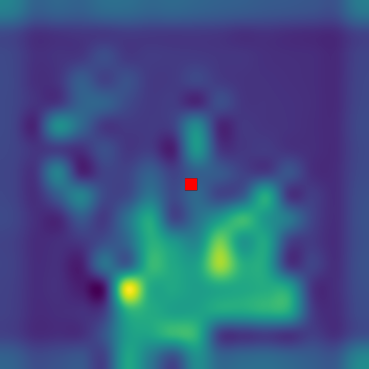} \\ 
    \end{tabular}
    \caption{Visual demonstration of the efficacy of the trunk-collateral structure. \textbf{Attn\_T} represents the attention map obtained using only the trunk section for optical flow encoding. \textbf{Attn\_T\&C} denotes the attention map generated by incorporating both the trunk and collateral branches for optical flow encoding. The small red squares in the images highlight the selected tokens used for analysis.}
    \label{Trunk-Collateral Effectiveness}
\end{figure}

We further illustrate the effectiveness of the trunk-collateral structure in Fig. \ref{Trunk-Collateral Effectiveness}. Specifically, we compare the attention maps of motion features encoded by both the trunk and collateral branches with those derived solely from the trunk branch, as how appearance information is encoded. We analyze the differences in the attention maps for both foreground and background tokens.

The results demonstrate that incorporating the collateral branch enhances the model’s ability to distinguish between foreground and background regions, clarify object boundaries, focus more on foreground details, and achieve higher confidence in the feature representation. These attention maps confirm that our trunk-collateral architecture successfully improves motion encoding, verifying its effectiveness for modeling the interplay between motion and appearance features.

\subsubsection{Impacts of the Collateral Branch at Different Positions}
We present the trunk-collateral structure based on a Transformer-like backbone, integrating the collateral branch into both the feed-forward network (FFN) and multi-head self-attention (MHSA) within each Transformer block. Table \ref{tab:Collateral position ablation} reveals the impact of incorporating the collateral branch at different positions within the Transformer block. The findings indicate that collateral branch at MHSA and FFN enhances the model's performance by 0.3\% and 0.2\% on the \(\mathcal{J} \& \mathcal{F}\) metric, respectively. When applied to MHSA and FFN simultaneously, the model achieves a total performance gain of 0.6\%.

\begin{table}[h]
\caption{Ablation study on the impact of the collateral branch's positions within the Transformer block}
\label{tab:Collateral position ablation}
\centering
\renewcommand{\arraystretch}{1.45}
\begin{tabularx}{0.95\linewidth}{p{3.7cm}|>{\centering\arraybackslash}p{0.7cm}|>{\centering\arraybackslash}p{0.7cm}|>{\centering\arraybackslash}X}
\hline
\textbf{Variant} & \(\boldsymbol{\mathcal{J}}\) & \(\boldsymbol{\mathcal{F}}\) &  \(\boldsymbol{\mathcal{J}}\) \& \(\boldsymbol{\mathcal{F}}\) \\
\hline
Baseline (Mit-b1) & 87.2 & 88.4 & 87.8\\
Baseline + FFN\_Collateral & 87.3 & 88.7 & 88 (\textbf{+0.2}) \\
Baseline + MHSA\_Collateral & 87.5 & 88.7 & 88.1 (\textbf{+0.3}) \\
Baseline + Both\_Collateral & 87.8 & 89 & 88.4 (\textbf{+0.6}) \\
\hline
\end{tabularx}
\end{table}

\subsubsection{Impact of LoRA rank value}
We adopt the low-rank adaptation (LoRA) paradigm in designing the collateral branch to reduce the model parameters and improve the runtime efficiency. LoRA achieves parameter reduction by lowering the intrinsic dimension of dense layers in the adapter, namely the rank value. We analyze the impact of different rank values on model performance through experiments. As shown in Table \ref{tab:Lora rank value ablation}, the model achieved optimal promotion with a rank value of eight, leading to a 0.8\% increase on the \(\mathcal{J} \& \mathcal{F}\) metric. As the rank increases, performance initially improves but later declines, highlighting the critical influence of rank size on model effectiveness.

Since the collateral branch is devised to capture specific motion features, an excessive number of parameters may be unnecessary. However, an insufficient number may hinder the learning of motion-appearance differences. Based on experimental findings, we speculate that rank value affects model performance by regulating the parameter capacity of the collateral branch, which influences its representational power.

\begin{table}[h]
\caption{Ablation study on the impact of the LoRA rank value on model performance}
\label{tab:Lora rank value ablation}
\centering
\renewcommand{\arraystretch}{1.45}
\begin{tabularx}{0.95\linewidth}{>{\centering\arraybackslash}p{3cm}|>{\centering\arraybackslash}p{0.8cm}|>{\centering\arraybackslash}p{0.8cm}|>{\centering\arraybackslash}X}
\hline
\textbf{Rank Value ($r$)} & \(\boldsymbol{\mathcal{J}}\) & \(\boldsymbol{\mathcal{F}}\) & \(\boldsymbol{\mathcal{J}} \& \boldsymbol{\mathcal{F}}\)\\
\hline
Baseline (Mit-b1) & 87.2 & 88.4 & 87.8\\
$r$=1  & 87.4 & 88.9 & 88.15 (\textbf{+0.2}) \\
$r$=2 & 87.8 & 89 & 88.4 (\textbf{+0.6}) \\
$r$=4 & 87.8 & 89.1 & 88.45 (\textbf{+0.65}) \\
$r$=8 & 87.8 & 89.4 & 88.6 (\textbf{+0.8}) \\
$r$=16 & 87.7 & 88.8 & 88.25 (\textbf{+0.35})\\
\hline
\end{tabularx}
\end{table}

\subsubsection{Impact of the Model Inputs}
Motion-appearance UVOS methods typically leverage optical flow and RGB images as inputs to obtain motion and appearance information. To evaluate the contributions of optical flow and RGB images, we conducted experiments and the results are summarized in Table \ref{tab:Image Motion ablation}. Our analysis reveals that using either optical flow or RGB images alone as input can achieve reasonably good performance, with  \(\boldsymbol{\mathcal{J}}\) \& \(\boldsymbol{\mathcal{F}}\) score of 78.8\% and 83.4\% respectively. Notably, images play a more critical role in segmentation due to their rich appearance features. However, images together with optical flow as input yield a noteworthy advancement, demonstrating the complementary nature of motion and appearance information in enhancing the overall effectiveness of UVOS methods.

\begin{table}[h]
\caption{Ablation study on the effect of model inputs}
\label{tab:Image Motion ablation}
\centering
\renewcommand{\arraystretch}{1.45}
\begin{tabularx}{0.95\linewidth}{p{3.5cm}|>{\centering\arraybackslash}p{0.8cm}|>{\centering\arraybackslash}p{0.8cm}|>{\centering\arraybackslash}X}
\hline
\textbf{Input} & \(\boldsymbol{\mathcal{J}}\) & \(\boldsymbol{\mathcal{F}}\) &  \(\boldsymbol{\mathcal{J}}\) \& \(\boldsymbol{\mathcal{F}}\) \\
\hline
Optical Flow & 79.4 & 78.1 & 78.8\\
RGB Image & 83.3 & 83.5 & 83.4 \textbf{(+4.6)}\\
Optical Flow \& RGB Image & 88.2 & 90.2 & 89.2 \textbf{(+10.4)}\\
\hline
\end{tabularx}
\end{table}

\section{Conclusion}
We present SMTC-Net, an innovative end-to-end UVOS framework, which better adapts to the motion-appearance intrinsic relationship while utilizing the model's inherent saliency to enhance performance. SMTC-Net advances a novel trunk-collateral paradigm that regards both the motion-appearance correspondence and uniqueness. Additionally, we put forward the intrinsic saliency guided refinement module, which leverages the model's inherent saliency to optimize high-level representations and guide the motion-appearance integration. Comprehensive experiments demonstrate the superiority of SMTC-Net, which achieved the state-of-the-art results on all commonly used UVOS and VSOD benchmarks. We hope that our model will provide some insights for better utilization of motion and appearance features and inspire future works.

\vfill

\end{document}